\documentclass[lettersize,journal]{IEEEtran}
\usepackage{amsmath,amsfonts}
\usepackage{amssymb}
\usepackage{algorithmic}
\usepackage{algorithm}
\usepackage{array}
\usepackage[caption=false,font=normalsize,labelfont=sf,textfont=sf]{subfig}
\usepackage{textcomp}
\usepackage{stfloats}
\usepackage{url}
\usepackage{verbatim}
\usepackage{cite}
\usepackage{lettrine}
\usepackage{tabularx}
\usepackage{algorithm,algorithmic}
\usepackage{multirow}
\usepackage{footnote}
\usepackage{footmisc}
\usepackage{cuted}
\usepackage{amsmath}
\newcommand{\RNum}[1]{\uppercase\expandafter{\romannumeral #1\relax}}

\ifCLASSINFOpdf
  \usepackage[pdftex]{graphicx}
\else
  \usepackage[dvips]{graphicx}
\fi

\usepackage[pdf]{pstricks}

\usepackage{tikz}
\usepackage{arydshln}

\usepackage[colorlinks=true,linkcolor=black,anchorcolor=black,citecolor=green,filecolor=black,menucolor=black,runcolor=black,urlcolor=black]{hyperref}

\begin{document}

\title{Shared Manifold Learning Using a Triplet Network for Multiple Sensor Translation and Fusion with Missing Data}

\author{Aditya~Dutt,~\IEEEmembership{Graduate Student Member,~IEEE}, Alina~Zare,~\IEEEmembership{Senior Member,~IEEE}, Paul~Gader,~\IEEEmembership{Fellow,~IEEE}


\IEEEcompsocitemizethanks{\IEEEcompsocthanksitem Aditya Dutt is with the Department of Computer \& Information Science \& Engineering, University of Florida, Gainesville, FL 32611 USA. (e-mail: \href{mailto:aditya.dutt@ufl.edu}{aditya.dutt@ufl.edu})
\IEEEcompsocthanksitem Alina Zare is with the Department of Electrical \& Computer Engineering, University of Florida, Gainesville, FL 32611 USA. (e-mail: \href{mailto:azare@ufl.edu}{azare@ufl.edu})
\IEEEcompsocthanksitem Paul Gader is with the Department of Computer \& Information Science \& Engineering, and the Department of Environmental Engineering Sciences, University of Florida, Gainesville, FL 32611 USA. (e-mail: \href{mailto:pgader@ufl.edu}{pgader@ufl.edu})
}
}

\markboth{Accepted to IEEE Journal of Selected Topics in Applied Earth Observations and Remote Sensing}%
{Shell \MakeLowercase{\textit{et al.}}: A Sample Article Using IEEEtran.cls for IEEE Journals}


\IEEEoverridecommandlockouts
\newcommand\copyrighttext{%
  \footnotesize \textcopyright 2022 IEEE. Personal use of this material is permitted.  Permission from IEEE must be obtained for all other uses, in any current or future media, including reprinting/republishing this material for advertising or promotional purposes, creating new collective works, for resale or redistribution to servers or lists, or reuse of any copyrighted component of this work in other works. DOI: {10.1109/JSTARS.2022.3217485}}
  
\newcommand\copyrightnotice{%
\begin{tikzpicture}[remember picture,overlay]
\node[anchor=south,yshift=10pt] at (current page.south) {\parbox{\dimexpr\textwidth-\fboxsep-\fboxrule\relax}{\copyrighttext}};

\end{tikzpicture}%
}

\maketitle

\copyrightnotice

\begin{abstract}
Heterogeneous data fusion can enhance the robustness and accuracy of an algorithm on a given task. However, due to the difference in various modalities, aligning the sensors and embedding their information into discriminative and compact representations is challenging. In this paper, we propose a Contrastive learning based MultiModal Alignment Network (CoMMANet) to align data from different sensors into a shared and discriminative manifold where class information is preserved. The proposed architecture uses a multimodal triplet autoencoder to cluster the latent space in such a way that samples of the same classes from each heterogeneous modality are mapped close to each other. Since all the modalities exist in a shared manifold, a unified classification framework is proposed. The resulting latent space representations are fused to perform more robust and accurate classification. In a missing sensor scenario, the latent space of one sensor is easily and efficiently predicted using another sensor's latent space, thereby allowing sensor translation. We conducted extensive experiments on a manually labeled multimodal dataset containing hyperspectral data from AVIRIS-NG and NEON, and LiDAR (light detection and ranging) data from NEON. Lastly, the model is validated on two benchmark datasets: Berlin Dataset (hyperspectral and synthetic aperture radar) and MUUFL Gulfport Dataset (hyperspectral and LiDAR). A comparison made with other methods demonstrates the superiority of this method. We achieved a mean overall accuracy of 94.3\% on the MUUFL dataset and the best overall accuracy of 71.26\% on the Berlin dataset, which is better than other state-of-the-art approaches.
\end{abstract}


\section{Introduction}
\label{sec:Introduction}
\begin{figure}
  \begin{centering}
  \includegraphics[scale=0.125]{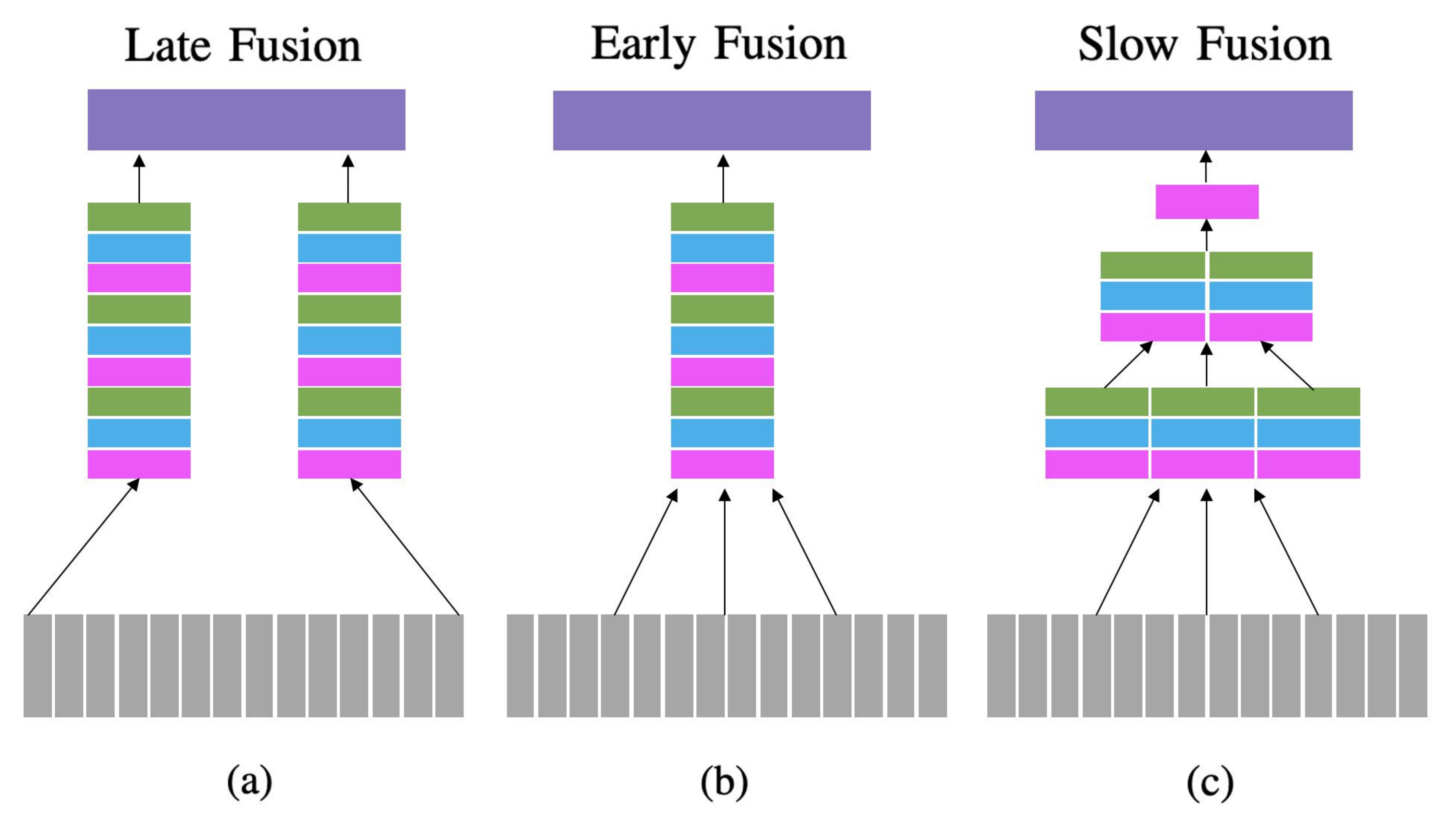}
  \caption{Data Fusion Methods: a) Late fusion, b) Early fusion, c) Intermediate/ Slow fusion}
  \label{FusionMethods} 
  \end{centering}
\end{figure}

\lettrine{M}{ultimodal} information fusion architectures have significantly outperformed unimodal models and achieved outstanding results on tasks like land-use and land-cover classification (LULC)\cite{khodadadzadeh2015fusion} \cite{matsuki2015hyperspectral}, mineral exploration \cite{guha2020mineral} \cite{noauthor_hunting_nodate} \cite{peyghambari2021hyperspectral}, urban planning \cite{hansch2020fusion}, biodiversity conservation \cite{kolmann2021hyperspectral}, sentiment detection \cite{chen2019complementary} \cite{yoon2018multimodal} \cite{tsai2019multimodal}, speech recognition, word sense disambiguation, fact extraction, and media description. In certain situations, one sensor is not sufficient to obtain robust performance. The conventional approach in multimodal fusion is to concatenate the representations of different modalities. This can be further divided into three categories as shown in Fig. \ref{FusionMethods}:

\begin{enumerate}
    \item \textit{Early Fusion:} In early fusion, the low-level features are extracted from each modality which are fused before being classified. However, the fusion of heterogeneous data sources into a fixed-size representation is challenging. Additionally, the model can lose important information to generate a common representation.
    \item \textit{Late Fusion:} In late fusion, the representation from each modality is classified, and a decision is made using methods like majority voting\cite{liao2014combining} \cite{pandeya2021deep}. This method is also called \textit{decision fusion}.
    \item \textit{Intermediate Fusion:} Intermediate fusion or Hybrid fusion is the most reliable and flexible fusion method. In this case, the intermediate representations of modalities are merged. In the context of a neural network, these representations are generated by the convolutional layers and fused gradually to form a shared representation layer.
\end{enumerate}

Fusion methods can be classified into two groups: concatenation and alignment-based methods. Usually, concatenation-based methods extract the features using deep learning or other machine learning models and fuse the information for classification. Several new techniques have recently been developed for joint data classification using concatenated representations. It is essential to extract rich structural and texture information from the sensors to make a robust classification. To address this issue, Liao et al. \cite{liao2014generalized} used morphological features to learn spatial information by using a structured morphological element of predefined size and shape. They proposed a graph-based model to couple the dimension reduction and fusion of information. However, using this method, the cloud-covered regions are not accurately classified because the morphological features of LiDAR are not computed properly. Morphological profiles have other limitations due to a fixed size structuring element. To overcome this issue, Rasti et al. \cite{rasti2017hyperspectral} used extinction profiles to extract spatial and spectral information from the LiDAR and hyperspectral data, which were fused using Orthogonal Total Variation Component Analysis (OTVCA). Extinction profiles are extrema-oriented filters, and experiments have shown that they perform better than morphological filters. However, after extracting the extinction profiles, a simple stacking of features is not efficient for classification. To address this, Zhao et al. \cite{zhao2020joint} proposed a hierarchical random walk network to capture the spatial and spectral features. The random walk also reduces the problem of weak localization around boundaries. To extract rich spectral features, Zhao et al. \cite{zhao2021fractional} used Gabor convolutional layers to extract the multi-directional, multi-scale, and semantic change features along with the Octave convolutional layers. Their model is able to capture directional texture features and frequency variation features efficiently. \par 

After extracting the spatial and spectral features, an efficient fusion strategy is required. Hong et al. \cite{hong2020more} proposed an end-to-end unified deep learning model for remote sensing imagery classification. They developed two deep learning models: Ex-Net to extract information and Fu-Net for data fusion. They used cross-modality learning and multi-modality learning to enhance classification accuracy. However, their model still depends on a large number of samples to yield good results. To reduce the dependency on the number of samples, Hang et al. \cite{hang2020classification} proposed an unsupervised coupled CNN framework for hyperspectral and LiDAR data. They provided each modality as the input to these CNNs. This coupled convolution guides the CNNs to learn useful representations from each other, which helps with the fusion process. Furthermore, they used both feature level and decision level fusion to enhance the fusion process. Several unsupervised CNN architectures were also proposed \cite{zhao2016efficient} \cite{hong2020deep} \cite{zhang2018feature} \cite{hong2020x} \cite{zhang2022artificial} to perform sensor fusion. \par

During the fusion process, it is essential to preserve complementary information from all the heterogeneous modalities. The presence of redundant features during fusion decreases the classification performance. To tackle this problem, Zhang et al. \cite{zhang2021information} used an Information Fusion network (IP-CNN) to learn complementary information from heterogeneous sensors. They utilized the Gram matrices to achieve this task. This concept is similar to Neural Style Transfer \cite{xu2021image}. They used the gram matrices from LiDAR as a texture reference to the fused embeddings. Similarly, the HSI gram matrix serves as a spectral reference to the fused embeddings. Therefore, the fused embeddings carry spatial and spectral information from the original modalities. Guo et al. \cite{guo2019degraded} also utilized the Gram matrices to control the image texture difference between the clean and degraded images. However, interpretability is a big issue in deep learning models. To increase the interpretability of fusion models, Hong et al. \cite{hong2021multimodal} proposed a shared and specific feature learning (S2FL) that is capable of decomposing data into modality-shared and modality-specific components, which enables a better information blending of multiple heterogeneous modalities. \par

Instead of the direct fusion process, many researchers have tried to develop alignment models where data from all modalities can be mapped onto a shared manifold while preserving the characteristics of all classes. This method provides an advantage by reducing the ambiguity in classification when only one modality is used. Therefore, the modalities with low discriminative ability can also perform a better classification. However, the decoupled parts of embeddings should not contribute when learning the common features. Hong et al. \cite{hong2019cospace} developed a common subspace learning (CoSpace) model that learns shared feature representations from hyperspectral and multispectral correspondences. CoSpace achieved state-of-the-art results on the classification task. Their shared space also allowed sensor translation. Pournemt et al. \cite{pournemat2021semisupervised} proposed a semi-supervised alignment approach that is based on utilizing only the common knowledge present in the shared representations. They claimed that the decoupled information from different modalities obstructs the alignment process. To achieve this, they used the joint spectral analysis of the graph Laplacians of the different modalities. Their model showed promising results for real-world multimodal problems.\par  

The alignment models are better than the concatenation models because they increase the classification performance in both unimodal and multimodal scenarios. In recent years, contrastive learning has gained much attention. Contrastive learning is an efficient way to address the data scarcity problem and align data from multiple modalities simultaneously. In 2015, Hoffer et al. \cite{hoffer2015deep} proposed Triplet Networks, which can learn representations by distance comparisons. After that, Triplet Networks have been successfully applied to several applications. In this paper, we explore the usage of contrastive learning to tackle the multimodal manifold alignment challenge. Most of the methods discussed so far focus on joint representation learning and do not perform sensor translation. According to Baltrusaitis et al. \cite{baltruvsaitis2018multimodal}, there are five technical challenges in a multimodal setting: Representation, Translation, Alignment, Fusion, and Co-learning. We simultaneously address the first four challenges in our framework.\par 



This paper focuses on generating a discriminative shared manifold for multimodal data. The shared manifold is generated in such a way that the samples of a class from all the modalities are mapped close to each other while dissimilar classes are pushed apart. The proposed architecture contains one Triplet autoencoder and one standard autoencoder. Each autoencoder takes a separate modality as its input. An objective function is proposed based on the triplet loss, which encourages the latent space of both the autoencoders to be similar for the same class and dissimilar for different classes. To further bring the latent space of both modalities close to each other, we proposed a similarity enhancement term. After training, the resulting latent space embeddings are highly discriminative. For classification, we combine the latent space of all modalities and apply a KNN and shallow neural network. In order to perform sensor translation, we developed a regression network to predict the latent space of a missing sensor from the available sensor's latent space. Since the latent space is already clustered, this process becomes significantly easier. Once the latent space of the missing sensor is predicted, it can be reconstructed to generate the full-scale data. Experimental results show that the overall classification accuracy outperforms all the other models on MUUFL Gulfport \cite{gader2017muufl} and Berlin Datasets \cite{hong2021multimodal}. \par 

The main contributions are highlighted as follows:

\begin{enumerate}
    \item \textit{Shared Manifold Representation/ Alignment:} The proposed architecture aligns the embeddings from all the sensors into a common shared manifold. Since an autoencoder is used in this architecture, it is ensured that the latent space embeddings capture all the information necessary to reconstruct the sensor data. Additionally, the latent space is discriminative. The proposed model is not limited to remote sensing applications.
    \item \textit{Sensor Translation:} A shallow regression network is developed to predict a missing sensor's embeddings from other available sensors. The predicted embeddings can be then reconstructed using a decoder, allowing sensor translation.
    \item \textit{Classification:} Classification is performed using the fused embeddings of all the sensors. Furthermore, classifiers are created for single modalities as well. The representations show high classification accuracy using even a simple model like KNN.

\end{enumerate}

\section{Proposed Method}
\label{sec:Method}
\subsection{CoMMANet: Shared Manifold Generation Architecture}
\begin{figure*}[ht]
  \begin{centering}
  \includegraphics[scale=0.185]{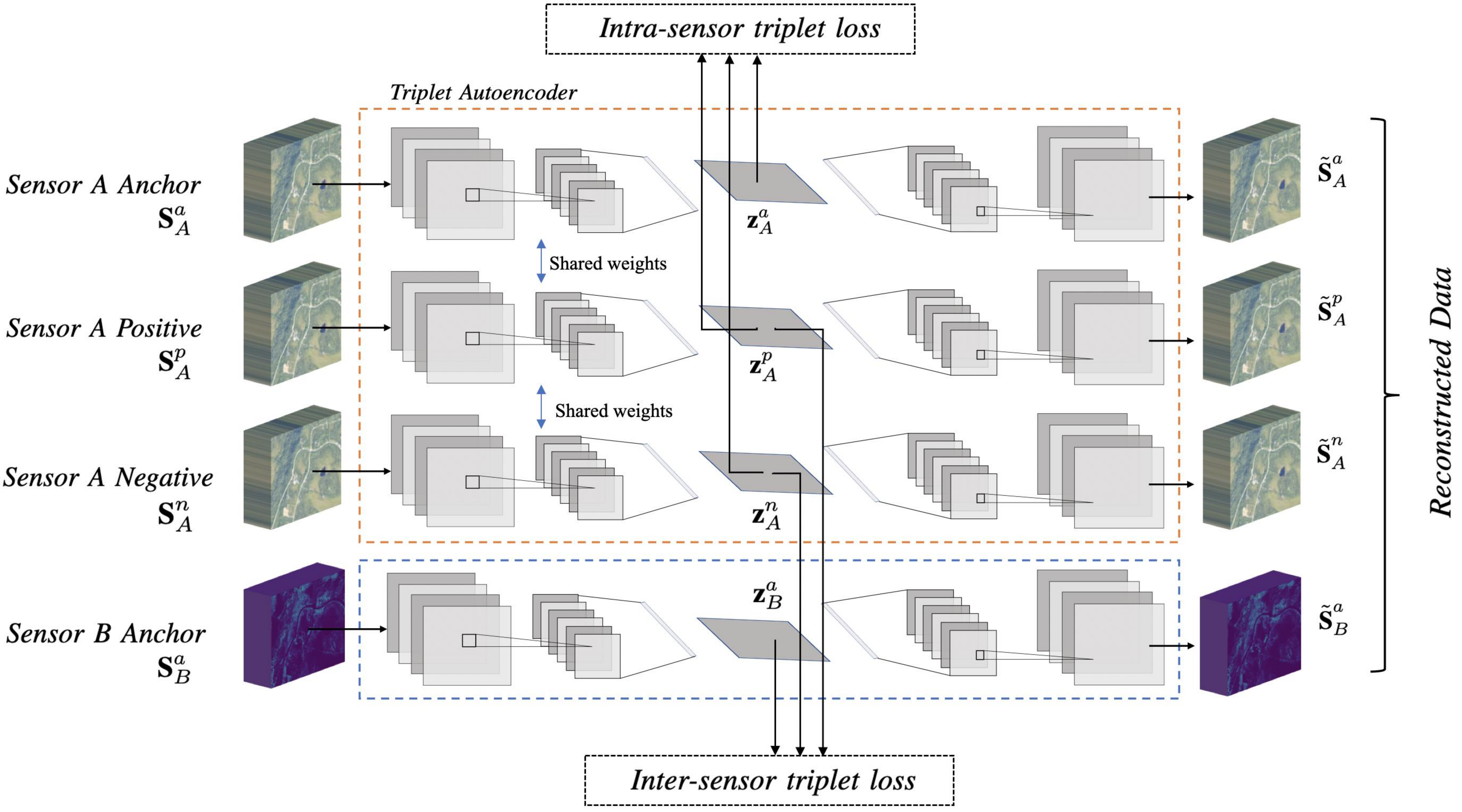}
  \caption{The proposed CoMMANet architecture for the shared manifold generation.}
  \label{SharedModel} 
  \end{centering}
\end{figure*}

The proposed framework, CoMMANet, consists of two autoencoders as shown in Fig. \ref{SharedModel}. The first autoencoder is a triplet autoencoder which is implemented on \textit{sensor} $A$. The autoencoder for \textit{sensor} $B$ is a standard autoencoder. The encoder of the \textit{sensor} $A$ is denoted by the embedding function, $\textbf{e}_{A}\left(.\right)$. The decoder of the \textit{sensor} $A$ is denoted by the embedding function, $\textbf{d}_{A}\left(.\right)$. Similarly, the encoder and decoder of \textit{sensor} $B$ are denoted by the embedding functions, $\textbf{e}_{B}\left(.\right)$ and $\textbf{d}_{B}\left(.\right)$, respectively. A standard CNN architecture is used for encoders ($\textbf{e}_{A}\left(.\right)$ and $\textbf{e}_{B}\left(.\right)$) and decoders ($\textbf{d}_{A}\left(.\right)$ and $\textbf{d}_{B}\left(.\right)$) in this paper.\par 

The inputs for the triplet autoencoder will be three samples from \textit{sensor} $A$ denoted by $\textbf{S}_{A}^{a}$, $\textbf{S}_{A}^{p}$, and $\textbf{S}_{A}^{n}$, where $\textbf{S}_{A}^{a}$ is the anchor, $\textbf{S}_{A}^{p}$ is the positive, and $\textbf{S}_{A}^{n}$ is the negative sample. The anchor and the positive samples share the same label. The negative sample belongs to a class other than the anchor's class. The latent space or embeddings of \textit{sensor} $A$ are denoted by: 

\begin{equation}
\label{eq1}
    \textbf{z}_{A} = \textbf{e}_{A}\left(\textbf{S}_{A}\right)
\end{equation}

\noindent \color{black}where $\textbf{S}_{A} \in {\rm I\!R}^{N_1}$ and $N_1$ is the dimensionality of  data extracted from sensor $A$, and $\mathbf{z}_A \in \mathbb{R}^{D}.$

\color{black}\textrm{This implies,} 

\begin{equation}
     \textbf{z}_{A}^{t} = \textbf{e}_{A}\left(\textbf{S}_{A}^{t}\right)\text{ for } t \in \{a,p,n\}
\end{equation}



\noindent where $\textbf{z}_{A}^{t} \in {\rm I\!R}^{D}$ for $t \in \{a,p,n\}$, and $D$ is the dimensionality of the latent space.

The reconstructed outputs of \textit{sensor} $A$ from the decoder are denoted by:


\begin{equation}
    \tilde{\textbf{S}}_{A}^{t}= \textbf{d}_{A}\left(\textbf{z}_{A}^{t}\right) \text{ for } t \in \{a,p,n\}
\end{equation}


\noindent where $\tilde{\textbf{S}}_{A}^{t} \in {\rm I\!R}^{N_1}$ for $t \in \{a,p,n\}$.

There is only one input from the \textit{sensor} $B$, $\textbf{S}_{B}^{a}$, which represents the anchor from \textit{sensor} $B$. Similarly, the latent space or embeddings of \textit{sensor} $B$ anchor are denoted by:

\begin{equation}
    \textbf{z}_{B}^{a} = \textbf{e}_{B}\left(\textbf{S}_{B}^{a}\right)
\end{equation}
\noindent where $\textbf{S}_{B}^{a} \in {\rm I\!R}^{N_2}$,  $\textbf{z}_{B}^{a} \in {\rm I\!R}^{D}$, $N_2$ is the dimensionality of data extracted from sensor $B$, and $D$ is the dimensionality of the latent space.

Similarly, the reconstructed output of \textit{sensor} $B$ anchor is denoted by:

\begin{equation}
    \tilde{\textbf{S}}_{B}^{a}= \textbf{d}_{B}\left(\textbf{z}_{B}^{a}\right)
\end{equation}
\noindent where $\tilde{\textbf{S}}_{B}^{a} \in {\rm I\!R}^{N_2}$.

The goal of training a standard triplet network is to minimize an objective function.  The objective function has the following properties:
\begin{itemize}

\item Decrease when the distances between the anchor and positive sample embeddings decrease i.e., the distance between samples of the same classes decreases.
\item Decrease when the distances between the anchor and negative sample embeddings increase i.e., the distance between dissimilar classes increases.
\end{itemize}


However, this triplet loss is limited to one modality only. For a multimodal setting, we propose a multimodal triplet loss objective function. The multi-modal triplet loss has the following properties:
\begin{itemize}

\item Decrease when the distances between embeddings of samples from the same class decrease irrespective of the sensor.
\item Decrease when the distances between embeddings of samples from different classes and the same/ different sensors increase.
\end{itemize}
The objective function to train the CoMMANet is shown below in Eq. \ref{eq:eq_commanet}.

\begin{equation}
    \label{eq:eq_commanet}
    \textbf{L}_{CommaNet} = \textbf{L}_T +  \textbf{L}_E + \textbf{L}_{SE}
\end{equation}

The working and influence of all these loss function terms are explained below:\par

\subsubsection{\textbf{Interpretation of Loss Term}, \texorpdfstring{$\textbf{L}_T$}{}} The loss function term, $\textbf{L}_T$, is the multimodal triplet loss term which is described by Eq. \ref{eq:LT}. The multimodal triplet loss function contains two terms: intra-sensor triplet loss and inter-sensor triplet loss.\par
\begin{figure}
    \begin{centering}
    \includegraphics[scale=0.1]{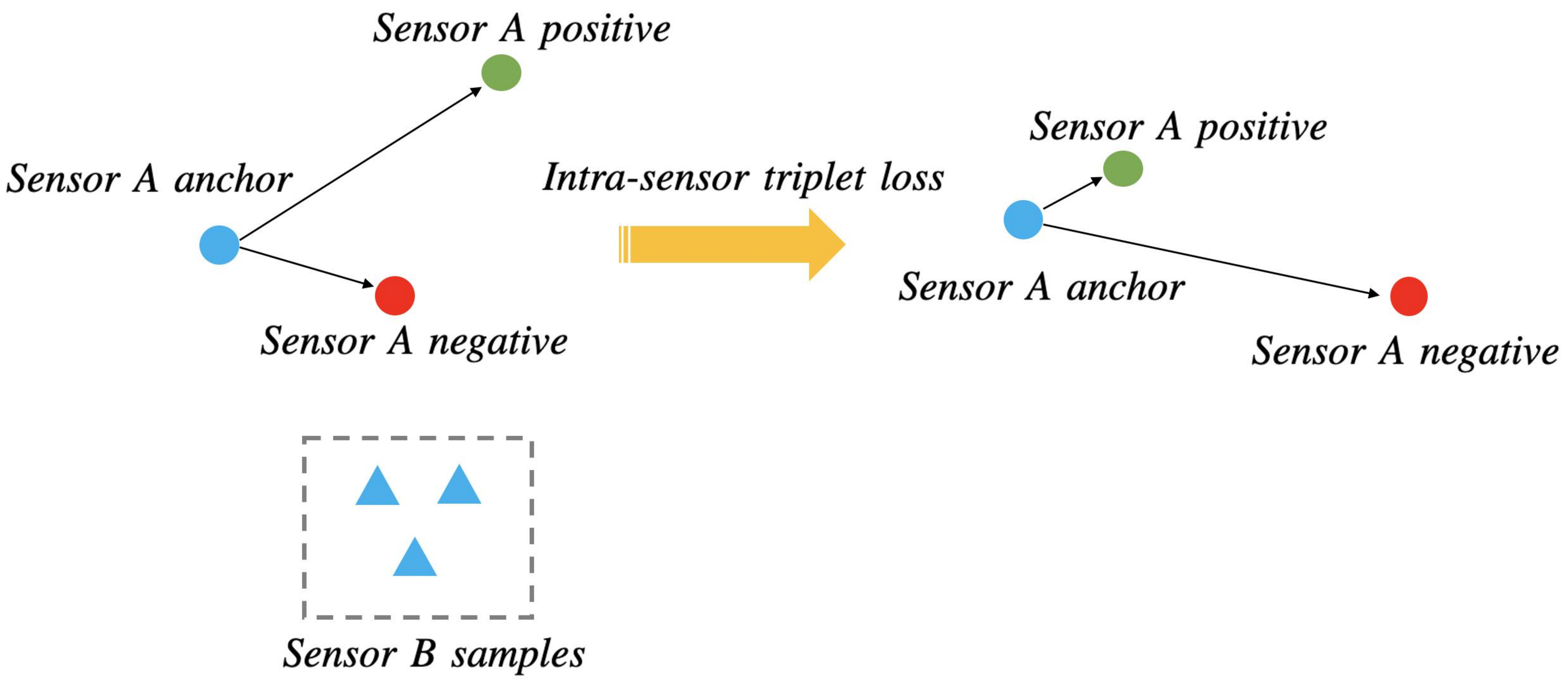}
  \caption{The effect of the intra-sensor triplet loss term is shown here. After applying this term, the network brings the anchor and positive embeddings from \textit{sensor} $A$ close to each other and pushes away the \textit{sensor} $A$ negative sample's embeddings.}
    \label{IntraSensor} 
    \end{centering}
\end{figure}

 
The intra-sensor triplet loss term is the standard triplet loss term to train a triplet network. This term results in discriminative embeddings of \textit{sensor} $A$. The effect of this term is demonstrated in Fig. \ref{IntraSensor}.\par
\begin{figure}
    \begin{centering}
    \includegraphics[scale=0.095]{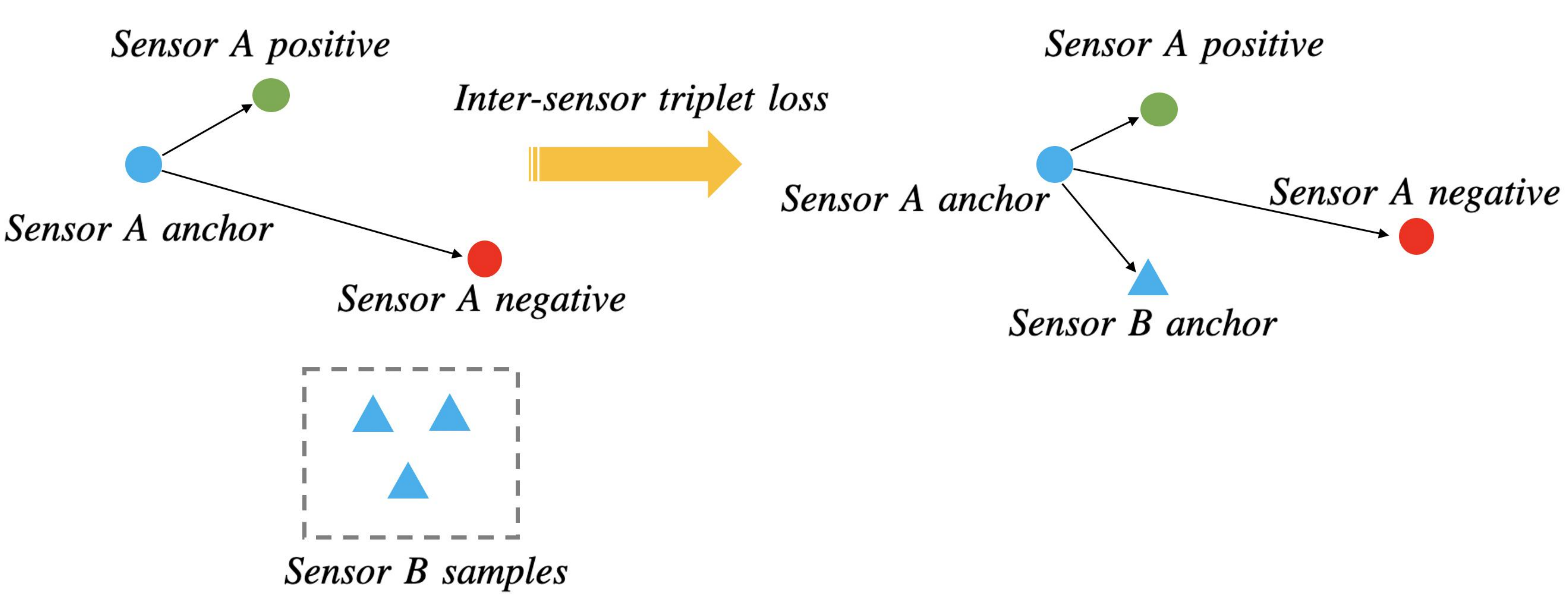}
  \caption{The effect of the inter-sensor triplet loss term is shown here. Before applying this term, only the \textit{sensor} $A$ embeddings are clustered. After applying this term, the \textit{sensor} $B$ anchor also moves close to the anchor of \textit{sensor} $A$.}
    \label{InterSensor} 
    \end{centering}
\end{figure}

 
The inter-sensor triplet loss term is a novel term that is introduced in this paper. This term maps the anchor of the \textit{sensor} $B$ close to the positive of \textit{sensor} $A$ and pushes away the negative of \textit{sensor} $A$. In other words, we are treating the anchor of \textit{sensor} $B$ similar to the anchor of \textit{sensor} $A$. This term is responsible for grouping the cross-modal embeddings in the shared latent space. The effect of the inter-sensor triplet loss term is demonstrated in Fig. \ref{InterSensor}.

\begin{equation}
    \label{eq:LT}
    \begin{aligned} 
    \textbf{L}_T = \boldsymbol{\sum}_{k=1} ^{K} \underbrace{\left\|\textbf{z}_{A,k}^{a} -\textbf{z}_{A,k}^{p} \right\|^{2} - \left\| \textbf{z}_{A,k}^{a} -\textbf{z}_{A,k}^{n} \right\|^{2}+ \mathit{\alpha} }_{\text{Intra-sensor triplet loss term}}  + \\
     \underbrace{\left\|\textbf{z}_{B,k}^{a} -\textbf{z}_{A,k}^{p} \right\|^{2} - \left\| \textbf{z}_{B,k}^{a} -\textbf{z}_{A,k}^{n} \right\|^{2}+ \mathit{\alpha}}_{\text{ Inter-sensor triplet loss term}}
    \end{aligned}
\end{equation}
\noindent where $\mathit{\alpha}$ is the margin hyperparameter, and there are $K$ random triplets selected from the dataset for training.

\subsubsection{\textbf{Interpretation of Loss Term}, \texorpdfstring{$\textbf{L}_E$}{}} This term consists of reconstruction loss terms of all the autoencoders. The three autoencoders are from the Triplet network for \textit{sensor} $A$. The fourth autoencoder is for \textit{sensor} $B$. The loss term, $\textbf{L}_E$ is described by Eq. \ref{Objective1}:

\begin{equation}
    \begin{aligned}
    \label{Objective1}
    \textbf{L}_E = \boldsymbol{\sum}_{k=1} ^{K} \left\|\textit{\textbf{S}}_{A,k}^{a} - \tilde{\textit{\textbf{S}}}_{A,k}^{a}\right\|^{2}+\left\|\textit{\textbf{S}}_{A,k}^{p} - \tilde{\textit{\textbf{S}}}_{A,k}^{p}\right\|^{2}+\\ \left\|\textit{\textbf{S}}_{A,k}^{n} - \tilde{\textit{\textbf{S}}}_{A,k}^{n}\right\|^{2}+ \left\|\textit{\textbf{S}}_{B,k}^{a} - \tilde{\textit{\textbf{S}}}_{B,k}^{a}\right\|^{2}
    \end{aligned}
\end{equation}

\noindent where $K$ is the number of triplets selected for training.

\subsubsection{\textbf{Interpretation of Loss Term}, \texorpdfstring{$\textbf{L}_{SE}$}{}} After using the loss term $\textbf{L}_T$, the embeddings of both the sensors are clustered in the shared manifold. However, the term $\textbf{L}_T$ is not sufficient to tightly cluster the embeddings of different sensors. The embeddings of the heterogeneous modalities are difficult to bring closer in the shared manifold because they have a different data structure and capture different kinds of information from the region of interest. Therefore, to enhance the clustering process, a similarity enhancement (SE) term is introduced in the loss function, which is shown in Eq. \ref{SETerm}.

\begin{equation}    
    \label{SETerm}  
    \textbf{L}_{SE} = \mathit{\gamma} \left\|\textbf{z}_{A}^{a} - \textbf{z}_{B}^{a}\right\|^{2}
\end{equation}

\noindent where $\mathit{\gamma}$ is the weight parameter which is assigned a small value between $0$ and $1$. In our experiments, we usually set the value of $\mathit{\gamma}$ less than $0.4$.

This term is weighted by a parameter $\gamma$. However, the first two terms, $\textbf{L}_T$ and $\textbf{L}_E$, are not weighted using a balancing parameter. The reason is that the term $\textbf{L}_T$ is primarily responsible for clustering the embeddings of all the classes irrespective of the sensors. The term $\textbf{L}_E$ is the reconstruction loss of all the autoencoders. Now, both $\textbf{L}_T$ and $\textbf{L}_E$ terms are given a weight of $1$ as both are equally important. However, the term $\textbf{L}_{SE}$ is simply enhancing the work done by the term $\textbf{L}_T$. It does so by reducing the distance between anchors of both the sensors, which results in more compact clusters, as shown in Fig. \ref{fig:ScatterSE} and \ref{fig:ScatterWSE}. We observed that using small values of $\gamma$ resulted in a significant improvement in clustering. Using larger values of $\gamma$ could result in a collapsed model. Therefore, assigning this term an equal weight as the other terms ($\textbf{L}_T$ and $\textbf{L}_E$) is not useful. That is why this is the only term to be assigned a weighting/ balancing parameter. \par

The experiments have been conducted with and without the $\textbf{L}_{SE}$ term in the loss function, and the comparison in performance is shown in Section \ref{sssec:se_effect}.

\subsection{Training Strategy}
A na\"{i}ve approach to train the CoMMANet is randomly selecting multiple triplets. Since this approach is computationally intensive, an offline semi-hard/ hard triplet selection strategy is used to train the network. According to Schroff et al. \cite{schroff2015facenet}, selecting hard triplets early on in training can lead to a collapsed model (i.e. $f\left(x\right) = 0$). The hard triplets mining strategy is also not robust to outliers. Therefore, the semi-hard triplet strategy is primarily used for the experiments (See Section \ref{sssec:triplet_effect} for ablation study). Providing the network with a mixture of easy and semi-hard triplets makes the training process more stable, which results in a better performance. We developed a strategy for selecting the semi-hard/ hard multimodal triplets (which is a variation of the standard hard triplet mining strategy). \par 

The multimodal triplet mining is divided into two stages: intra-sensor and inter-sensor triplets selection.

\begin{enumerate}
    \item In the first stage, the semi-hard/ hard triplets from the \textit{sensor} $A$ are sampled.
    \item In the second stage, the anchor of \textit{sensor} $B$ is treated as the anchor of \textit{sensor} $A$, and again the semi-hard/ hard triplets are sampled. So, essentially, now the anchor is from \textit{sensor} $B$, and the positive and negative are from \textit{sensor} $A$. 
\end{enumerate}

Note that the triplet selection process requires latent space embeddings and not the original sensor data.\par

Let $\textbf{S}_{A}^{a}$, $\textbf{S}_{A}^{p}$, and $\textbf{S}_{A}^{n}$ denote the anchor, positive, and negative samples from \textit{sensor} $A$. $\textbf{S}_{B}^{a}$ is the anchor from the \textit{sensor} $B$. Let $\textbf{z}_{A}$ and $\textbf{z}_{B}$ denote the latent space embeddings of \textit{sensor} $A$ and \textit{sensor} $B$, respectively. The pseudo-code of the algorithm is presented in Algorithm \ref{alg:Algo}.

\begin{algorithm}
 \caption{Multimodal Offline Hard Triplet Mining}
 \label{alg:Algo}
 \begin{algorithmic}[1]
 \renewcommand{\algorithmicrequire}{\textbf{Input:}}
 \renewcommand{\algorithmicensure}{\textbf{Output:}}
 \REQUIRE $\textbf{S}_{A}$, $\textbf{S}_{B}$
 \ENSURE  $\textbf{S}_{A}^{a}$, $\textbf{S}_{A}^{p}$, $\textbf{S}_{A}^{n}$, $\textbf{S}_{B}^{a}$
 \\ \textit{Initialization} : Checkpoint counter initialized
  \STATE i = 1
  \STATE Select $K$ random triplets. Train the model for a specific number of epochs and save the model as Checkpoint $1$.
  \FOR {$i = 2$ to $N$}
  \STATE Predict the embeddings of each sensor: $\textbf{z}_{A}$, $\textbf{z}_{B}$, using the Checkpoint $\left(i-1\right)$ model.
  \STATE Select $K$ anchor, positive, and negative from \textit{sensor} $A$ in such a way that: $d\left(\textbf{z}_{A}^{a}, \textbf{z}_{A}^{n} \right) < d\left(\textbf{z}_{A}^{a}, \textbf{z}_{A}^{p} \right)$, where $d$ is a distance metric. The corresponding \textit{sensor} $B$ anchor samples are selected randomly.
  \STATE Select $K$ anchor $\left(\textit{sensor}\:B\right)$, positive $\left(\textit{sensor}\:A\right)$, and negative $\left(\textit{sensor}\:A\right)$ in such a way that: $d\left(\textbf{z}_{B}^{a}, \textbf{z}_{A}^{n} \right) < d\left(\textbf{z}_{B}^{a}, \textbf{z}_{A}^{p} \right)$, where $d$ is a distance metric. The corresponding \textit{sensor} $A$ anchor samples are selected randomly.

 \ENDFOR
 \RETURN $\textbf{S}_{A}^{a}$, $\textbf{S}_{A}^{p}$, $\textbf{S}_{A}^{n}$, $\textbf{S}_{B}^{a}$
 \end{algorithmic} 
\end{algorithm}

\begin{figure*}[!htb]
  \begin{centering}
  \includegraphics[scale=0.2]{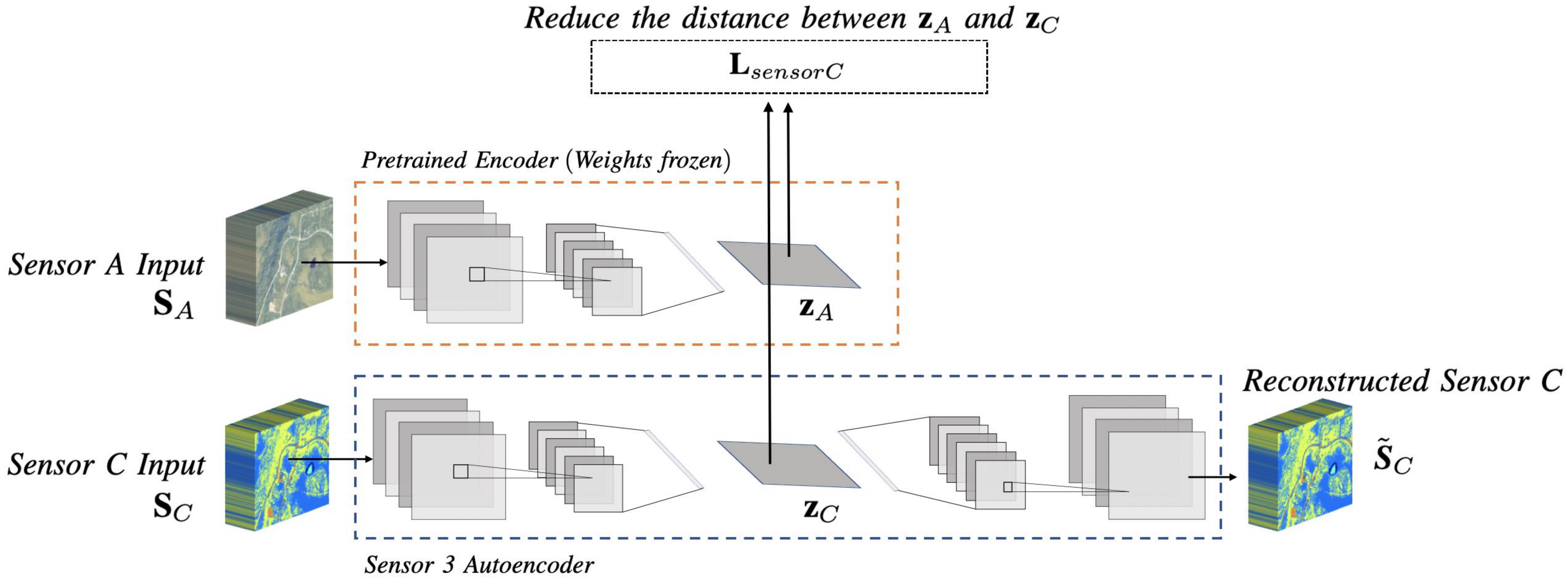}
  \caption{The architecture to map the additional sensors onto the shared manifold is shown here. The embeddings of a new sensor are brought closer to the pretrained encoder's embeddings by reducing the distance between them.}
  \label{Sensor3Model} 
  \end{centering}
\end{figure*}

\subsection{Learning Mapping of Additional Sensors}
The current architecture allows only two sensors to be mapped onto a shared manifold. However, there can be more than one sensor surveying the region of interest, for example, \textit{LiDAR}, \textit{SAR}, and \textit{HSI}. In such cases, firstly, two sensors are mapped onto the shared manifold. Then, the rest of the sensors are mapped using the following algorithm (architecture shown in Fig. \ref{Sensor3Model}):

\begin{enumerate}
    \item Use the pretrained encoder from any of the first 2 sensors. Let's say, the \textit{sensor} $A$ pretrained encoder is used. According to Eq. \ref{eq1}, for an input, $\textbf{S}_{A}$, the embeddings, $\textbf{z}_{A} \in {\rm I\!R}^{D}$, can be expressed as: $\textbf{e}_{A}\left(\textbf{S}_{A}\right)$, where $D$ is the dimensionality of the shared latent space. 
    \item Develop a standard autoencoder for \textit{sensor} $C$.  
    \item The latent space of \textit{sensor} $C$ is forced to be similar to the pretrained encoder’s latent space.
    For a given input, $\textbf{S}_{C}$ $\in {\rm I\!R}^{N_3}$, where $N_3$ is the dimensionality of data extracted from \textit{sensor} $C$, the encoder of \textit{sensor} $C$ is represented by the function $\textbf{e}_{C}\left(.\right)$ and the decoder is represented by the function $\textbf{d}_{C}\left(.\right)$. The latent space, $\textbf{z}_{C}$, is denoted by $\textbf{e}_{C}\left(\textbf{S}_{C}\right)$, where $\textbf{z}_{C} \in {\rm I\!R}^{D}$, and $D$ is the dimensionality of the shared latent space. The reconstructed output of the \textit{sensor} $C$, $\tilde{\textbf{S}}_{C}$, can be expressed as: $\textbf{d}_{C}\left(\textbf{z}_{C}\right)$, where $\tilde{\textbf{S}_{C}} \in {\rm I\!R}^{N_3}$.
    \item The model is trained using the following objective function:
    \begin{equation}
        \textbf{L}_{sensorC} = \boldsymbol{\sum}_{k=1} ^{K} \left\|\textbf{z}_{A,k} -\textbf{z}_{C,k} \right\|^{2} +\newline
        \underbrace{\left\|\textit{\textbf{S}}_{C,k} - \tilde{\textit{\textbf{S}}}_{C,k}\right\|^{2}}_{\textrm{Reconstruction term}}
    \end{equation}
        
    \noindent where $K$ is the number of training samples. 
    \item After training the network using the above-mentioned objective function, the \textit{sensor} $C$ embeddings are mapped close to the \textit{sensor} $A$ embeddings. All three sensors now have embeddings in the shared manifold in a discriminative manner.

\end{enumerate}

\subsection{Classification Strategy}
\begin{figure}[!htb]
  \begin{centering}
  \includegraphics[scale=0.18]{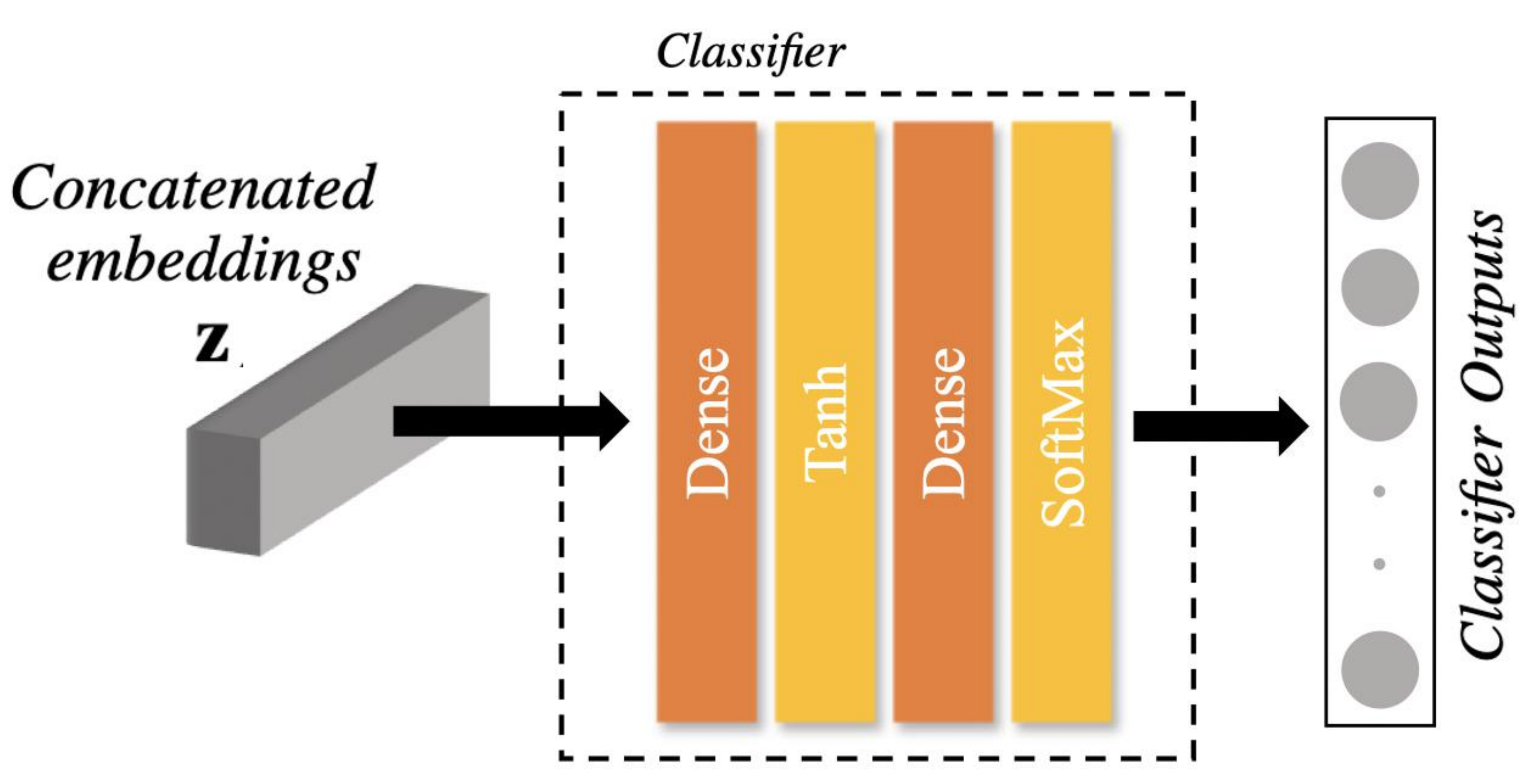}
  \caption{Shared embeddings classification model. The term ``Dense" signifies a Fully Connected layer whose neurons are connected to every neuron in the previous layer.}
  \label{Classifier} 
  \end{centering}
\end{figure}

For classification, the embeddings of different modalities are concatenated, and a shallow fully connected neural network is applied. The classifier architecture is shown in Fig. \ref{Classifier}. Alternatively, the nearest neighbor classifier is also used to classify the concatenated embeddings. Since the embeddings are already clustered in the latent space, the nearest-neighbors classifier also gives a good performance which is comparable to the neural network classifier results.\par

To compute the nearest neighbors, the distance of a test sample is computed with every training sample. All the distances are averaged by the class. Now, the label of the class having the minimum distance is assigned to the test sample. It is not necessary to use all the training samples, and k-nearest neighbors can also be used. \par

To develop a unified classification model, instead of using the concatenated embeddings, the classification can be performed using the latent space of any of the sensors. Since the embeddings of all the sensors lie close to each other in the shared latent space, a classifier trained on one sensor can be used for other sensors as well. However, the classifier trained on the concatenated embeddings of all the sensors yields more accurate and robust results. Let $\textbf{z}$ denote the flattened latent space of a single sensor. Let the hidden state $\mathbf{h_1}$ contain $\mathbf{U}$ hidden units (neurons). We use the convention that the addition of a matrix, $A \in {\rm I\!R}^{N \times M}$, and a vector, $B \in {\rm I\!R}^{1 \times M}$, means that the vector $B$ is added to every row of matrix $A$. The classifier output is given by:

\begin{equation}
    \textbf{h}_{1} = tanh\left(\textbf{z} \textbf{W}_{1} + \textbf{b}_{1} \right)
\end{equation}

\noindent where $\mathbf{z} \in {\rm I\!R}^{N \times D}$, $D$ is the dimensionality of the latent space, $N$ is the number of samples, $\mathbf{W_1} \in {\rm I\!R}^{D \times U}$ is the hidden state weight matrix, and $\mathbf{b_1} \in {\rm I\!R}^{1 \times U}$ is the bias. Thus, $\mathbf{h_1} \in {\rm I\!R}^{N \times U}$.

Let $\sigma$ be the softmax function applied on the final layer, and $\mathbf{O}$ be the number of units (neurons) in the output layer.

\begin{equation}
    \textbf{\^{y}} = \sigma\left(\textbf{h}_{1} \textbf{W}_{2} + \textbf{b}_2\right)
\end{equation}

\noindent where $\mathbf{W_2} \in {\rm I\!R}^{U \times O}$ is the hidden state weight matrix, $\mathbf{b_2} \in {\rm I\!R}^{1 \times O}$ is the bias, and $\mathbf{h_1} \in {\rm I\!R}^{N \times U}$. Thus, $\textbf{\^{y}} \in {\rm I\!R}^{N \times O}$ is the classifier output, and $N$ is the number of samples.

If the latent spaces of both \textit{sensor} $A$ and \textit{sensor} $B$ are fused to perform classification, the input of the classifier will be the concatenated flattened latent space, $\mathbf{z}_{fused}$, of both the sensors, where $\mathbf{z}_{fused} \in {\rm I\!R}^{N \times (2*D)}$, $D$ is the dimensionality of the latent space of each sensor, and $N$ is the number of samples.

\subsection{Missing Sensor's Embeddings Prediction/ Sensor Translation}
\begin{figure}[!htb]
  \begin{centering}
  \includegraphics[scale=0.18]{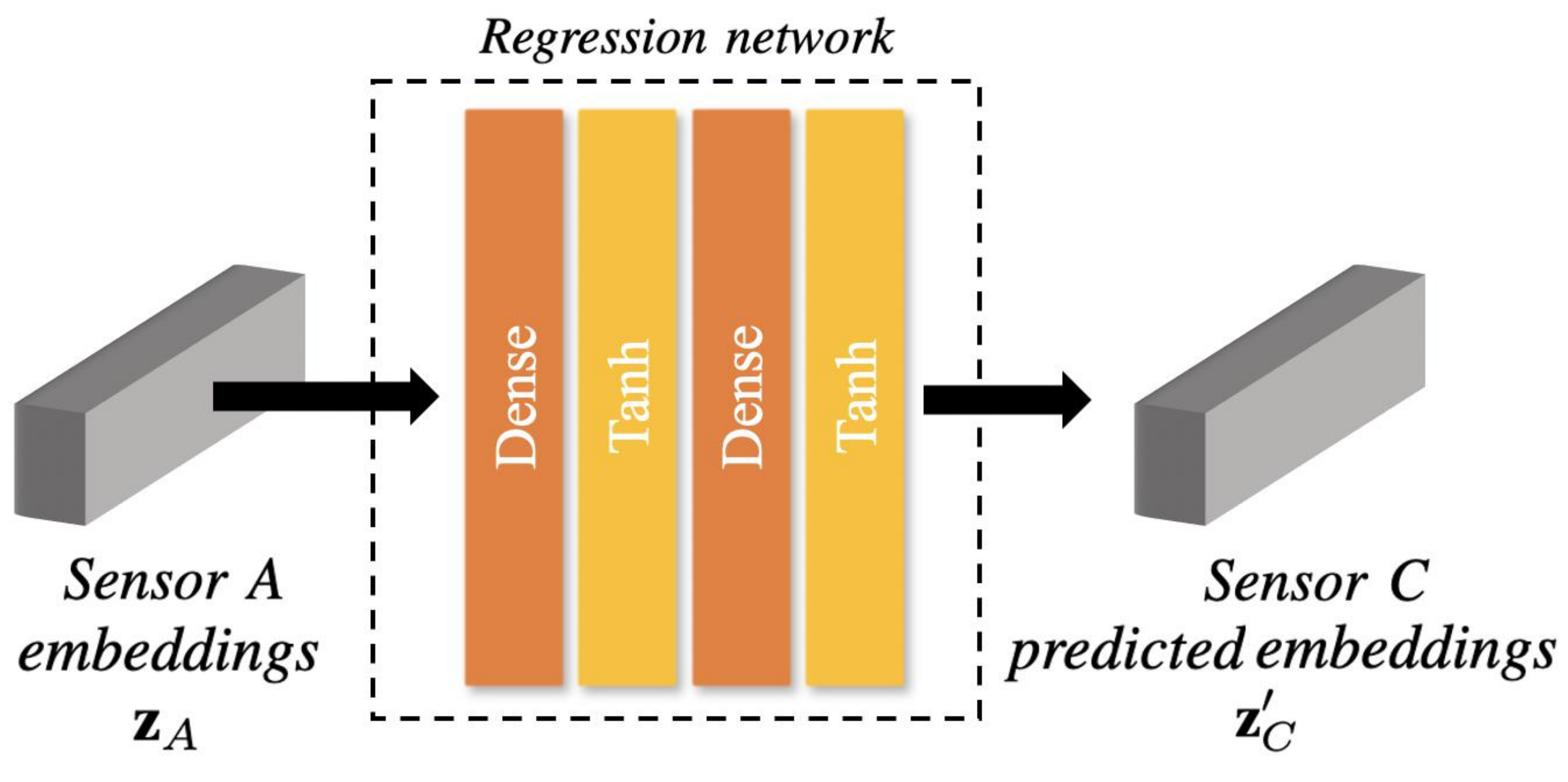}
  \caption{Missing sensor's embeddings prediction/ Sensor translation model. The term ``Dense" signifies a Fully Connected layer whose neurons are connected to every neuron in the previous layer.}
  \label{Regressor} 
  \end{centering}
\end{figure}

After following the steps in the subsection $A$, the embeddings of all the sensors are successfully mapped onto the shared manifold. Let $\textbf{S}_{A}$ and $\textbf{S}_{C}$ be the two sensors, where $\textbf{S}_{C}$ is the missing sensor. The clustered embeddings of both the sensors will lie close to each other in the shared manifold. In an ideal scenario, the sensors would be homogeneous, and their embeddings will completely overlap. However, if the sensors are heterogeneous, the embeddings will not overlap entirely due to the variation in data structure and information captured by different sensors. Therefore, a shallow regression network is developed, which predicts the embeddings of the missing sensor using the available sensor. The architecture of the regression neural network is shown in Fig. \ref{Regressor}. \par

Let $\textbf{z}_{A}$ be the flattened latent space of \textit{sensor} $A$ (the available/ predictor sensor). Here also, we use the convention that the addition of a matrix, $A \in {\rm I\!R}^{N \times M}$, and a vector, $B \in {\rm I\!R}^{1 \times M}$, means that the vector $B$ is added to every row of matrix $A$. The predicted latent space of \textit{sensor} $C$ (the missing sensor), $\textbf{z}'_{C}$, is given by:

\begin{equation}
    \textbf{h}_{3} = tanh\left(\textbf{z}_{A} \textbf{W}_{3} + \textbf{b}_{3} \right)
\end{equation}

\noindent where $\mathbf{z}_A \in {\rm I\!R}^{N \times D}$, $D$ is the dimensionality of the latent space, and $N$ is the number of samples. $\mathbf{W_3} \in {\rm I\!R}^{D \times V}$ is the hidden state weight matrix, $V$ is the number of units (neurons) in the hidden layer $\mathbf{h}_3$, and $\mathbf{b_3} \in {\rm I\!R}^{1 \times V}$ is the bias. Thus, $\mathbf{h_3} \in {\rm I\!R}^{N \times V}$.

\begin{equation}
    \textbf{z}'_{C} = tanh\left(\textbf{h}_{3} \textbf{W}_{4} + \textbf{b}_4\right)
\end{equation}


\noindent where $\mathbf{W_4} \in {\rm I\!R}^{V \times D}$ is the hidden state weight matrix, $D$ is the number of units (neurons) in the output layer $\mathbf{z}'_C$, and $N$ is the number of samples. $\mathbf{h_3} \in {\rm I\!R}^{N \times V}$ is the previous layer output, and $\mathbf{b_4} \in {\rm I\!R}^{1 \times D}$ is the bias. Thus, $\mathbf{z}'_C \in {\rm I\!R}^{N \times D}$ is the predicted latent space of \textit{sensor C}.

After predicting the missing sensor's latent space, its decoder can be used to reconstruct the original data. This way sensor translation can be performed. Here, the predicted latent space of \textit{sensor} $C$ can be reconstructed using its decoder, $\textbf{d}_{C}\left(.\right)$. The reconstructed \textit{sensor} $C$ data, $\tilde{\textbf{S}}_{C}$, can be expressed as:

\begin{equation}
    \tilde{\textbf{S}}_{C} = \textbf{d}_{C}\left( \textbf{z}'_{C} \right)
\end{equation}
\noindent where $\tilde{\textbf{S}}_{C} \in {\rm I\!R}^{N \times N_3}$, $N$ is the number of samples, and $N_3$ is the dimensionality of  data extracted from sensor $C$.

In our experiments, the latent space values are between $-1$ and $1$, therefore, a \textit{tanh} activation is used. However, any other activation function can also be used according to convenience.

\section{Experiments}
\label{sec:Experiments}
\subsection{Datasets Description}

\begin{enumerate}
    \item \textit{AVIRIS-NG and NEON Data:} For initial experiments, the data is used from AVIRIS-NG and NEON sensors. The HSI data is acquired from NEON and AVIRIS-NG sensors. The LiDAR data is available only from NEON. The data was acquired over Healy, Alaska, in 2017. The NEON HSI data, which is captured by the NEON Airborne Observation Platform (AOP) Imaging Spectrometer (NIS), comprises $426$ bands, each with a spectral resolution of $~5$ nm covering the range from $380$ $\mu$m to $2510$ $\mu$m. The LiDAR data was acquired by the Optech Gemini sensor. The data is composed of $1000$ x $1000$ pixels with a spatial resolution of $1$ m. \par 
    The AVIRIS-NG sensor comprises $425$ bands covering the range from $380$ $\mu$m to $2510$ $\mu$m with a spectral resolution of $5$ nm. The spatial resolution of AVIRIS-NG data is $5$ m. QGIS software was used to extract the common region of interest (ROI) between NEON and AVIRIS-NG data. The original AVIRIS-NG data is composed of $213$ x $213$ pixels, which was upscaled to $1000$ x $1000$ pixels. The RGB image of ROI is shown in Fig. \ref{NEON_ROI}. The NEON Canopy Height Model was used to generate the ground truth for trees. The rest of the data were manually labeled. For the experiments, four classes are used: mixed ground, road, building, and trees. The distribution of the classes in the dataset is described in Table \ref{table:neon_dist}.

    \begin{figure}[!htb]
      \begin{centering}
      \includegraphics[scale=0.15]{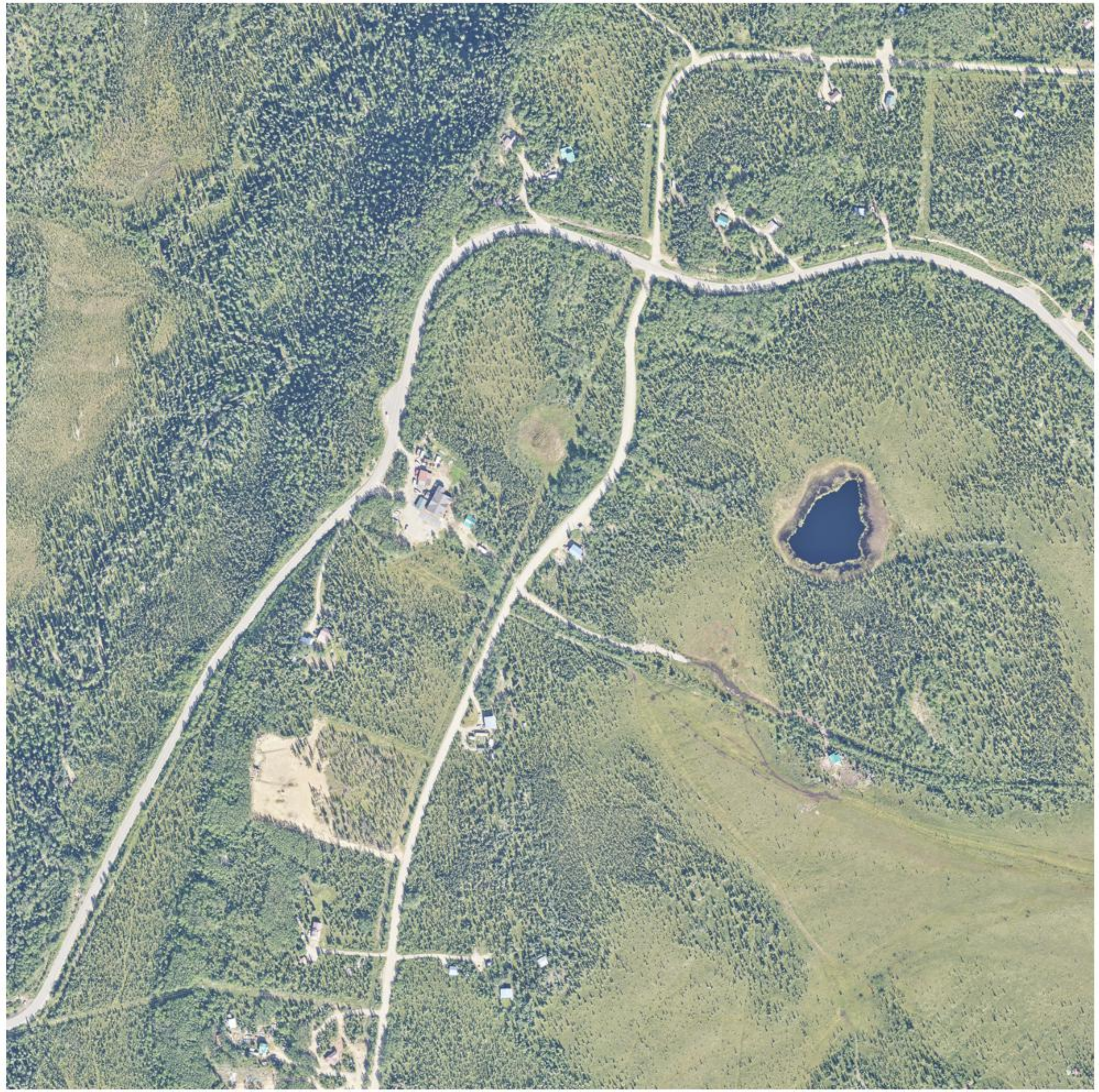}
      \caption{The RGB image of the region of interest used from the AVIRIS-NG and NEON data}
      \label{NEON_ROI} 
      \end{centering}
    \end{figure}

    \item \textit{MUUFL Gulfport Data:} The Gulfport dataset was acquired in November 2010, over the University of Southern Mississippi Gulfport Campus, Long Beach, Mississippi, USA. The dataset is composed of coregistered HSI and LiDAR-based digital surface model (DSM). The HSI data was acquired by the Compact Airborne Spectrographic Imager (CASI)-1500 sensor and the LiDAR data is acquired by the Gemini airborne laser terrain mapper (ALTM) LiDAR sensor. The data consists of $325$ × $220$ pixels with $64$ spectral channels covering the range from $375$ nm to $1050$ nm at a spectral sampling of $10$ nm. The spatial resolution of HSI is $0.54$ m across the track and $1.0$ m along the track. The spatial resolution of LiDAR data is $0.60$ m across the track and $0.78$ m along the track. In this dataset, 11 classes are investigated for the land cover classification task. The distribution of the classes in the dataset is described in Table \ref{table:muufl_dist}.

    \item \textit{HS-SAR Berlin data:} This dataset describes the Berlin urban and its neighboring rural area. It consists of an EnMAP HS image which is simulated from the HyMap HS data. A corresponding Sentinel-1 SAR data of the same region was prepared using the ESA toolbox SNAP after applying orbit profile, radiometric calibration, speckle reduction, and terrain correction. The HSI data consists of $797$ × $220$ pixels with $244$ spectral bands covering the range from $400$ nm to $2500$ nm. The SAR data are a dual-polarSAR containing four bands. The processed SAR image consists of $1723$ x $476$ pixels and has a $13.89$ m GSD. A nearest-neighbor interpolation is performed on the HSI image to make SAR and HSI images of the same size. The ground truth map is created using OpenStreetMap data. This dataset contains eight categories for the classification task. The distribution of the classes in the dataset is described in Table \ref{table:berlin_dist}.

    
\end{enumerate}

\begin{table}[!htb]
\caption{\textsc{Description of the AVIRIS-NG/ NEON data}}

\centering
 \begin{tabular}{||c | c | c||} 
 \hline\hline
 No. & Class & Number of Samples \\ [0.5ex] 
 \hline\hline
 1 & Mixed Grass & 46324 \\
 2 & Buildings & 22538 \\
 3 & Trees & 19054 \\
 4 & Road/ Ground & 1300 \\
 \hline
  & \textbf{Total} & \textbf{89216} \\
 \hline\hline

 \end{tabular}
\label{table:neon_dist}
\end{table}

\begin{table}[!htb]
\caption{\textsc{Description of the MUUFL data}}

\centering
 \begin{tabular}{||c | c | c||} 
 \hline\hline
 No. & Class & Number of Samples \\ [0.5ex] 
 \hline\hline
 1 & Trees & 23246 \\
 2 & Mostly Grass & 4270 \\
 3 & Mixed ground surface & 6882 \\
 5 & Road & 6687 \\
 6 & Water & 466 \\
 7 & Building shadow & 2233 \\
 8 & Building & 6240 \\
 9 & Sidewalk & 1385 \\
 10 & Yellow curb & 183 \\
 11 & Cloth panels & 269 \\
 \hline
  & \textbf{Total} & \textbf{53687} \\
 \hline\hline

 \end{tabular}
\label{table:muufl_dist}
\end{table}

\begin{table}[!htb]
\caption{\textsc{Description of the HS-SAR Berlin Data}}

\centering
 \begin{tabular}{||c | c | c | c||} 
 \hline\hline
 No. & Class & Training Sample & Testing Samples \\ [0.5ex] 
 \hline\hline
 1 & Forest & 443 & 54511 \\
 2 & Residential Area & 423 & 268219 \\
 3 & Industrial Area & 499 & 19067 \\
 4 & Low Plants & 376 & 58906 \\
 5 & Soil & 331 & 17095 \\
 6 & Allotment & 280 & 13025 \\
 7 & Commercial Area & 298 & 24526 \\
 8 & Water & 170 & 6502 \\
 \hline
  & \textbf{Total} & \textbf{2820} & \textbf{461851} \\
 \hline\hline

 \end{tabular}
\label{table:berlin_dist}
\end{table}

\subsection{Experimental Setup}
To measure the classification performance, Overall Accuracy (OA), Average Accuracy (AA), and Kappa coefficient are used. In the missing sensor scenario, the mean squared error metric is used to evaluate the performance of predicted latent space embeddings and reconstructed sensor data. 

\begin{enumerate}
    \item \textbf{\textit{AVIRIS-NG/ NEON Data}} \\
    \textbf{\textit{Shared embeddings:}} The CoMMANet is applied on AVIRIS-NG and NEON HSI data. 40\% of the data from each class is used for training and the rest 60\% for testing. From HSI data, a single pixel is used for training. The HSI data from both sensors lies between $0$ and $1$. The weight for the similarity enhancement term, $\gamma$, is set to $0.4$. The margin, $\alpha$, is set to 1. A \textit{tanh} activation is used on the latent space so that the embeddings remain bounded. The embeddings are $32$-dimensional vectors. To find the optimal latent space size, experiments are conducted by setting the latent space dimensionality to 16, 32, 64, and 128. The best performance is achieved when the latent space is 32-dimensional. Therefore, 32-dimensional latent space is used to generate shared embeddings.\par For training the triplet network, the semi-hard triplets mixed with a few easy triplets are used instead of hard triplets because hard triplets are sensitive to outliers. The Similarity Enhancement (SE) term is also added to the loss function. The CoMMANet is trained for $10$ checkpoints for $450$ epochs. At the beginning of each checkpoint, new triplets are computed to focus on the samples showing a higher loss. In each checkpoint, $100,000$ triplets are used. The learning rate is set to $0.0005$, the batch size is set to $256$, and the model is trained using the Adam optimizer. \par 

    \textbf{\textit{Classification:}} The CoMMANet embeddings of NEON HSI and AVIRIS-NG HSI are concatenated and used for classification. The classification is performed using a single neural network having two hidden layers with $128$ and $64$ hidden units, respectively. The classification results are shown in Table \ref{table:neon_result}. A K-fold Monte Carlo experiment is conducted, and results are reported along with the mean and standard deviation of accuracy.\par
    Additionally, to demonstrate the effectiveness of a unified classification model, a classifier is trained on the embeddings of one sensor and tested on the embeddings of other sensors. Since all the embeddings are clustered in the shared manifold, the models trained on one sensor show comparable accuracy on other sensors as well. The results are shown in Table \ref{table:neon_unified}.

    \begin{table}[!htb]
    \caption{\textsc{Classification performance of the CoMMANet embeddings on the AVIRIS-NG/ NEON Data is shown.}}

    \centering
     \begin{tabular}{||c | c | c||} 
     \hline\hline
      No. & Class & NEON HSI + AVIRIS-NG HSI\\ [0.5ex] 
      \cline{3-3}
      &  & Neural Network \\ [0.5ex] 


     \hline\hline
     1 & Mixed Grass & $98.3\pm0.05$ \\
     2 & Buildings & $98.1\pm0.07$ \\
     3 & Trees & $99.2\pm0.02$ \\
     4 & Road/ Ground & $98.0\pm0.05$ \\
     \hline
     \multicolumn{2}{||c|}{OA (\%)} & $\textbf{98.3}\pm\textbf{0.02}$ \\
      \multicolumn{2}{||c|}{AA (\%)} & $\textbf{98.7}\pm\textbf{0.05}$ \\
      \multicolumn{2}{||c|}{Kappa (\%)} & $\textbf{98.9}\pm\textbf{0.02}$ \\
     \hline\hline
     \end{tabular}
    \label{table:neon_result}
    \end{table}

    \begin{table*}[!htb]
    \caption{\textsc{AVIRIS-NG/ NEON Data: The effectiveness of a unified classification model is shown here. The classifier is trained on embeddings of one sensor and tested on embeddings of another sensor.}}

    \centering
     \begin{tabular}{||c | c | c | c | c ||} 
     \hline\hline
     Neural Network Classifier \textit{trained} & Evaluation Metric & \multicolumn{3}{c||}{Neural Network Classifier \textit{tested} on embeddings of}\\
     on embeddings of & (\%) & \multicolumn{3}{c||}{}
     \\\cline{3-5}
     & & NEON HSI & AVIRIS-NG HSI & LiDAR\\ [0.5ex]
     \hline
      & OA & $98.6\pm0.06$ & $98.5\pm0.05$ & $93.16\pm0.12$
     \\\cline{2-5}
     NEON HSI & AA & $98.4\pm0.12$ & $95.6\pm0.15$ & $93.1\pm0.12$
     \\\cline{2-5}
      & Kappa & $97.7\pm0.13$ & $97.8\pm0.02$ & $95.6\pm0.08$ \\

     \hline
      & OA & $95.2\pm0.08$ & $98.5\pm0.05$ & $96.2\pm0.10$
     \\\cline{2-5}
     AVIRIS-NG HSI & AA & $94.6\pm0.10$ & $95.5\pm0.11$ & $90.5\pm0.15$
     \\\cline{2-5}
      & Kappa & $92.3\pm0.12$ & $97.4\pm0.04$  & $93.8\pm0.12$ \\

     \hline
      & OA & $97.5\pm0.10$ & $97.2\pm0.03$ & $98.4\pm0.04$
     \\\cline{2-5}
     LiDAR & AA & $97.1\pm0.20$ & $94.4\pm0.16$ & $95.7\pm0.10$
     \\\cline{2-5}
      & Kappa & $95.9\pm0.05$ & $95.5\pm0.10$  & $97.4\pm0.06$ \\

     \hline\hline
    
    \end{tabular}
    \label{table:neon_unified}
    \end{table*}

    \textbf{\textit{Mapping additional sensor (LiDAR):}} To map an additional sensor (NEON LiDAR) onto the shared manifold, the pretrained encoder of NEON HSI is used. The size of the input patches from LiDAR is $5$ x $5$ pixels. Each channel of LiDAR data is scaled between $0$ and $1$. The model is trained for $300$ epochs with a learning rate of $0.0005$ and a batch size of $256$. However, the model converges very fast (in approx. 200 epochs). The shared embeddings of NEON HSI, AVIRIS-NG HSI, and NEON LiDAR are shown in Fig. \ref{fig:ScatterSE}. \par 
    
    \textbf{\textit{Missing sensor prediction/ reconstruction:}} To simulate a missing sensor scenario, one sensor's embeddings are predicted from another sensor's embeddings. The values of the $\alpha$ and $\gamma$ parameters are the same as used for classification. A shallow neural network with two hidden layers is used to predict the embeddings. The two hidden layers contain 128 and 64 hidden units, respectively. A \textit{tanh} activation is used on the final layer. The batch size is set to $64$. The model is trained using 5-fold validation for $50$ epochs in each fold using the Adam optimizer. The prediction results are shown in Table \ref{table:neon_missing}. After the embeddings of a sensor are predicted, the decoder is used to reconstruct the original data.
    
    \begin{table}[!htb]
    \caption{\textsc{AVIRIS-NG/ NEON Data: The latent space of one sensor is predicted using another sensor and then reconstructed using the sensor's decoder.}}
    
    \centering
     \begin{tabular}{||c | c | c | c||} 
     \hline\hline
      &  & \multirow{1}{*}{Latent} & \multirow{1}{*}{Reconstructed} \\ [0.5ex] 
      Sensor & Sensor & Space & Data MSE \\ [0.5ex] 
      (Predictor) &(Predicted)  & MSE & ($\pm0.0001$)\\ [0.5ex] 
      &  & ($\pm0.0005$) & \\ [0.5ex] 

     \hline\hline
     NEON HSI & LiDAR & $0.042$ & $0.008$ \\
     \hline
     NEON HSI & AVIRIS-NG HSI & $0.048$ & $0.002$ \\
     \hline
     AVIRIS-NG HSI & LiDAR & $0.056$ & $0.008$ \\
     \hline
     AVIRIS-NG HSI & NEON HSI & $0.038$ & $0.001$ \\
     \hline
     LiDAR & AVIRIS-NG HSI & $0.180$ & $0.012$ \\
     \hline
     LiDAR & NEON HSI & $0.157$ & $0.010$ \\
     \hline\hline
     \multicolumn{4}{|c|}{}  \\
     \multicolumn{4}{|c|}{* \textit{Latent space values $\in$ [-1,1]} \qquad \textit{All sensors data $\in$ [0,1]}}  \\
     \hline
    
     \end{tabular}
    \label{table:neon_missing}
    \end{table}

    \item \textbf{\textit{MUUFL Gulfport Data}} \\
    \textbf{\textit{Shared embeddings:}} In this dataset also, 40\% of the data from each class is used for training and the rest 60\% for testing. For LiDAR, the input patch size of $13$ x $13$ pixels is found to be optimal. Similar to the previous dataset, a single pixel is used from HSI. The HSI data already exists between $0$ and $1$. The LiDAR height and intensity are scaled between $0$ and $1$. The optimal weight for the similarity enhancement term, $\gamma$, is $0$, and the margin, $\alpha$, is set to $0.4$. The value of $\alpha$ depends on the number of classes and the activation function used on the latent space. If the number of classes is high, then a higher value of $\alpha$ will lead to a collapsed model (i.e. $f\left(x\right) = 0$). The embeddings of each sensor are $32$-dimensional vectors, and a \textit{tanh} activation is applied on the latent space so that the embeddings remain bounded between $-1$ and $1$. The semi-hard triplets mixed with a few easy triplets are used for training the triplet network. The Similarity Enhancement (SE) term is also added to the loss function. The checkpoints are created in a similar way as the previous dataset. The CoMMANet is trained for $10$ checkpoints with $50$ epochs in each checkpoint. In each checkpoint, $800,000$ triplets are used. The learning rate is set to $0.001$, the batch size is set to $1024$, and the model is trained using the Adam optimizer.\par 
    
    \textbf{\textit{Classification:}} The CoMMANet embeddings of both sensors are concatenated and classified using an ensemble of three shallow neural networks which yielded an overall classification accuracy of $93.1\% \pm 0.15$. The best average classification accuracy is $94.0\% \pm 0.10$ which is achieved using KNN (with $k=35$). However, the best overall accuracy is $94.3\% \pm 0.12$, which is achieved using an ensemble of KNN (with $k = 35$) and three neural networks (initialized with different weights). The sensitivity of the KNN model to the value of $k$ is shown in Fig. \ref{fig:MUUFL_KNN}. Similar to the previous dataset, a K-fold Monte Carlo experiment is conducted here, and the mean and standard deviation of accuracy are reported in Table \ref{table:muufl_result}. The classification map and ground truth are shown in Fig. \ref{fig:MapMUUFL}. \par 
    For this dataset also, unified classification experiments are conducted, and the results are shown in Table \ref{table:muufl_unified}. The classifier trained on HSI embeddings show a good accuracy on LiDAR embeddings as well and vice-versa.

     \begin{figure}[!hb]
        \centering
        \includegraphics[width=7.8cm]{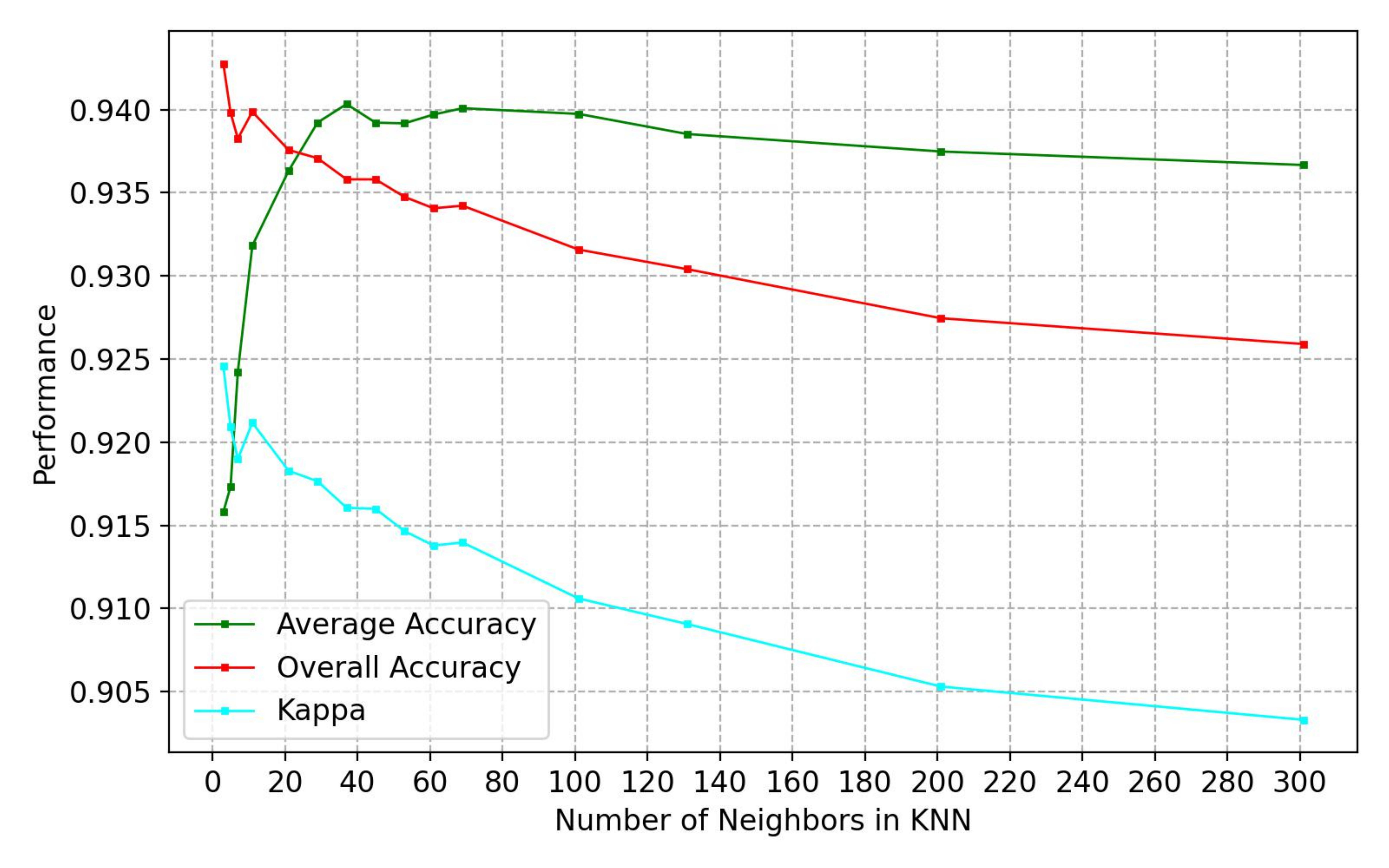}
        \caption{The sensitivity of the KNN model to the value of $k$ is shown for the MUUFL dataset. The KNN is used for the classification of concatenated HSI and LiDAR embeddings from CoMMANet.}
        \label{fig:MUUFL_KNN}
    \end{figure}

    \begin{figure*}[!htb]
        \centering
        \subfloat[\footnotesize{RGB Image}]{{\includegraphics[width=3.9cm]{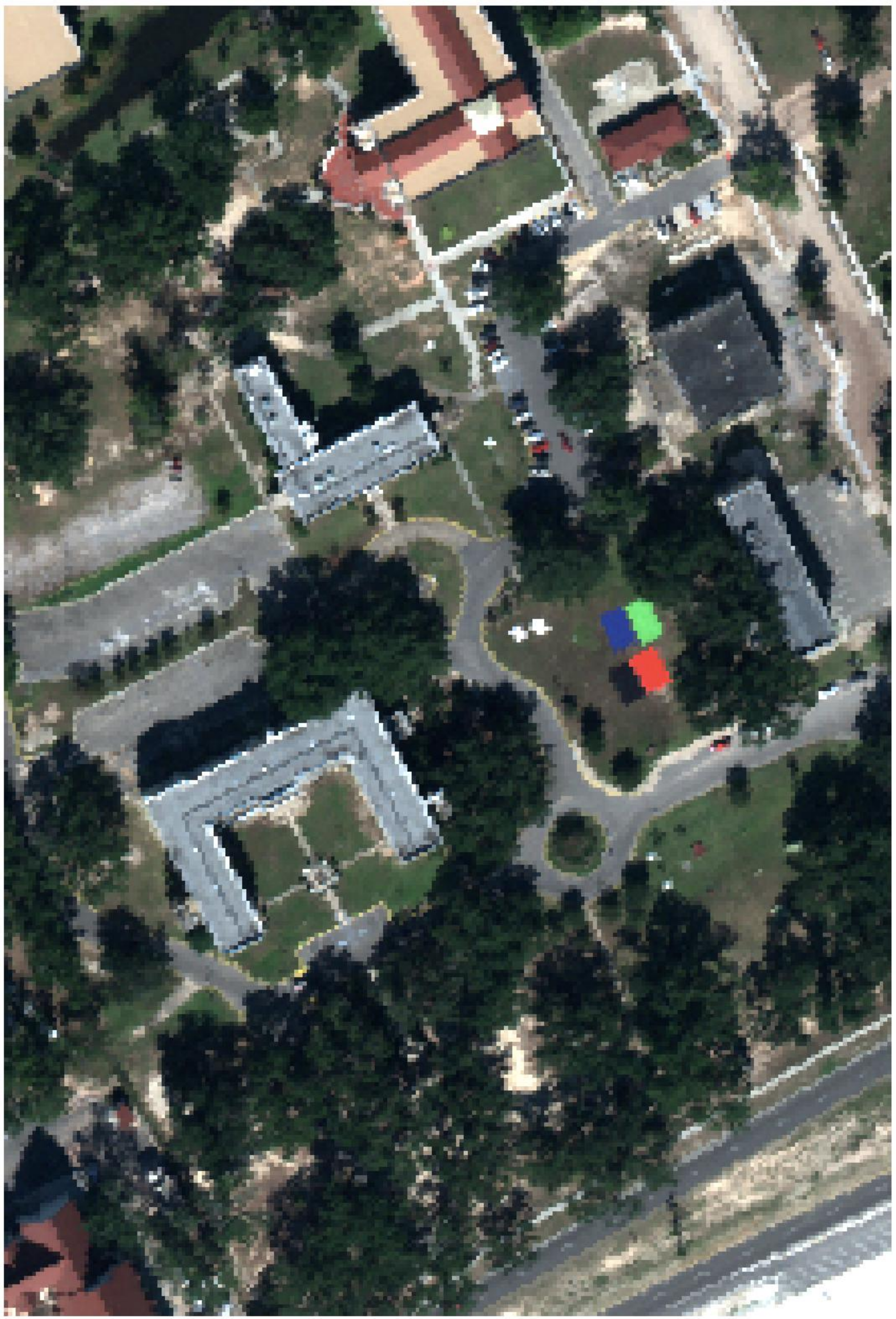} }}
        \qquad
        \subfloat[\footnotesize{Ground Truth Map}]{{\includegraphics[width=3.9cm]{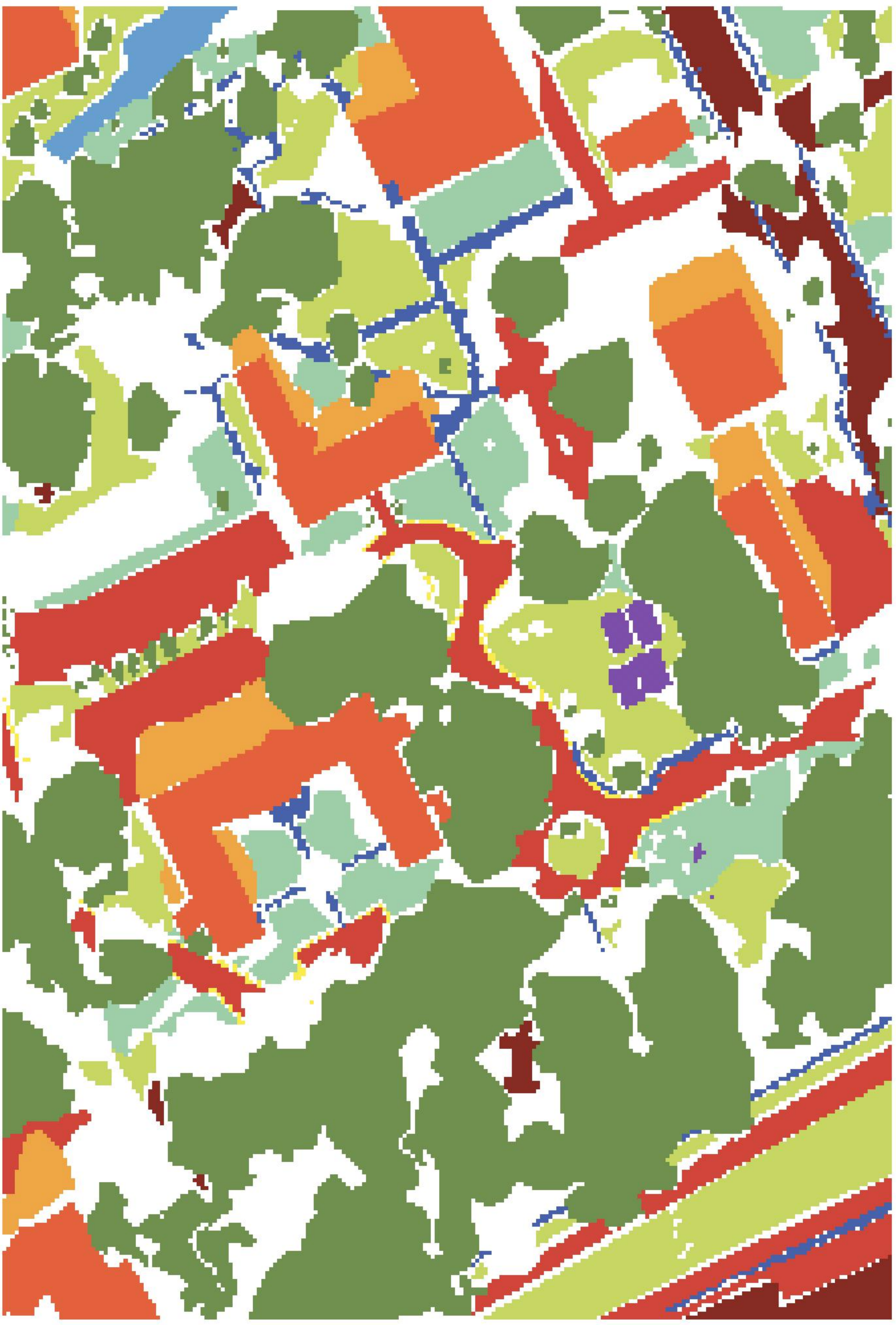} }}
        \qquad
        \subfloat[\footnotesize{Classification Map}]{{\includegraphics[width=3.9cm]{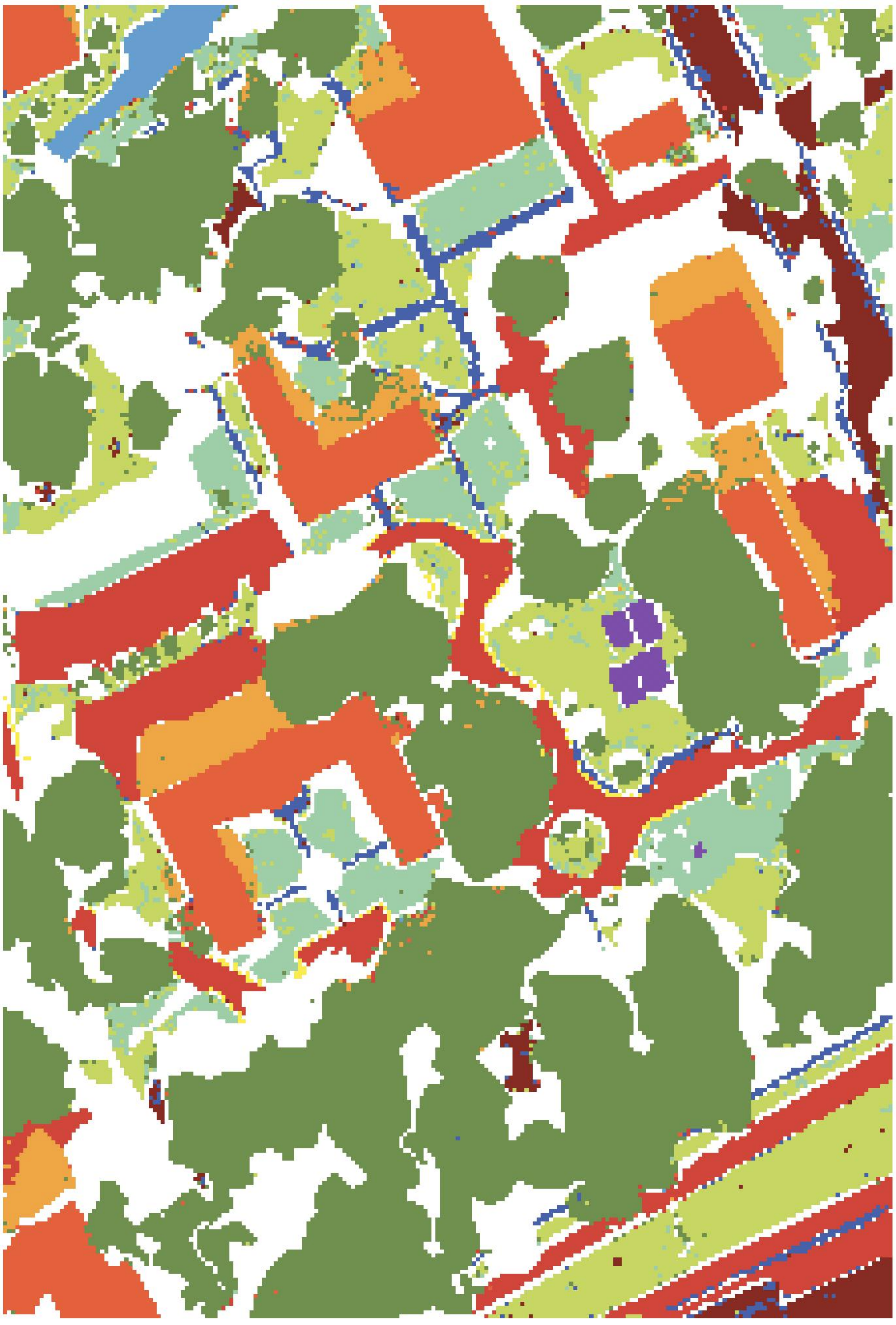} }}
        \\
        \subfloat{{\includegraphics[width=15cm]{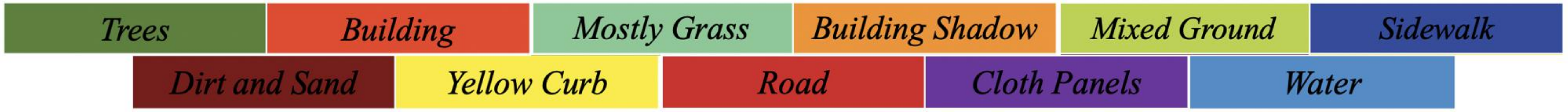} }}

        \caption{Visualization of the classification maps from the MUUFL data}
        \label{fig:MapMUUFL}
    \end{figure*}

    \begin{table*}[!htb]
    \caption{\textsc{Classification performance (\%) on the MUUFL Data is shown here. The concatenated CoMMANet embeddings of HSI and LiDAR are classified using a neural network, KNN, and a KNN and neural network ensemble.}}

    \centering
     \begin{tabular}{||c | c | c | c | c ||} 
     \hline\hline
     \multirow{2}{*}{} & \multirow{2}{*}{} & \multicolumn{3}{c||}{HSI + LiDAR}
     \\\cline{3-5}
     No. & Class & \multirow{2}{*}{Neural Network} & \multirow{2}{*}{KNN ($k=35$)} & \multirow{2}{*}{Neural Network + KNN ($k=35$)}\\ [0.5ex] 
      & & Accuracy & Accuracy & Accuracy\\ [0.5ex] 

     \hline\hline
     1 & Trees  & $96.8\pm0.03$ & $96.6\pm0.03$ & $97.0\pm0.02$\\
     2 & Mostly Grass  & $84.5\pm0.15$ & $87.1\pm0.15$ & $87.3\pm0.12$\\
     3 & Mixed ground surface & $86.9\pm0.25$ & $89.6\pm0.48$ & $89.2\pm0.26$\\
     4 & Dirt and sand  & $90.6\pm0.40$ & $94.8\pm0.32$ & $94.8\pm0.30$\\
     5 & Road  & $95.5\pm0.18$ & $96.3\pm0.12$ & $96.3\pm0.10$\\
     6 & Water & $94.6\pm0.28$ & $96.2\pm0.23$ & $96.5\pm0.21$\\
     7 & Building shadow & $83.5\pm0.50$ & $80.9\pm0.21$ & $83.4\pm0.21$\\
     8 & Building  & $72.6\pm0.08$ & $98.0\pm0.12$ & $98.0\pm0.08$\\
     9 & Sidewalk  & $82.6\pm1.10$ & $84.3\pm0.24$ & $81.5\pm1.00$\\
     10 & Yellow curb  & $85.5\pm0.52$ & $90.9\pm0.40$ & $89.9\pm0.40$\\
     11 & Cloth panels  & $95.8\pm0.35$ & $98.7\pm0.43$ & $98.1\pm0.38$\\
     \hline
     \multicolumn{2}{||c|}{OA (\%)}  & $\textbf{93.1}\pm\textbf{0.15}$ & $\textbf{94.0}\pm\textbf{0.10}$ & $\textbf{94.3}\pm\textbf{0.12}$\\
     \multicolumn{2}{||c|}{AA (\%)}  & $\textbf{87.6}\pm\textbf{0.08}$ & $\textbf{93.1}\pm\textbf{0.15}$ & $\textbf{91.9}\pm\textbf{0.10}$\\
     \multicolumn{2}{||c|}{Kappa (\%)}  & $\textbf{90.8}\pm\textbf{0.10}$ & $\textbf{92.1}\pm\textbf{0.12}$ & $\textbf{92.3}\pm\textbf{0.08}$\\
    
     \hline\hline
    
     \end{tabular}
    \label{table:muufl_result}
    \end{table*}

    \begin{table}[!htb]
    \caption{\textsc{MUUFL Data: The effectiveness of a unified classification model is shown here. The classifier is trained on embeddings of one sensor and tested on embeddings of another sensor.}}

    \centering
     \begin{tabular}{||c | c | c | c ||} 
     \hline\hline
     Neural Network &  & \multicolumn{2}{c||}{Neural Network }\\
     Classifier & Evaluation & \multicolumn{2}{c||}{Classifier \textit{tested} }\\
     \textit{trained} on & Metric & \multicolumn{2}{c||}{on embeddings of}\\
     embeddings of & (\%) & \multicolumn{2}{c||}{}
     \\\cline{3-4}
     & & HSI & LiDAR\\ [0.5ex]
     \hline
      & OA & $88.22\pm0.05$ & $87.04\pm0.10$
     \\\cline{2-4}
     HSI & AA & $84.84\pm0.45$ & $74.85\pm0.40$
     \\\cline{2-4}
      & Kappa & $84.58\pm0.08$ & $82.89\pm0.21$  \\

     \hline
      & OA & $88.01\pm0.20$ & $87.47\pm0.12$
     \\\cline{2-4}
     LiDAR & AA & $84.82\pm0.32$ & $76.12\pm0.42$
     \\\cline{2-4}
      & Kappa & $84.35\pm0.12$ & $83.50\pm0.18$  \\
     \hline\hline
    
    \end{tabular}
    \label{table:muufl_unified}
    \end{table}

    \begin{table}[!htb]
    \caption{\textsc{MUUFL Data: The latent space of one sensor is predicted using another sensor, and then reconstructed using the sensor's decoder.}}
    
    \centering
     \begin{tabular}{||c | c | c | c||} 
     \hline\hline
     Sensor & Sensor & Latent space & Reconstructed \\ [0.5ex] 
     (Predictor) & (Predicted) & MSE  & Data MSE \\ [0.5ex] 
     \hline\hline
     HSI & LiDAR & $0.10\pm0.04$ & $0.015\pm0.003$ \\
     \hline
     LiDAR & HSI & $0.10\pm0.06$ & $0.009\pm0.004$ \\
     \hline\hline
     \multicolumn{4}{|c|}{}  \\
     \multicolumn{4}{|c|}{* \textit{Latent space values $\in$ [-1,1]} \qquad \textit{HSI and LiDAR data $\in$ [0,1]}}  \\
     \hline
    
     \end{tabular}
    \label{table:muufl_missing}
    \end{table}

     \begin{figure}[!ht]
        \centering
        \subfloat[\footnotesize{LiDAR Height Ground Truth (in meters) }]{{\includegraphics[width=3.4cm]{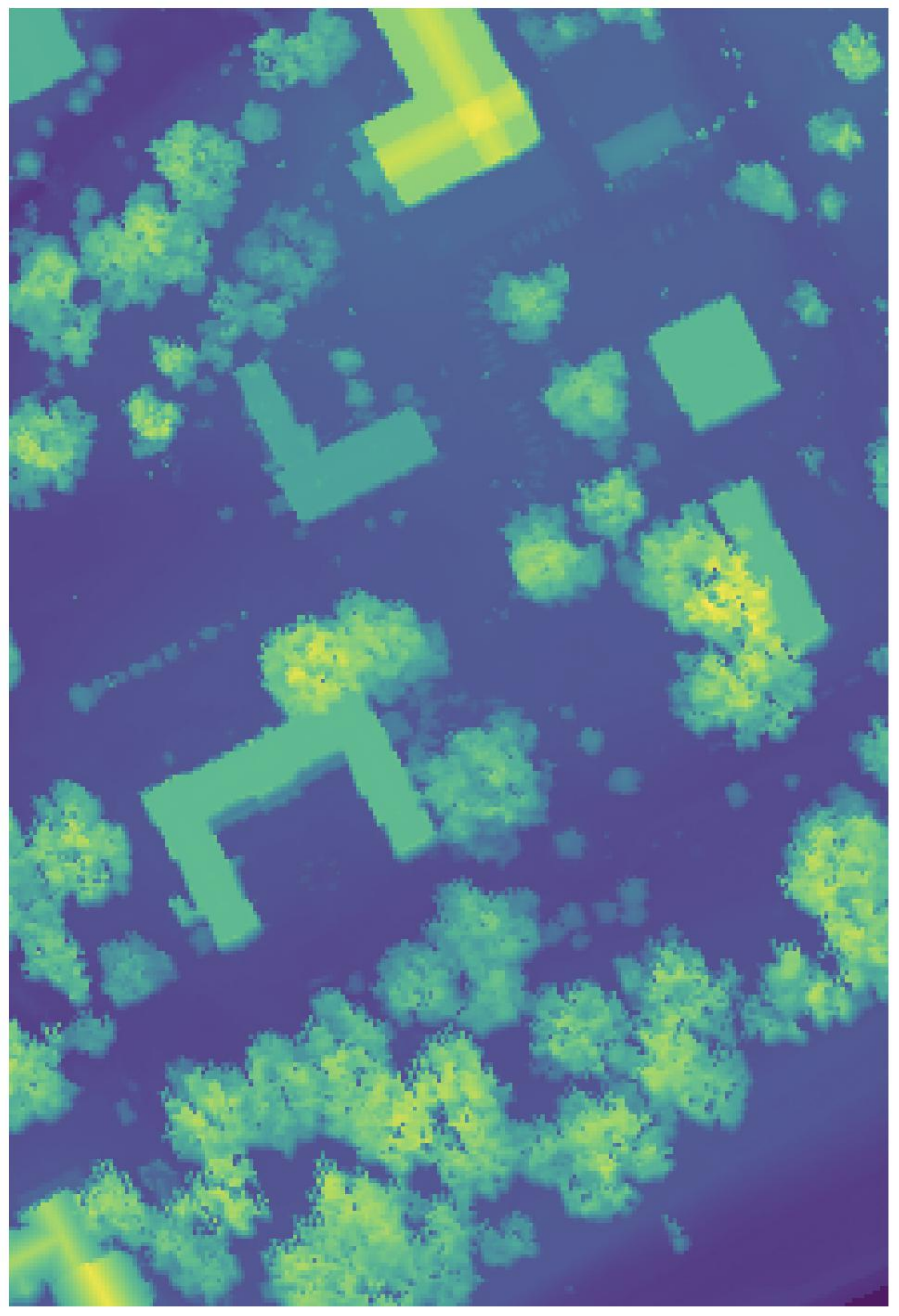} }}
        \hspace{0.1mm}
        \subfloat[\footnotesize{LiDAR Height Predicted from HSI (in meters)}]{{\includegraphics[width=3.4cm]{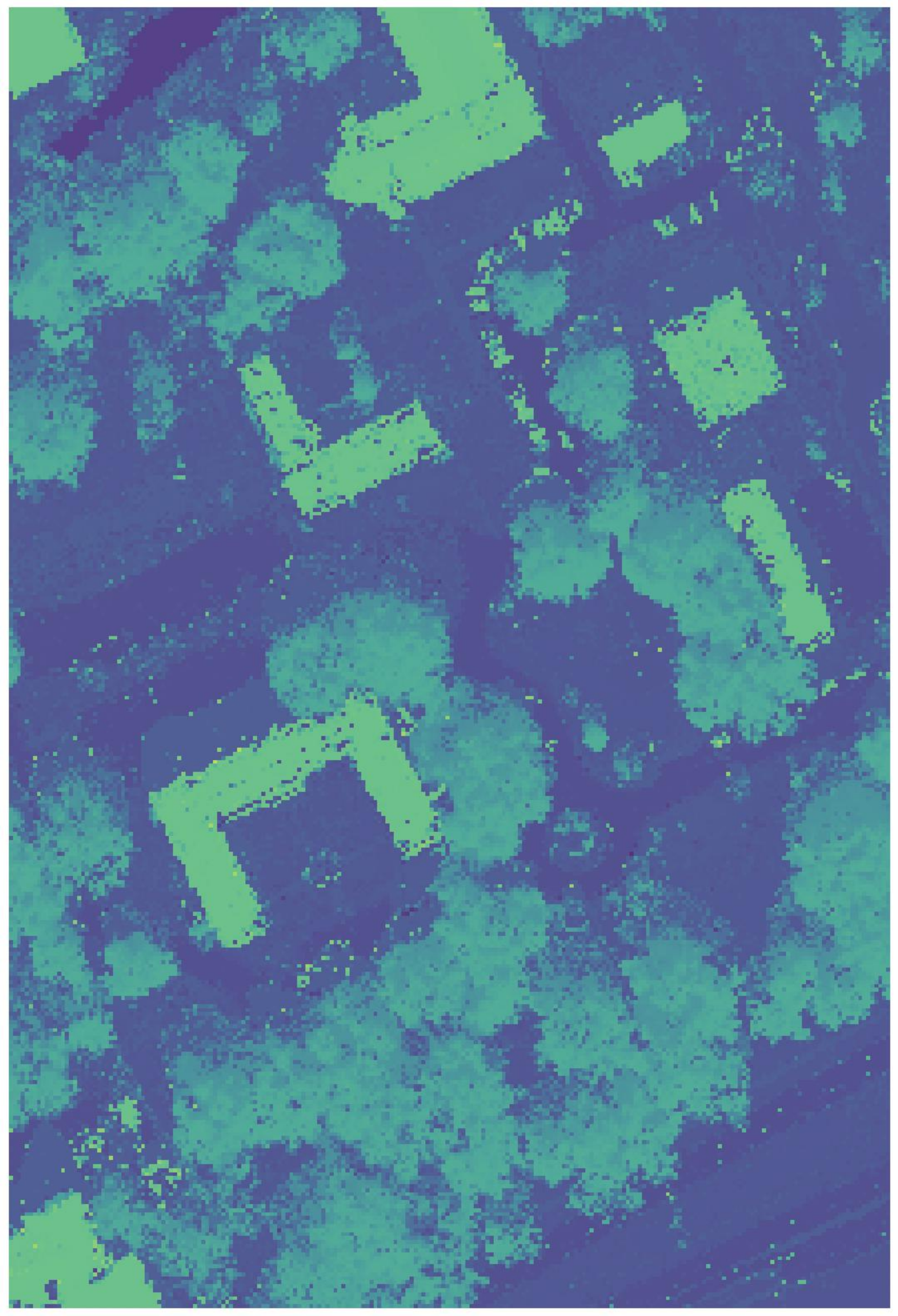} }}
        \hspace{0.1mm}
        \subfloat{{\includegraphics[width=0.86cm]{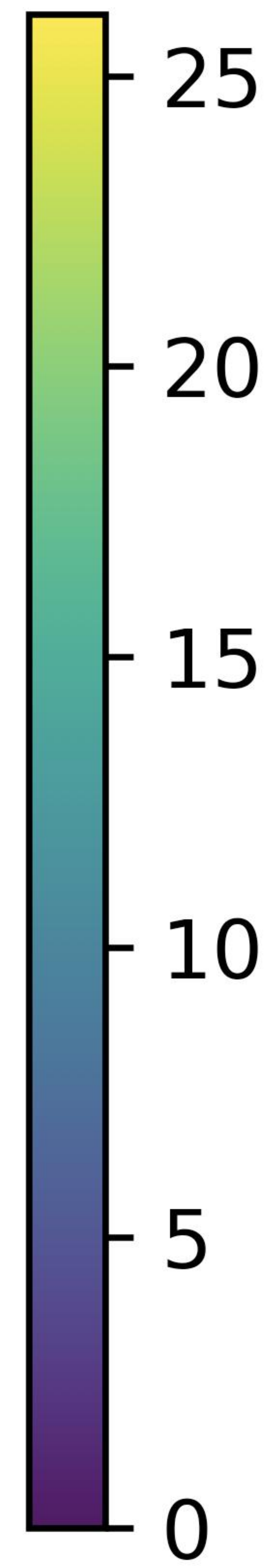} }}
        \\
        \subfloat[\footnotesize{LiDAR Intensity Ground Truth (between 0 and 1) }]{{\includegraphics[width=3.4cm]{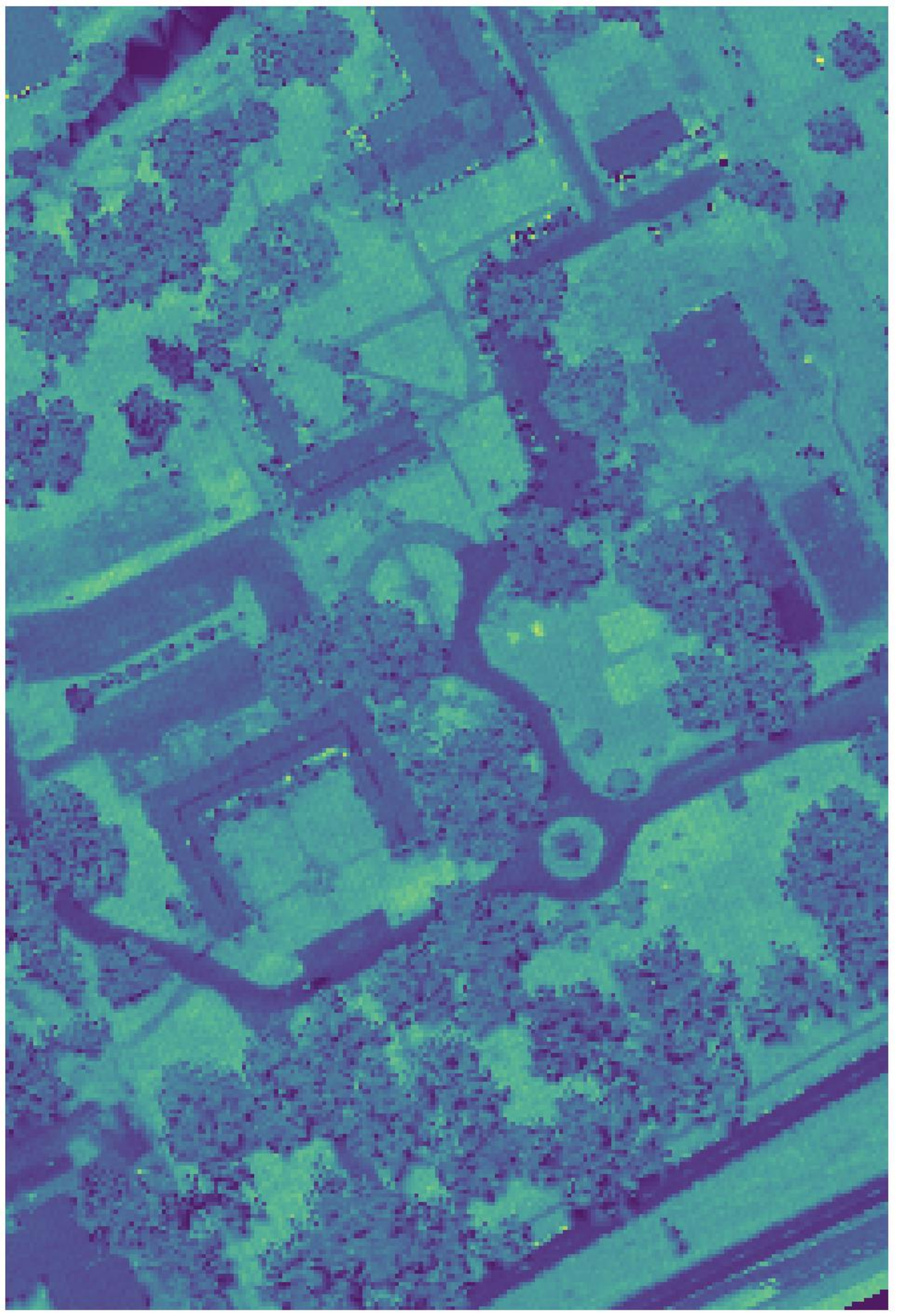} }}
        \hspace{0.1mm}
        \subfloat[\footnotesize{LiDAR Intensity Predicted from HSI  (between 0 and 1)}]{{\includegraphics[width=3.4cm]{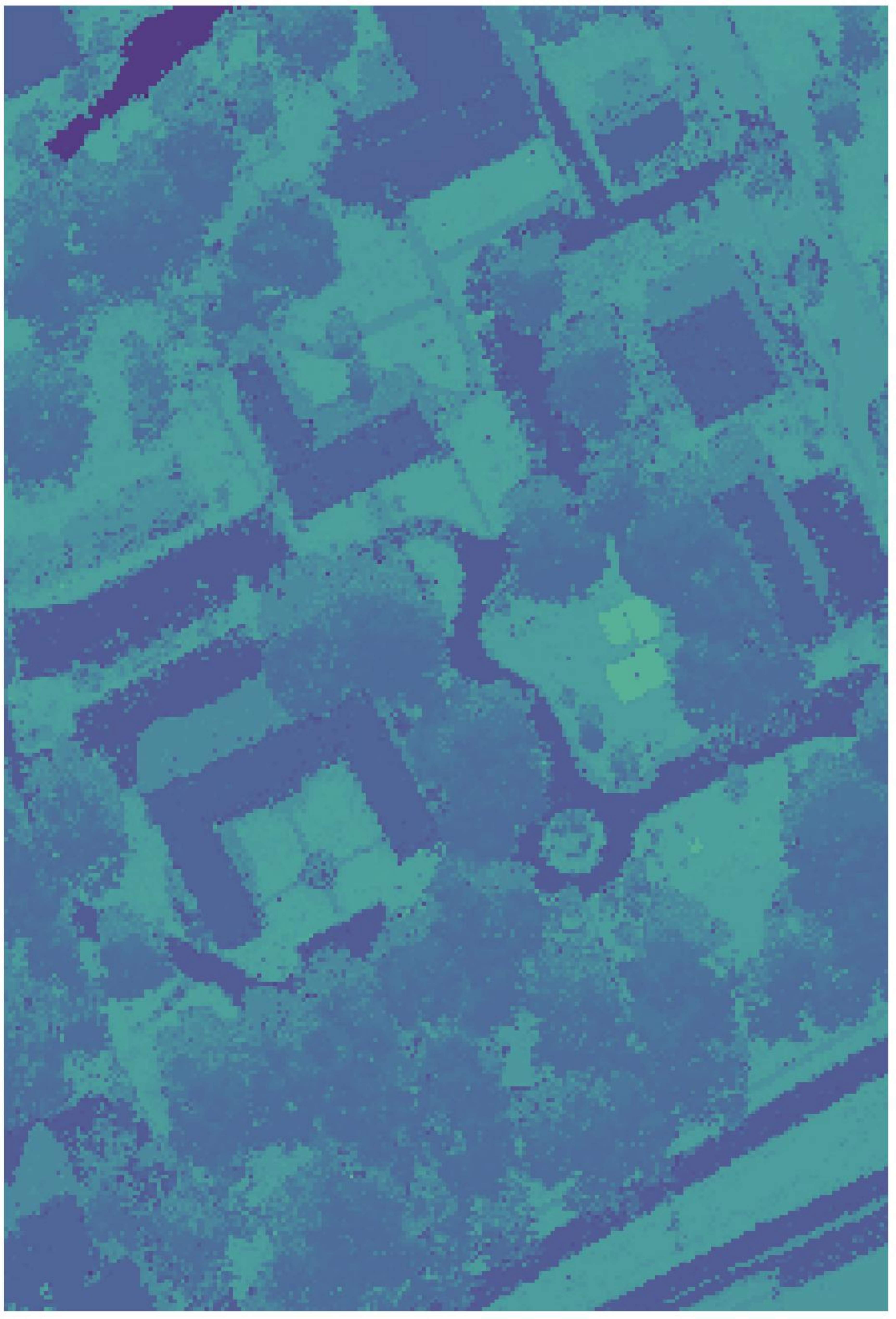} }}
        \hspace{0.1mm}
        \subfloat{{\includegraphics[width=1.05cm]{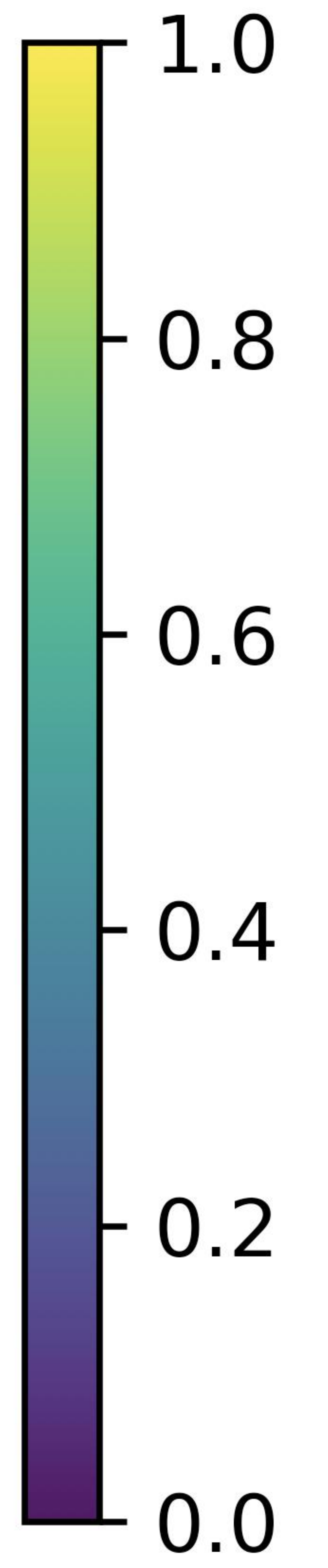} }}
        \caption{The LiDAR latent space is predicted using the HSI latent space from the MUUFL dataset. Using the LiDAR decoder, the predicted latent space is reconstructed to get the LiDAR height and intensity, which are shown here.}
        \label{fig:PredLidar}
    \end{figure}


    \textbf{\textit{Missing sensor prediction/ reconstruction:}} For the prediction of one sensor's embeddings from another sensor, a shallow neural network is used with two hidden layers. The embeddings are 32-dimensional vectors. The two hidden layers contain $128$ and $64$ hidden units, respectively. A \textit{tanh} activation is used on the final layer. The batch size is set to $32$. The model is trained using 5-fold validation for $100$ epochs in each fold using the Adam optimizer. The mean squared error of the latent space predictions are least when $\alpha$ and $\gamma$ are both set to $0.4$ while training the CoMMANet to generate shared embeddings. \par 
    
    After the embeddings of a sensor are predicted, the decoder is used to reconstruct the original data. The reconstructed LiDAR height and intensity are shown in Fig. \ref{fig:PredLidar}. Similarly, HSI embeddings are predicted using the LiDAR embeddings and then reconstructed using the decoder model. The prediction metrics are shown in Table \ref{table:muufl_missing}. The images of original and predicted HSI spectra are available here\footnote{\url{https://github.com/GatorSense/AdaptiveManifoldLearning_CBL}\label{foot}}. The data is reconstructed with a low mean squared error. In LiDAR data reconstruction, the classes having a lot of variation like trees and mixed ground show slightly higher reconstruction loss. It is slightly more challenging to reconstruct the HSI spectra using LiDAR only. Nevertheless, the reconstruction loss of spectra is very low.

    \item \textbf{\textit{HS-SAR Berlin Data}} \\
    \textbf{\textit{Shared embeddings:}} In this dataset, the training and testing samples are provided separately. The optimal size of the input patches from SAR is $11$ x $11$ pixels. A single pixel is used from HSI for training. The HSI data already lies between $0$ and $1$. Each channel of SAR is scaled between $0$ and $1$. The weight for the similarity enhancement term, $\gamma$, is set to $0.4$ and the margin, $\alpha$, is set to $1$ for the best performance. For training the triplet network, the semi-hard triplets mixed with a few easy triplets are used. The Similarity Enhancement (SE) term is also added to the loss function. \par 
    The CoMMANet is trained for $10$ checkpoints with $50$ epochs in each checkpoint. In each checkpoint, $280,000$ triplets are used. The embeddings of each sensor are $32$-dimensional vectors, and a \textit{tanh} activation is applied on the latent space. The learning rate is set to $0.001$, the batch size is set to $512$, and the model is trained using the Adam optimizer. Due to fewer samples in the training dataset, the model is prone to overfitting. Therefore, a 10-fold validation is used to avoid overfitting.\par 
    
    \textbf{\textit{Classification:}} Firstly, the CoMMANet embeddings of HSI and SAR are concatenated and classified using an ensemble of three neural networks and KNN. Using the neural network ensemble, the best overall classification accuracy is $71.26$\%. The best average accuracy is $63.26$\% which is achieved using KNN (with $k=51$). Since all the previous methods reported their best accuracy, the best accuracy achieved is reported here instead of mean accuracy for the sake of comparison. The sensitivity of the KNN model to the value of $k$ is shown in Fig. \ref{fig:Berlin_KNN}. The classification performance on the Berlin dataset is shown in Table \ref{table:berlin_result}. The classification map is shown in Fig. \ref{fig:MapBerlin}.\par 
    Secondly, a classifier is trained on one sensor's embeddings and tested on other sensor's embeddings. The results listed in Table \ref{table:berlin_unified} show that the classification models developed for one sensor's embeddings can accurately predict the embeddings of other sensors also.

     \begin{figure}[!ht]
        \centering
        \includegraphics[width=7.8cm]{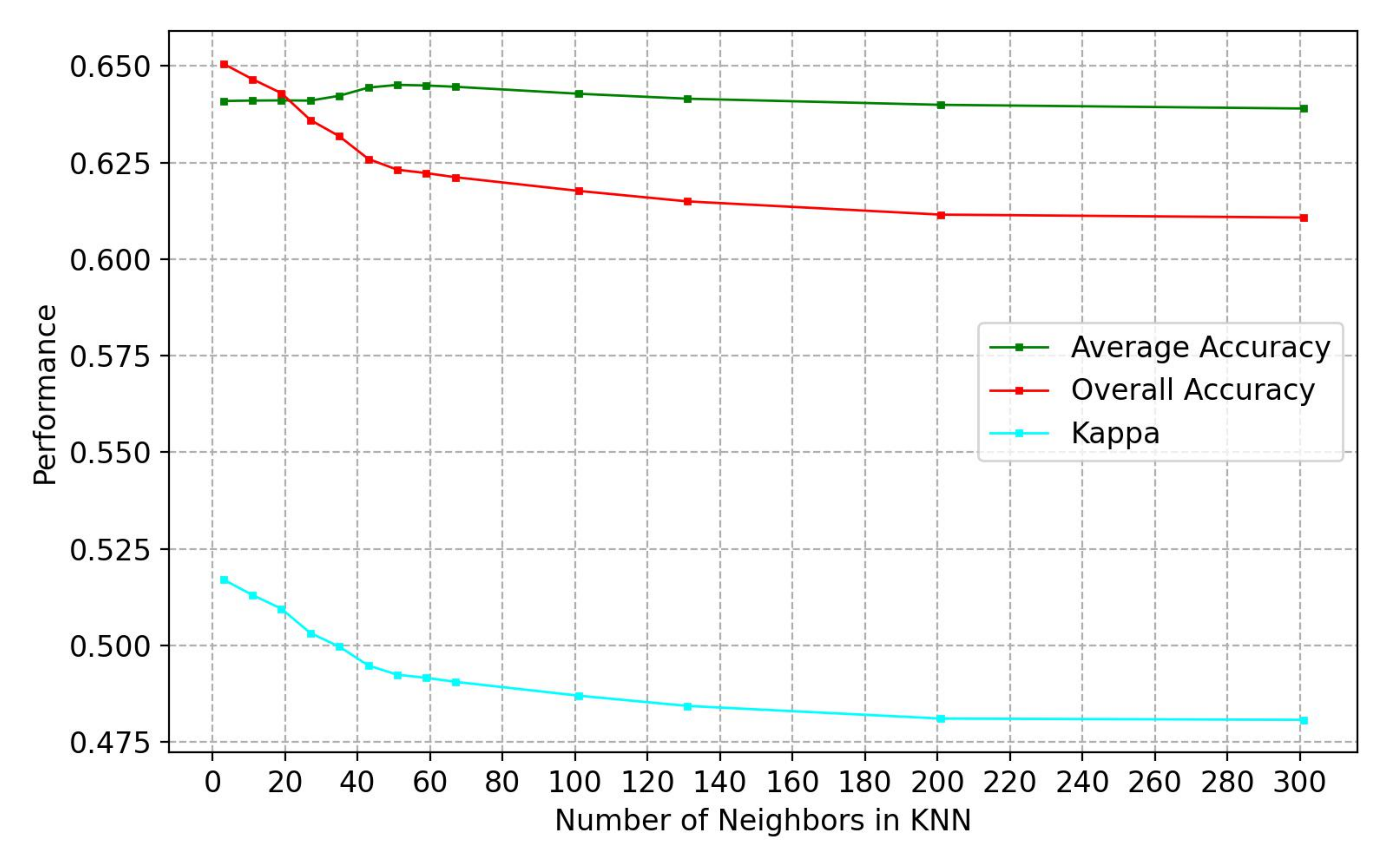}
        \caption{The sensitivity of the KNN model to the value of $k$ is shown for the HS-SAR Berlin dataset. The KNN is used for the classification of concatenated HSI and SAR embeddings from CoMMANet.}
        \label{fig:Berlin_KNN}
    \end{figure}

    \begin{table*}[!ht]
    \caption{\textsc{Classification performance (\%) on the HS-SAR Berlin data is shown here. Since all the other methods reported their best accuracy, the best accuracy is shown here for comparison. The concatenated CoMMANet embeddings of HSI and SAR are classified using a neural network, KNN, and a KNN and neural network ensemble.}}

    \centering
     \begin{tabular}{||c | c | c | c | c||} 
     \hline\hline
     \multirow{2}{*}{} & \multirow{2}{*}{} & \multicolumn{3}{c||}{HSI + SAR}  \\ [0.5ex] 
     \cline{3-5}
     No. & Class  & \multirow{2}{*}{Neural Network } & \multirow{2}{*}{KNN ($k=51$)} & \multirow{2}{*}{Neural Network + KNN ($k=51$)} \\ [0.5ex] 
      &  & Best Accuracy & Best Accuracy & Best Accuracy \\ [0.5ex] 

     \hline\hline
     1 & Forest & $78.64$ & $73.65$ & $78.81$ \\
     2 & Residential Area  & $81.76$ & $72.82$ & $81.32$ \\
     3 & Industrial Area  & $35.60$ & $36.96$ & $35.96$ \\
     4 & Low Plants  & $76.48$ & $74.30$ & $77.05$ \\
     5 & Soil  & $63.71$ & $69.57$ & $65.96$ \\
     6 & Allotment  & $32.84$ & $28.83$ & $33.59$ \\
     7 & Commercial Area  & $18.85$ & $19.58$ & $18.67$ \\
     8 & Water  & $62.20$ & $61.69$ & $63.06$ \\
     \hline
     \multicolumn{2}{||c|}{OA (\%)}  & $\textbf{71.26}$ & $\textbf{63.26}$  & $\textbf{71.10}$\\
     \multicolumn{2}{||c|}{AA (\%)} & $\textbf{60.50}$ & $\textbf{64.19}$  & $\textbf{61.15}$\\
     \multicolumn{2}{||c|}{Kappa (\%)}  & $\textbf{57.20}$ & $\textbf{50.03}$  & $\textbf{57.31}$\\
     \hline\hline
    
     \end{tabular}
    \label{table:berlin_result}
    \end{table*}

    \begin{table}[!htb]
    \caption{\textsc{HS-SAR Berlin Data: The effectiveness of a unified classification model is shown here. The classifier is trained on embeddings of one sensor and tested on embeddings of another sensor. The best accuracy (\%) is shown.}}

    \centering
     \begin{tabular}{||c | c | c | c ||} 
     \hline\hline
     Neural Network Classifier & Evaluation & \multicolumn{2}{c||}{Neural Network Classifier}\\
     \textit{trained} on & Metric & \multicolumn{2}{c||}{ \textit{tested} on embeddings of}\\
     embeddings of & (\%) & \multicolumn{2}{c||}{}
     \\\cline{3-4}
     & & HSI & SAR\\ [0.5ex]
     \hline
      & OA & $59.01$ & $50.84$
     \\\cline{2-4}
     HSI & AA & $64.24$ & $44.70$
     \\\cline{2-4}
      & Kappa & $46.28$ & $34.29$  \\

     \hline
      & OA & $59.82$ & $50.78$
     \\\cline{2-4}
     SAR & AA & $64.44$ & $43.96$
     \\\cline{2-4}
      & Kappa & $46.85$ & $33.98$  \\
     \hline\hline
    
    \end{tabular}
    \label{table:berlin_unified}
    \end{table}

     \begin{figure}[!ht]
        \centering
        \subfloat[\footnotesize{Ground Truth Map }]{{\includegraphics[width=2.6cm]{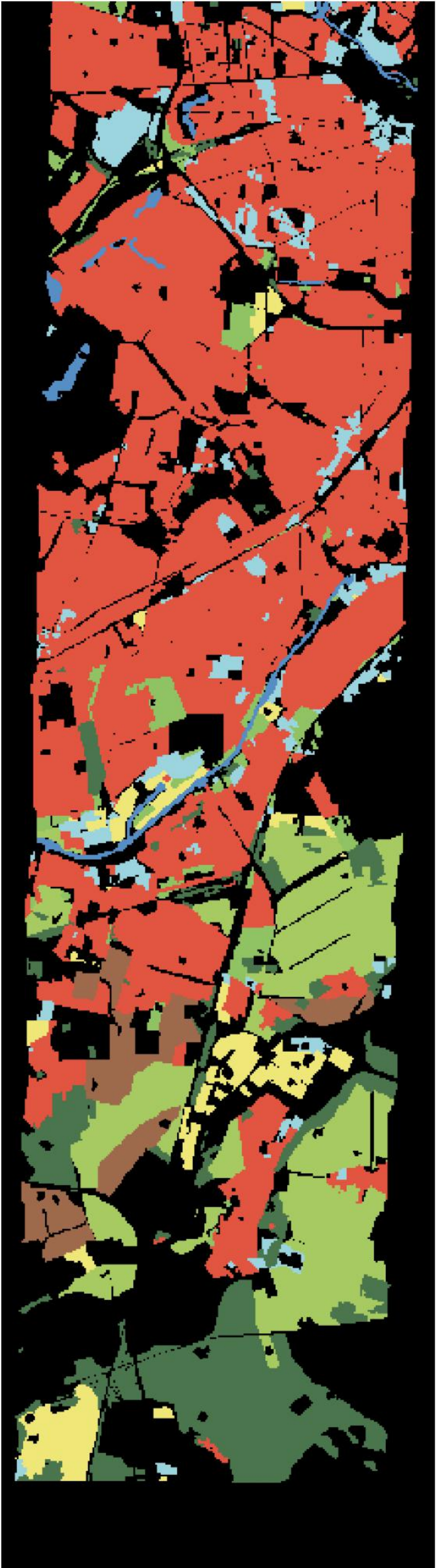} }}
        \qquad
        \subfloat[\footnotesize{Classification Map }]{{\includegraphics[width=2.6cm]{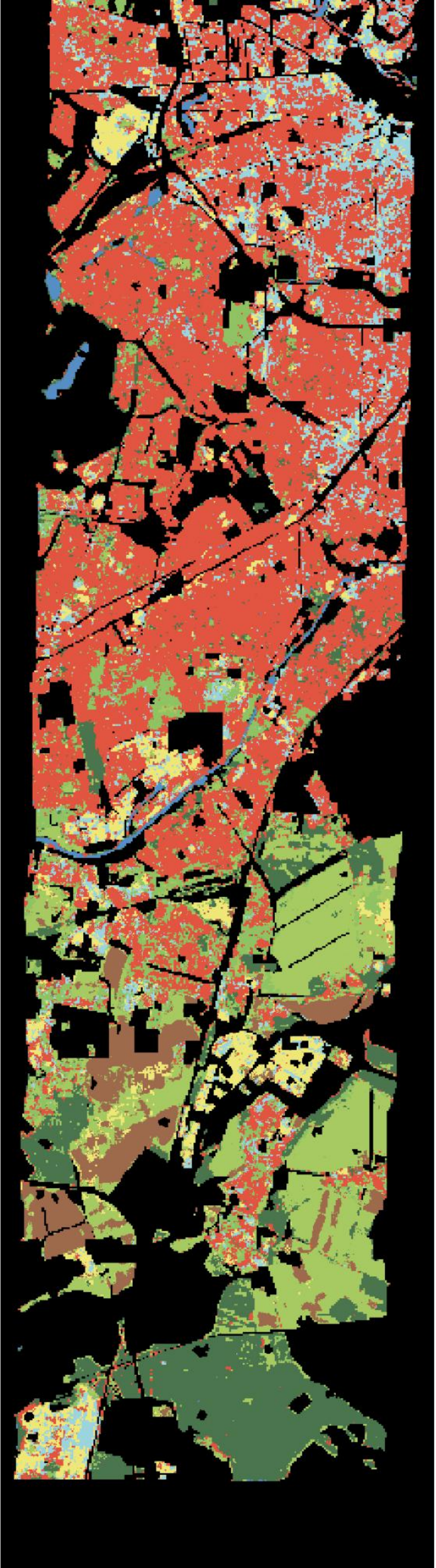} }}\\
        \subfloat{{\includegraphics[width=8.2cm]{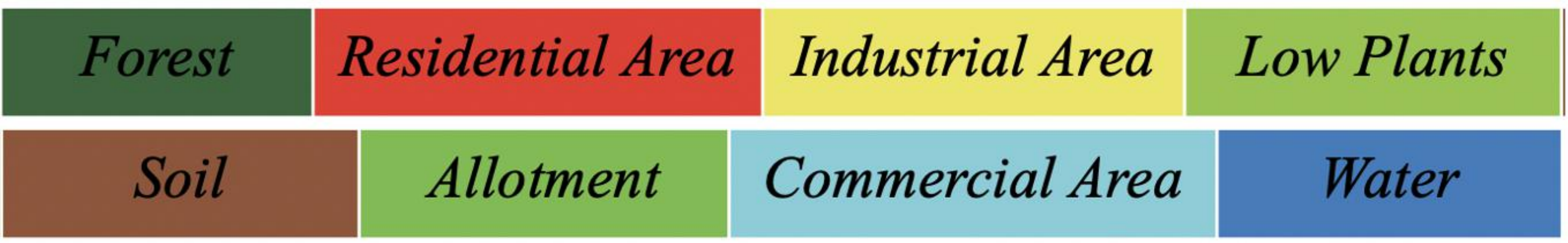} }}

        \caption{Visualization of the classification maps from the HS-SAR Berlin data}
        \label{fig:MapBerlin}
    \end{figure}

 \textbf{\textit{Missing sensor prediction/ reconstruction:}} For the missing sensor prediction, the same model as the MUUFL data is used. The mean squared error of the predictions is least when $\alpha$ is set to $1$, and $\gamma$ is set to $0.4$ in the CoMMANet. All the training parameters are the same as the parameters used for the MUUFL data. The prediction metrics are shown in Table \ref{table:berlin_missing}. \par
 After the embeddings of a sensor are predicted, the decoder is used to reconstruct the original data. The images of original and predicted SAR and HSI bands are available here\footnote{See footnote \ref{foot}}. The data is reconstructed with a reasonably low mean squared error. The SAR data has a smaller reconstruction error compared to the HSI reconstruction. In the testing data, the number of samples is significantly higher than the number of training samples. Therefore, the spectra of testing data have a significant amount of variation, which causes the accuracy of predicted spectra to drop. However, the spectra are successfully predicted with a low mean squared error.

    \begin{table}[!ht]
    \caption{\textsc{HS-SAR Berlin Data: The latent space of one sensor is predicted using another sensor, and then reconstructed using the sensor's decoder.}}
    
    \centering
     \begin{tabular}{||c | c | c | c||} 
     \hline\hline
     Sensor & Sensor & Latent space & Reconstructed \\ [0.5ex] 
     (Predictor) & (Predicted) & MSE & Data MSE \\ [0.5ex] 
     \hline\hline
     HSI & SAR & $0.92\pm0.03$ & $0.026\pm0.005$ \\
     \hline
     SAR & HSI & $0.95\pm0.06$ & $0.023\pm0.011$ \\
     \hline\hline
     \multicolumn{4}{|c|}{}  \\
     \multicolumn{4}{|c|}{* \textit{Latent space values $\in$ [-1,1]} \qquad \textit{HSI and SAR data $\in$ [0,1]}}  \\
     \hline
    
     \end{tabular}
    \label{table:berlin_missing}
    \end{table}

\end{enumerate}

\subsection{Comparison and Analysis}
To validate the effectiveness of the proposed model, a comparison is made with the state-of-the-art models like CNN-PPF \cite{li2016hyperspectral}, FDSSCN \cite{wang2018fast}, CNN-MRF \cite{cao2018hyperspectral}, CRNN\cite {wu2017convolutional}, IAP \cite{hong2020invariant}, Coupled CNN \cite{hang2020classification}, IP-CNN \cite{zhang2021information}, FusAtNet \cite{mohla2020fusatnet}, SpectralFormer \cite{hong2021spectralformer}, FIT \cite{lee2021fnet}, FrIT \cite{zhao2022fractional}, and SOTNet \cite{zhang2022hyperspectral} using their reported results, \cite[Tab. \RNum{11}]{zhang2021information}, and \cite[Tab. \RNum{2}]{zhao2022fractional}. The comparison results are shown in Table \ref{table:muufl_compare}. Using the MUUFL dataset, it shows $3.33\%$, $9.47\%$, $5.36\%$, $2.92\%$, $8.25\%$, $3.37\%$, $0.44\%$, $2.82\%$, $13.09\%$, $8.84\%$, $7.69\%$, and $0.43\%$ improvement in overall accuracy (OA) over CNN-PPF, FDSSCN, CNN-MRF, CRNN, IAP, Coupled CNN, IP-CNN, FusAtNet, SpectralFormer, FIT, FrIT, and SOTNet. The proposed CoMMANet outperforms the other methods in terms of the kappa coefficient and overall accuracy (OA) on the MUUFL Dataset. Moreover, applying a simple method like KNN also gives a high average accuracy and overall accuracy. The CoMMANet even outperforms vision transformer based models \cite{hong2021spectralformer} \cite{zhao2022fractional}. \par 
For many classes, the accuracy is lower compared to the other methods. The reason is that in our model, to extract features from each modality, we use standard convolutional layers instead of Gabor filters \cite{chen2017hyperspectral} \cite{zhao2021fractional} and complementary-structure control \cite{zhang2021information}. Therefore, other methods are able to extract rich spatial information and achieve higher accuracy on some ambiguous classes like ``sidewalk", ``yellow curb", and ``dirt/ sand" from the MUUFL dataset. Gabor filters are known to extract rich texture information from an image. It can increase the robustness of learned representations while reducing the training complexity of the neural networks \cite{li2016road} \cite{rai2020review} \cite{alekseev2019gabornet}. In the CoMMANet, increasing the number of CNN layers becomes very computationally expensive because of the triplet networks. Therefore, fewer convolutional layers are used in the model. The information fusion is performed during the classification by simply concatenating the embeddings.\par 
Zhang et al. \cite{zhang2021information} used Gram matrices to maintain the complementary structure of their fusion block. They used the Gram matrix from LiDAR as a texture reference for the fused features. Similarly, a Gram matrix from HSI was used as a spectral reference for the fused features. The joint Gram matrices preserve the complementary information from both sensors. The concept is similar to Image Style transfer \cite{xu2021image}. Several other methods \cite{jahan2020inverse} \cite{brell2017hyperspectral} \cite{gu2017discriminative} use different techniques to remove redundant information from the fused representation. But, in this paper, our goal is to show the effectiveness of the proposed architecture without using any additional components/ techniques. The proposed model shows superior results compared to all the other methods in terms of overall accuracy and kappa coefficient on the MUUFL Dataset. 

    \begin{table*}[!htb]
    \caption{\textsc{Comparison of the Classification Accuracy (\%) using the MUUFL Data. The mean accuracy of the proposed model is compared with the other methods' mean accuracy.}}
    
    \centering
    \scalebox{0.9}{%
     \begin{tabular}{||c | c | c | c | c | c | c | c | c | c | c | c | c | c | c ||} 
     \hline\hline
      & &  & &  & & &  & & &  & & & & \multicolumn{1}{c||}{CoMMANet} \\ [0.5ex] 
     \cline{15-15}
     \multirow{2}{*}{No.} & \multirow{2}{*}{Class} & CNN & \multirow{2}{*}{FDSSCN} & \multirow{1}{*}{CNN} & \multirow{2}{*}{CRNN} & \multirow{2}{*}{IAP} & Coupled & \multirow{1}{*}{IP} & \multirow{2}{*}{FusAtNet} & \multirow{1}{*}{Spec} & \multirow{2}{*}{FIT} & \multirow{2}{*}{FrIT} & \multirow{1}{*}{SOT} & Neural \\ [0.5ex] 
      & & PPF & & MRF & & & CNN & CNN & & Former & & & Net & Network \\ [0.5ex] 
      & & & & & & & & & & & & & & + KNN \\ [0.5ex] 
      & & & & & & & & & & & & & & ($k=35$) \\ [0.5ex] 

     \hline\hline
     1 & Trees & 89.07 & 87.37 & 93.04&91.43 & 85.32& 98.90& 94.40 & 98.10 &81.47 & 90.72& 89.61& 96.50 & 97.0 \\
     2 & Mostly Grass & 85.71 & 32.37 & 60.17 & 63.16& 81.99& 78.60& 92.26 & 71.66 &87.54 &78.88 &81.43 & 88.23 & 87.3 \\
     3 & Mixed ground & 80.15 & 88.12 & 90.60&90.20 & 78.51& 90.66& 87.96 & 87.65 &57.47 &71.08 &70.43 & 87.11 & 89.2 \\
      & surface & & & & & & & & & & & & & \\

     4 & Dirt and sand & 93.10 & 94.51 & 97.20& 93.44 & 94.63& 90.60& 97.15 &86.42 & 86.14&92.55 &94.03 & 96.42 & 94.8  \\
     5 & Road & 88.98 & 97.84 & 92.00& 87.62& 86.81& 96.90 & 94.38 & 95.09 &89.13 &83.68 &87.99 & 94.39 & 96.3  \\
     6 & Water &98.93& 96.20& 99.68& 95.89 & 99.79 & 75.98& 99.79 & 90.73 &99.14 &98.93 &98.93 & 99.37 & 96.5 \\
     7 & Building shadow & 89.07& 89.92& 95.39& 90.16 & 90.91& 73.54 & 96.30 & 74.27 &87.37 &90.19 & 90.42& 91.89 & 83.4  \\
     8 & Building &92.15& 87.44& 94.71& 89.29& 95.46&96.66 & 96.13 & 97.55 &90.61 & 87.13&94.57 & 94.84 & 98.0 \\
     9 & Sidewalk &75.45& 85.75& 30.53& 82.91& 73.94 & 64.93 & 94.01 & 60.44 & 66.79& 64.04&69.39 & 90.61& 81.5 \\
     10 & Yellow curb &100.00& 72.73& 36.36& 96.97& 98.91 & 19.47 & 100.00 & 9.39 & 95.08& 97.27 & 73.77 & 100.00 & 89.9 \\
     11 & Cloth panels &100.00& 99.16&95.80 & 96.64& 99.63& 62.76& 99.63 & 93.02&98.88 & 99.63 & 98.88 & 100.00 & 98.1 \\
     \hline
     \multicolumn{2}{|c|}{OA (\%)} &90.97& 84.83 & 88.94 & 91.38 & 86.05& 90.93 & 93.86 & 91.48 & 81.21 & 85.46 & 86.61 & 93.87 & \textbf{94.3}\\
     \multicolumn{2}{|c|}{AA (\%)} &90.24& 84.70&85.02 & 88.88& 89.63& 77.18 & \textbf{95.64} & 78.58 & 76.20 & 81.19 & 82.69 & 94.49 & 91.9 \\
     \multicolumn{2}{|c|}{Kappa (\%)} &84.46& 80.24& 85.55 & 84.41& 82.12& 88.22& 91.99 & 88.65 & 85.42 & 86.74& 86.31& 91.84 & \textbf{92.3} \\
    
     \hline\hline
    
     \end{tabular}
    }
    \label{table:muufl_compare}
    \end{table*}

For the Berlin Dataset, the comparison is made with previous methods using \cite[Tab. \RNum{5}]{wu2021convolutional} and \cite[Tab. \RNum{6}]{li2022asymmetric}, and the results are shown in Table \ref{table:berlin_compare}. The proposed model shows $6.73\%$, $4.55\%$, $4.71\%$, $2.75\%$, $1.41\%$, $4.95\%$, $5.02\%$, and $0.75\%$ improvement in overall accuracy (OA) over CoSpace \cite{hong2019cospace}, LeMA \cite{hong2019learnable}, CapsNet \cite{li2020robust}, Co-CNN \cite{hang2020classification}, CCR-Net \cite{wu2021convolutional}, ContextCNN \cite{lee2017going}, DFINet \cite{gao2021hyperspectral}, and AsyFFNet \cite{li2022asymmetric}. However, classes like ``Commercial Area", ``Residential Area", and ``Allotment" show a lower accuracy compared to the other state-of-the-art methods. Wu et al. \cite{wu2021convolutional} used a cross-channel reconstruction module (CCR) which makes the fusion process more efficient. The CCR module achieves effective information exchange between the two modalities and results in a more compact fusion at the feature level. If the LiDAR or SAR latent space is capable of reconstructing HSI latent space and vice-versa, it indicates that the latent space of both the sensors is highly similar and carries enough information about the target class. Due to this factor, they achieve higher accuracy on some classes. Another problem with the Berlin Dataset is that the number of testing samples (461851) is significantly higher than the number of training samples (2820) (See Table \ref{table:berlin_dist}). Therefore, the difference in training and testing data distributions causes the performance to drop, and the proposed model could not learn the representation of classes having fewer samples effectively. As a result, the average accuracy (AA) gets slightly lower than the other methods. However, the proposed model still outperformed all the other methods in terms of overall accuracy on the Berlin Dataset.\par

Additionally, in all three datasets, the classification models trained on one sensor's embeddings are able to classify other sensor's embeddings accurately. But if a classifier is trained on embeddings of a sensor having a low discriminative ability, the classification accuracy on all sensors drops. However, it is successfully demonstrated that using the proposed CoMMANet, the embeddings from different sensors can be aligned, and a unified classification/ analysis model is possible, which is sensor agnostic. It eliminates the need to develop separate classification models for every sensor.

    \begin{table*}[!htb]
    \caption{\textsc{Comparison of the Classification Accuracy (\%) using the HS-SAR Berlin data. For comparison, the best accuracy is shown since all the state-of-the-art methods reported their best accuracy.}}

    \centering
     \begin{tabular}{||c | c | c | c | c | c | c | c | c | c | c | c ||} 
     \hline\hline
     & & & & & & & & & & \multicolumn{2}{c||}{CoMMANet}\\ [0.5ex] 
     \cline{11-12}
     No. & Class & CoSpace & LeMA & CapsNet & CoCNN & CCRNet & Context & DFINet & AsyFFNet & Neural & KNN \\ [0.5ex] 
     & & & & & & & CNN & & & Network & ($k=51$) \\ [0.5ex] 

     \hline\hline
     1 & Forest & 85.09 & 85.11 & 84.96 & 84.09 & 85.93 & 77.22& 80.29& 76.65 & 78.64 & 73.65 \\
     2 & Residential Area & 61.60 & 64.84 & 65.22 & 68.48 & 68.07 & 63.69 & 61.93& 70.76 & 81.76 & 72.82\\
     3 & Industrial Area & 51.18 & 48.94 & 48.42 & 49.09 & 53.17 &61.44 & 47.44 & 60.16 & 35.60 & 36.96 \\
     4 & Low Plants & 75.44 & 80.04 & 80.80 & 79.43 & 82.62& 73.77 & 80.01 & 74.66 & 76.48 & 74.30\\
     5 & Soil  &  82.50 & 80.66 & 69.18 & 81.25 & 85.10 & 87.22 & 77.54 & 79.18 & 63.71 & 69.57\\
     6 & Allotment & 54.66 & 54.07 & 55.08 & 50.68 & 63.02 & 82.88 & 73.42& 79.24 & 32.84 & 29.83\\
     7 & Commercial Area & 28.81 & 27.40 & 26.12 & 26.16 & 29.23 & 31.13& 49.11 & 37.94 & 18.85 & 19.58\\
     8 & Water & 60.78 & 57.75 & 59.69 & 59.52 & 68.78 & 74.24 & 77.59 & 83.90 &  62.20 & 61.69\\
     \hline
     \multicolumn{2}{|c|}{OA (\%)} & 64.53 & 66.71 & 66.55 & 68.51 & 69.85 & 66.31 & 66.24& 70.51 & \textbf{71.26} & 63.26\\
     \multicolumn{2}{|c|}{AA (\%)} & 62.51 & 62.35 & 61.18 & 62.34 & 66.99 & 68.95 & 68.42& \textbf{70.31} & 60.50 & 64.19\\
     \multicolumn{2}{|c|}{Kappa (\%)} & 50.93 & 53.12 & 52.77 & 54.76 & 57.16 & 54.03 & 53.98 & \textbf{58.24} & 57.20 & 50.03\\
     \hline\hline
    
     \end{tabular}
    \label{table:berlin_compare}
    \end{table*}



\section{Ablation Studies}
\label{sec:Ablation}

The proposed CoMMANet is affected by several components and hyperparameters. To investigate the effect of each component on the performance of the model, ablation studies are conducted by varying one component at a time.

\subsection{Effect of the Similarity Enhancement (SE) Term}\label{sssec:se_effect}
The similarity enhancement term is added to enhance the clustering process. However, there is a trade-off between the SE weight parameter, $\gamma$, and the classification accuracy. If the value of $\gamma$ is high, the ambiguity between the classes is not represented properly because the samples get too close to a particular class. Therefore, the classification accuracy decreases. For example, in MUUFL data, the two classes: ``Dirt and Sand" and ``Sidewalk", are slightly ambiguous. Therefore, a low value of $\gamma$ should be chosen for classification tasks. \par 

\begin{figure*}[!ht]
        \centering
        \subfloat[\footnotesize{AVIRIS-NG/ NEON Data Embeddings}]{{\includegraphics[width=5.8cm]{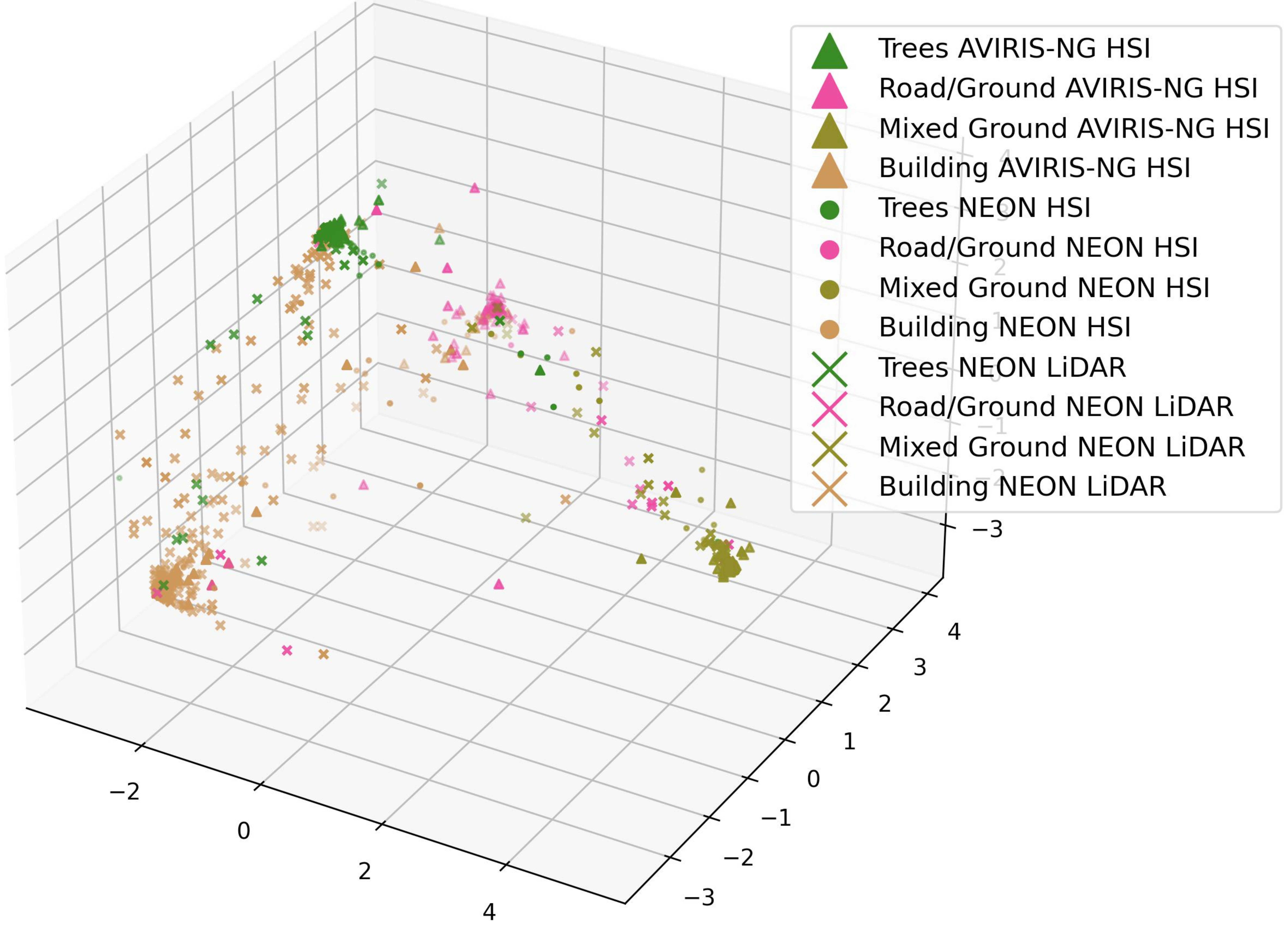} }}
        \hspace{0.01em}
        \subfloat[\footnotesize{MUUFL Data Embeddings}]{{\includegraphics[width=5.8cm]{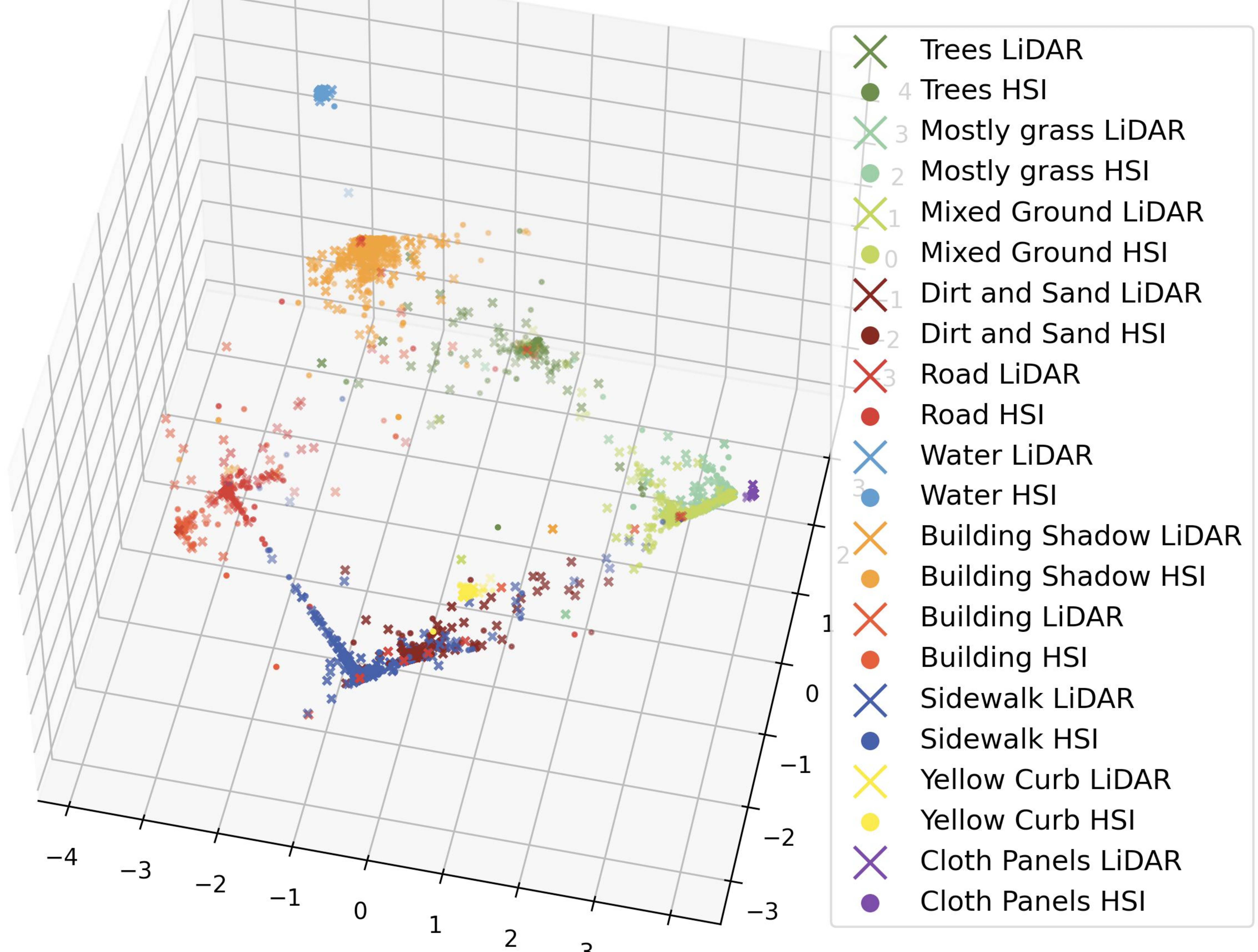} }}
        \hspace{0.01em}
        \subfloat[\footnotesize{Berlin Data Embeddings}]{{\includegraphics[width=5.8cm]{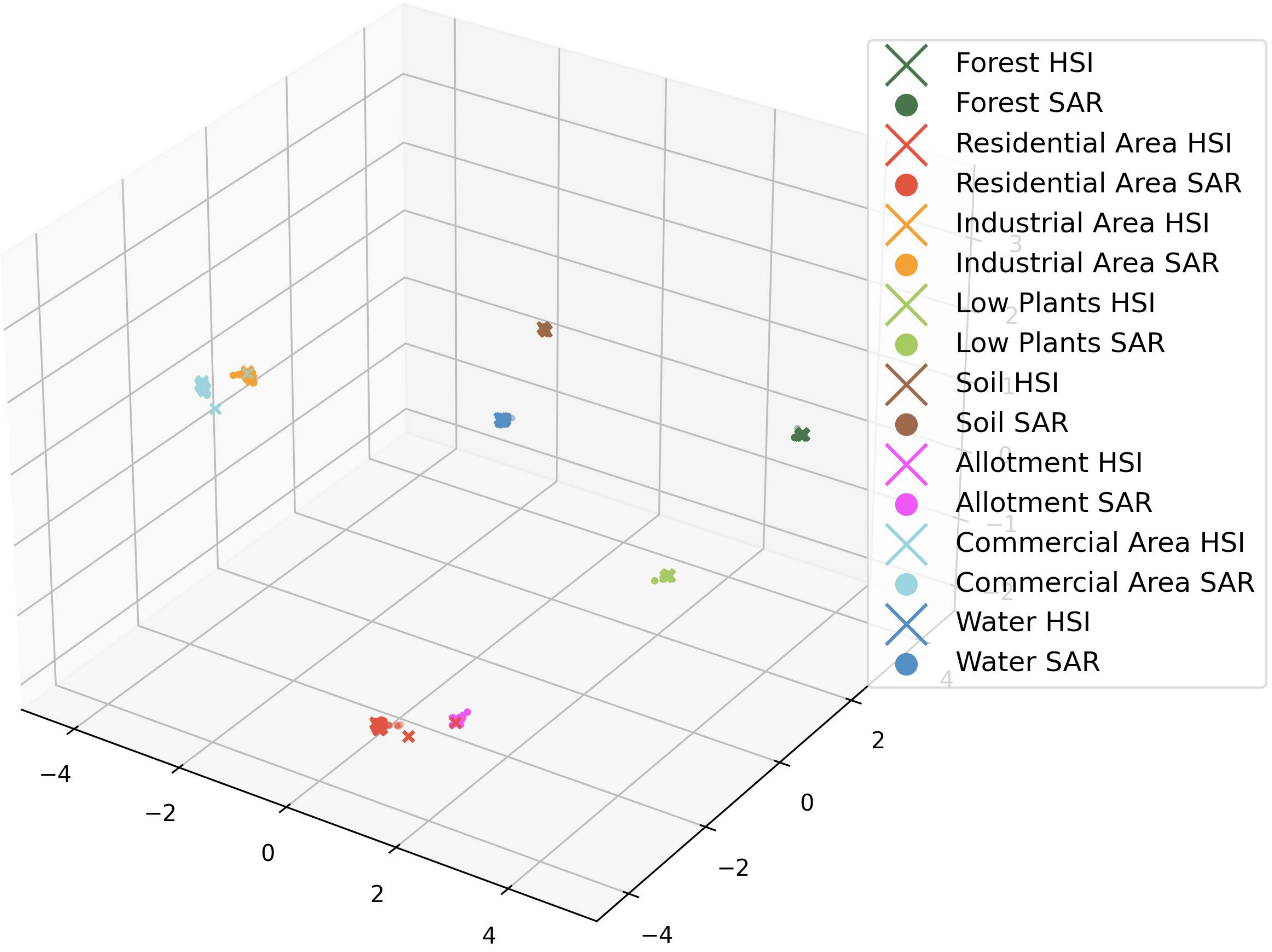} }}
        \caption{Visualization of shared embeddings from different datasets with the Similarity Enhancement (SE) term}
        \label{fig:ScatterSE}
    \end{figure*}

\begin{figure*}[!ht]
        \centering
        \subfloat[\footnotesize{AVIRIS-NG/ NEON Data Embeddings}]{{\includegraphics[width=5.9cm]{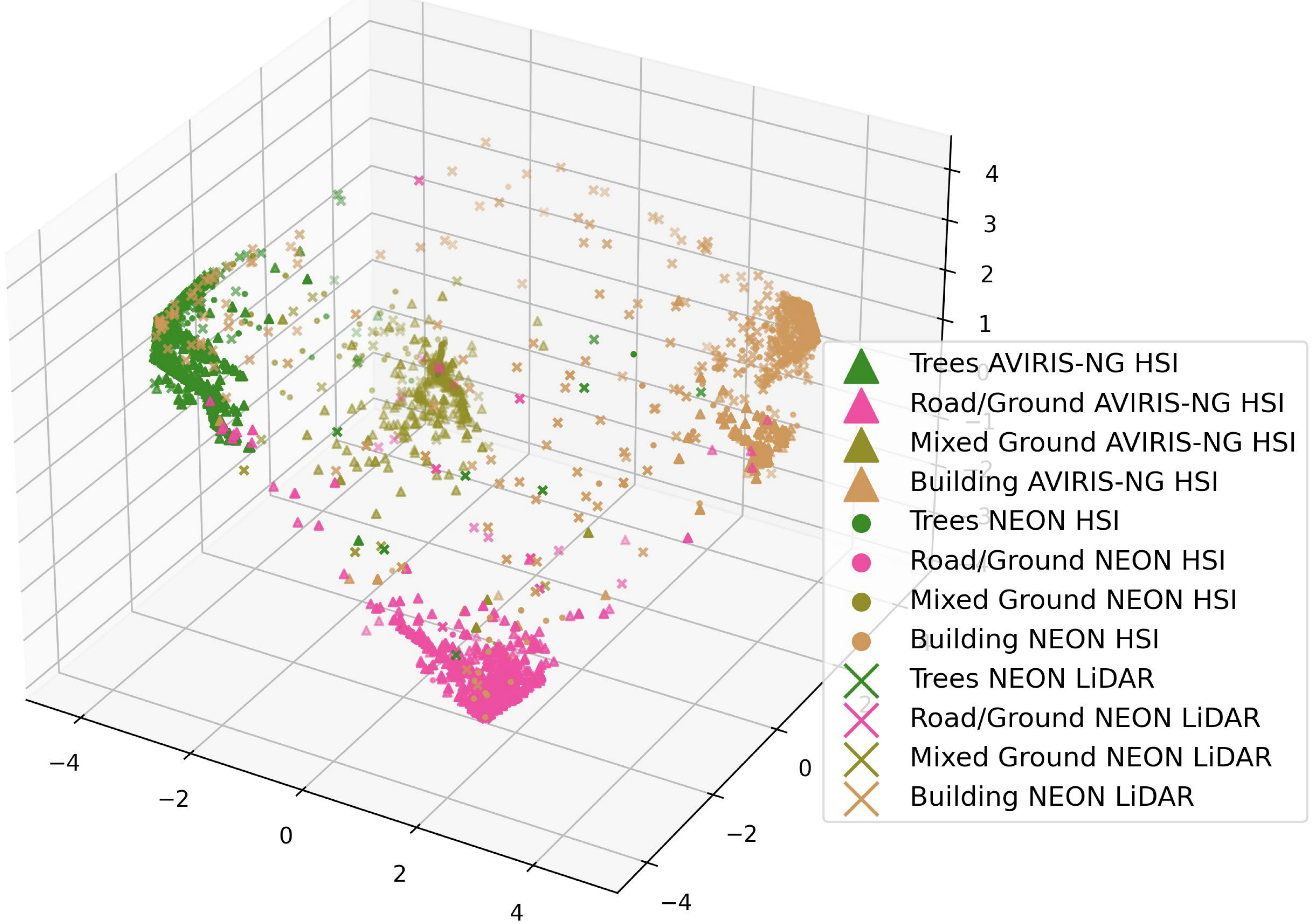} }}
        \hspace{0.01em}
        \subfloat[\footnotesize{MUUFL Data Embeddings}]{{\includegraphics[width=5.8cm]{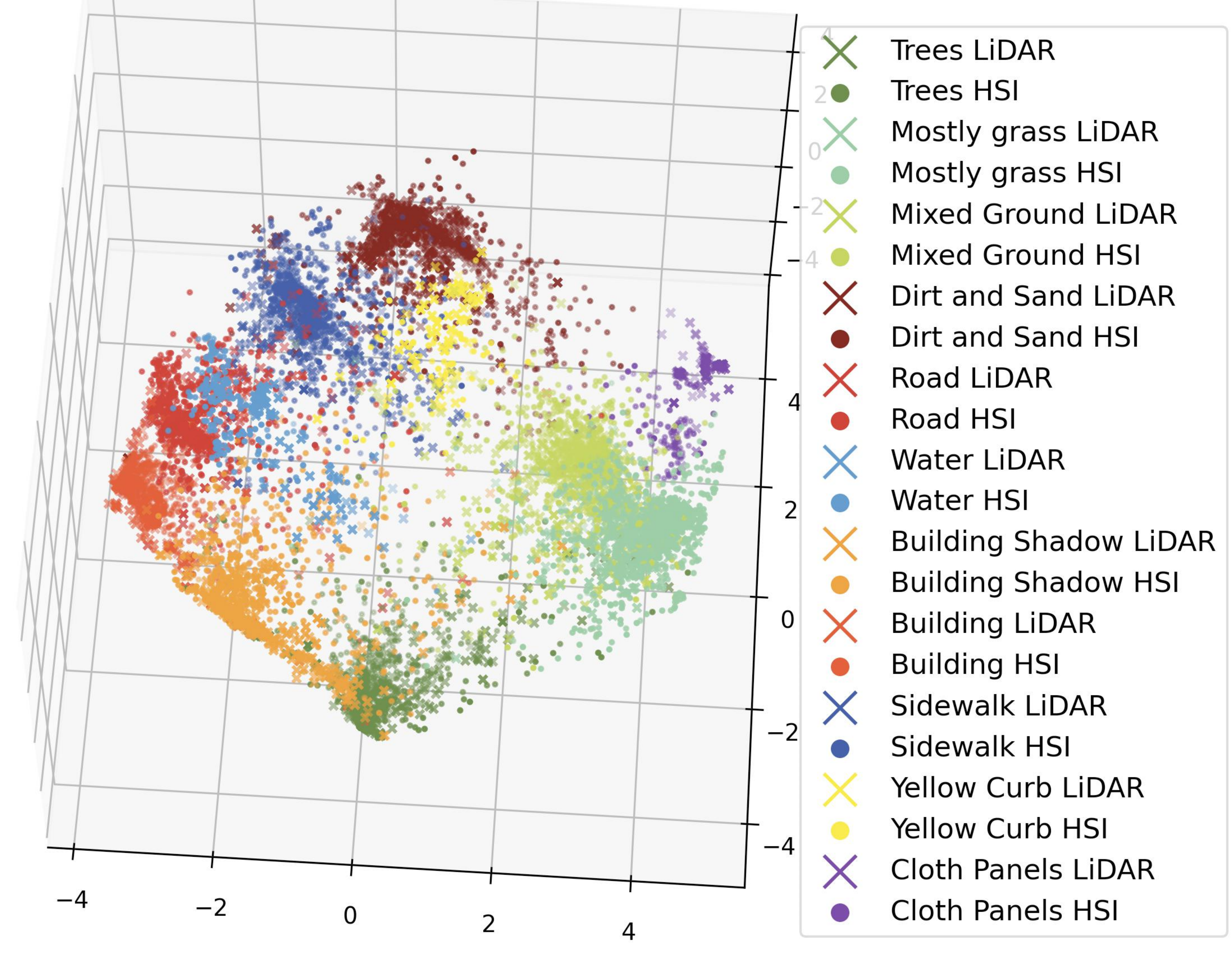} }}
        \hspace{0.01em}
        \subfloat[\footnotesize{Berlin Data Embeddings}]{{\includegraphics[width=5.8cm]{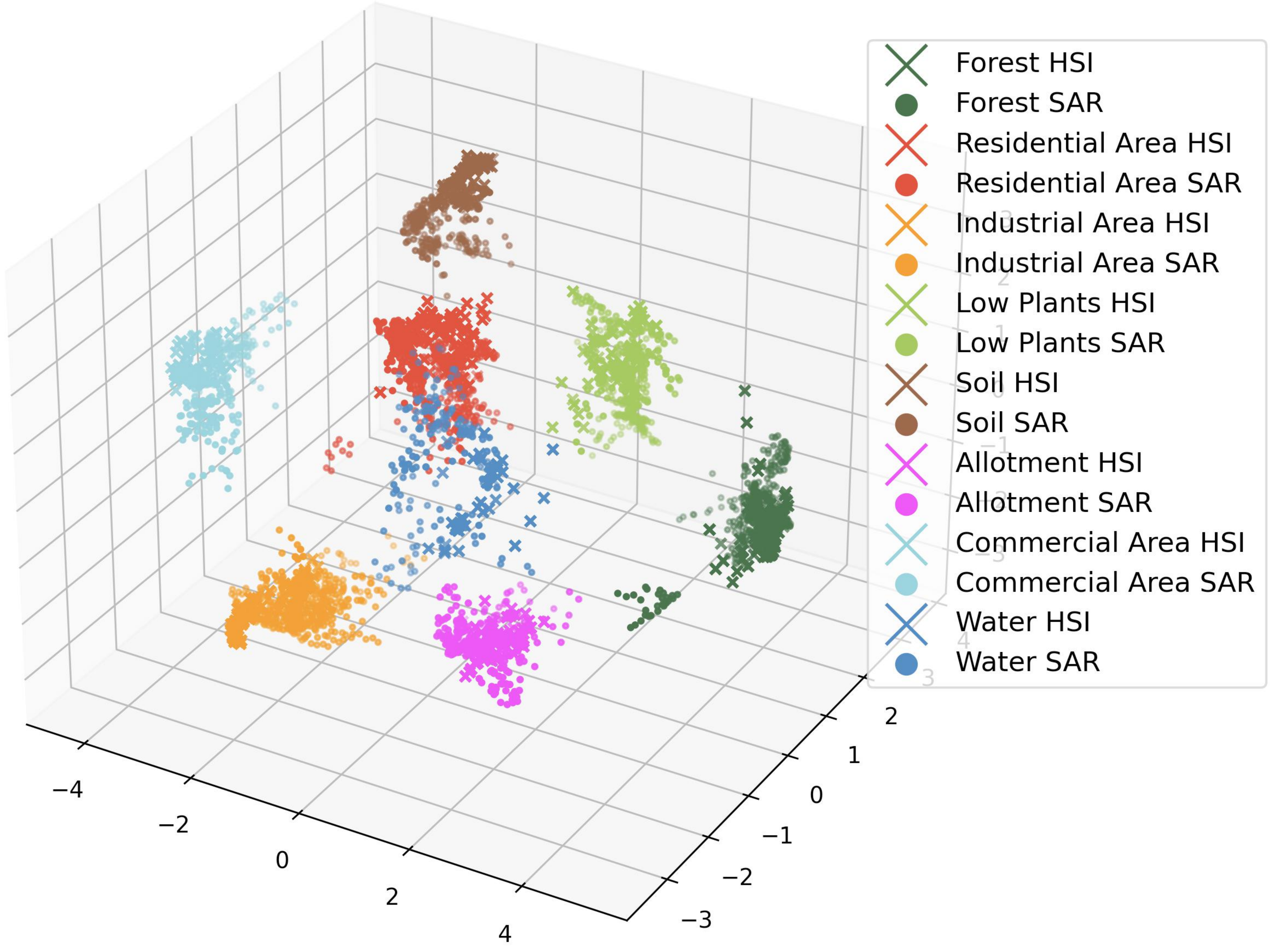} }}
        \caption{Visualization of shared embeddings from different datasets without the Similarity Enhancement (SE) term}
        \label{fig:ScatterWSE}
    \end{figure*}

\begin{figure*}[!ht]
        \centering
        \subfloat[\footnotesize{Comparison of models trained on the NEON HSI Data}]{{\includegraphics[width=5.8cm]{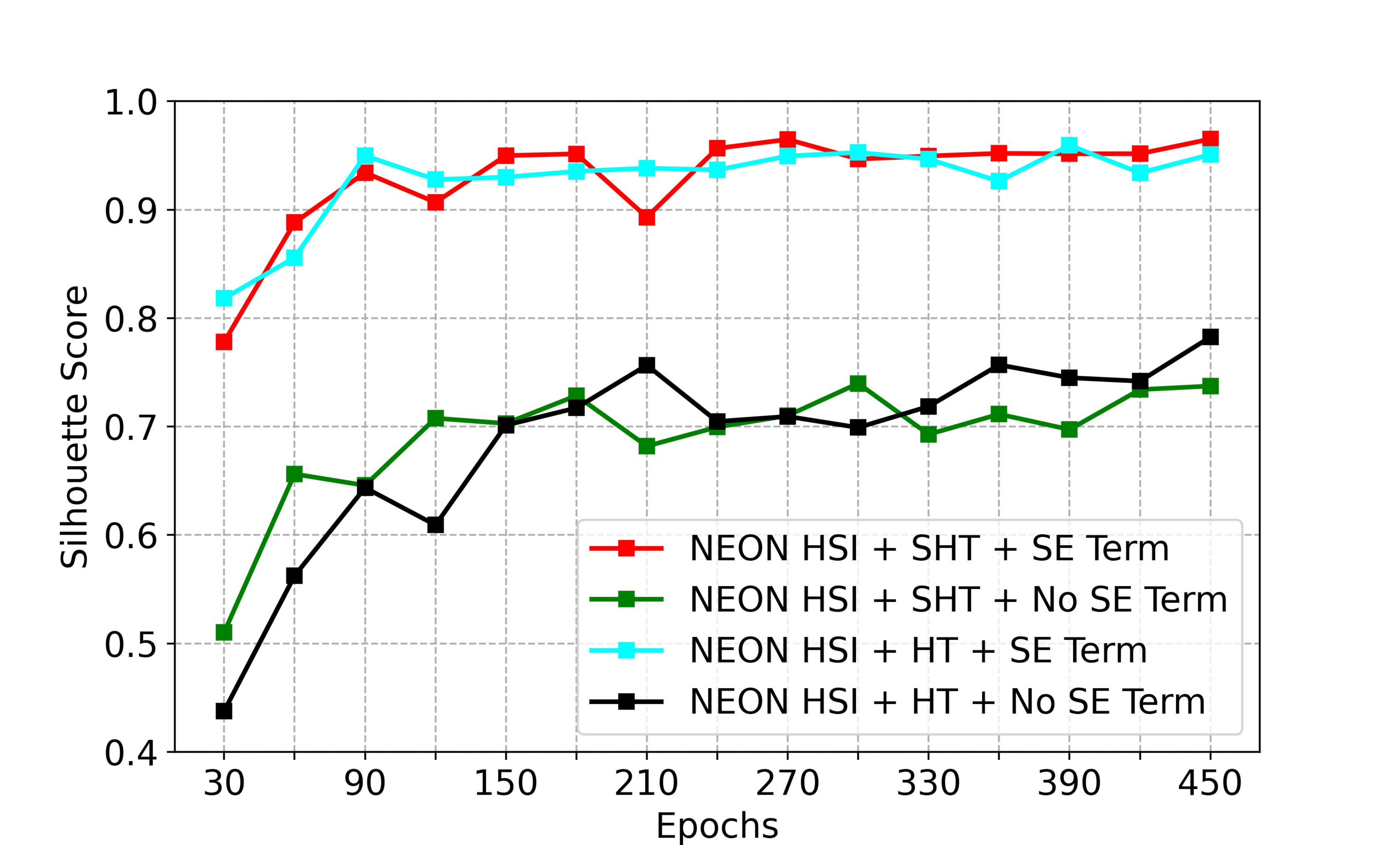} }}
         \hspace{0.02em}
        \subfloat[\footnotesize{Comparison of models trained on the AVIRIS-NG HSI Data }]{{\includegraphics[width=5.8cm]{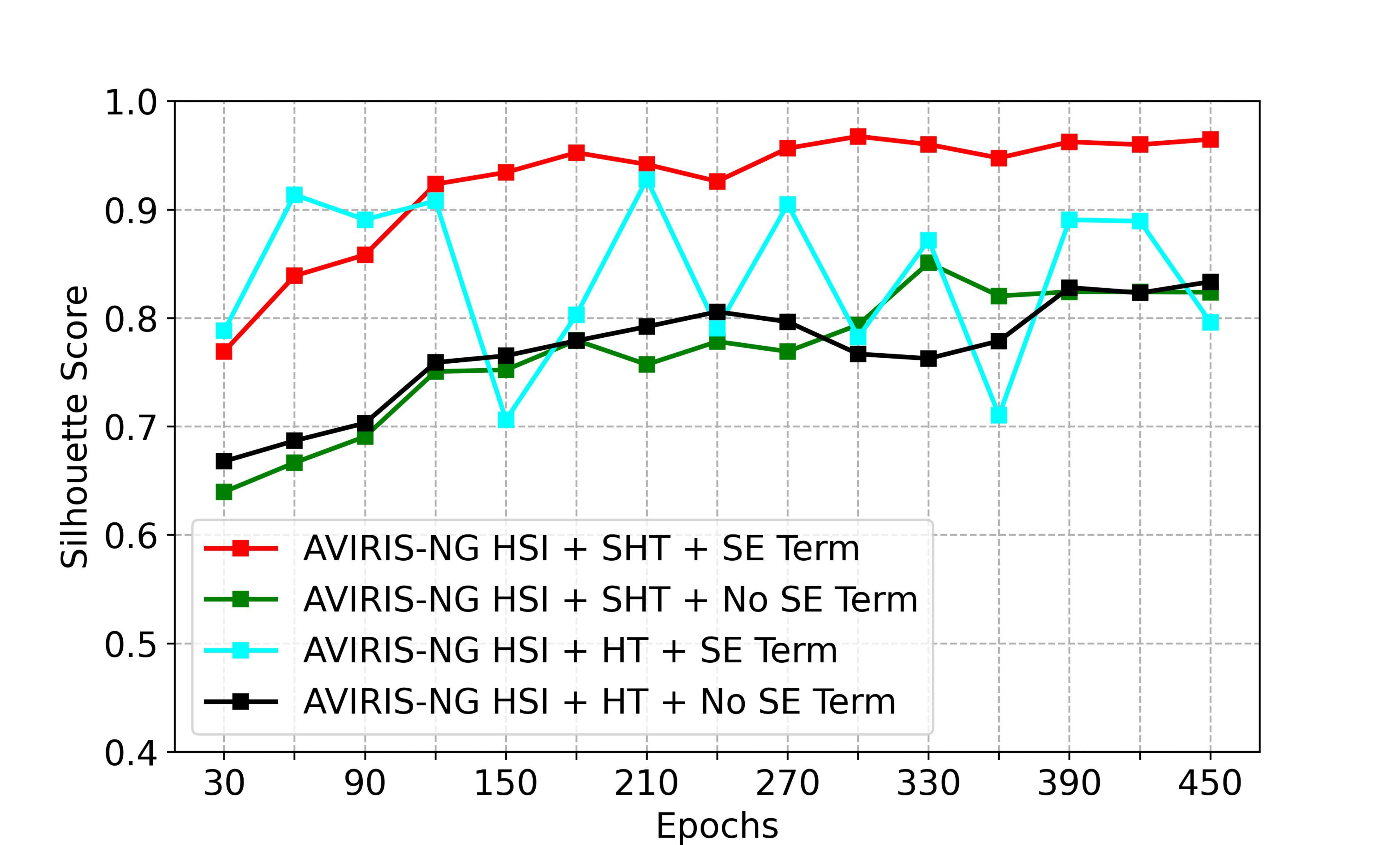} }}
         \hspace{0.02em}
        \subfloat[\footnotesize{AVIRIS-NG/NEON Data: Effect of the Sensor used for the Triplet Network }]{{\includegraphics[width=5.8cm]{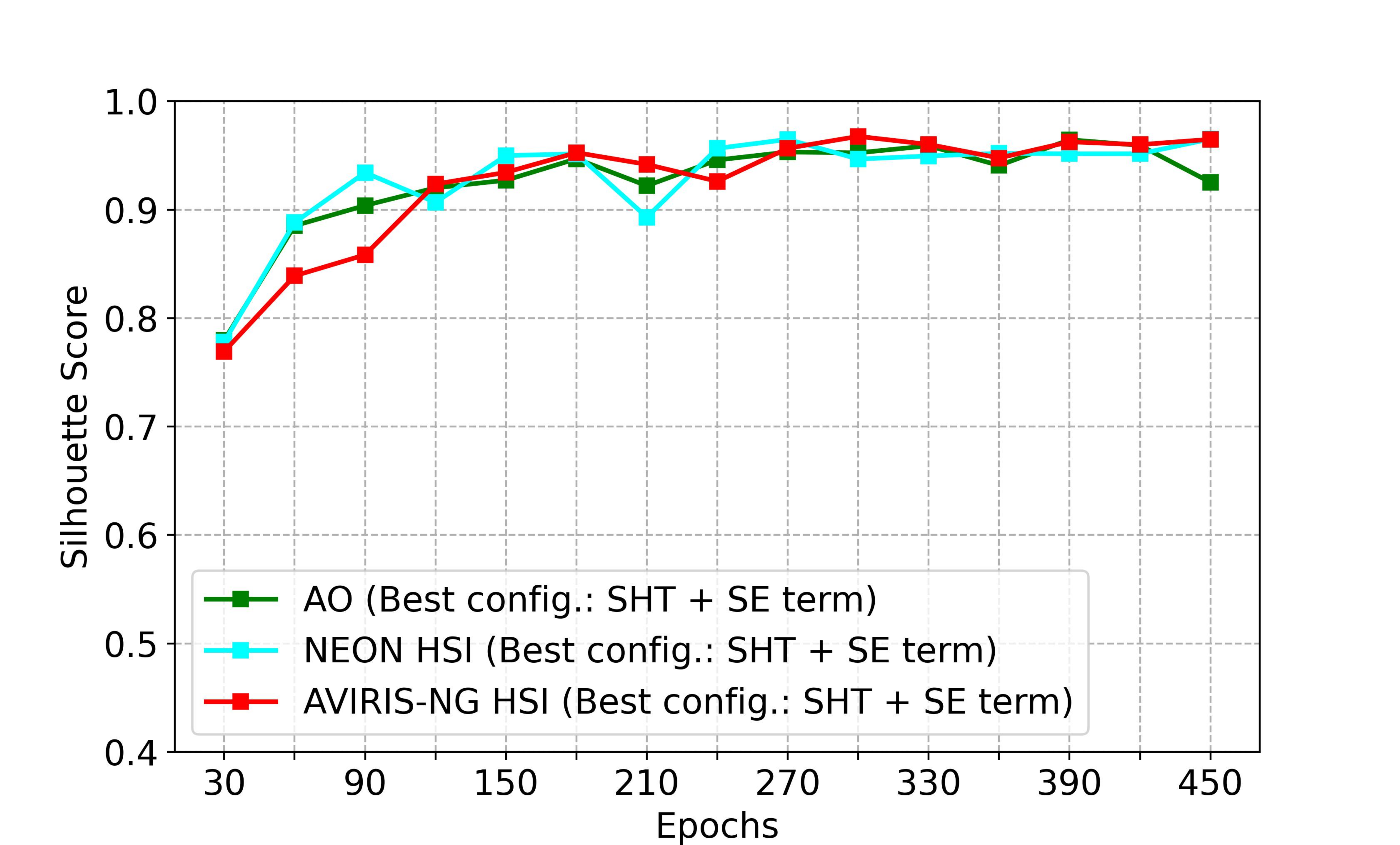} }}
        \\
        \subfloat[\footnotesize{Models trained on the MUUFL HSI Data }]{{\includegraphics[width=5.8cm]{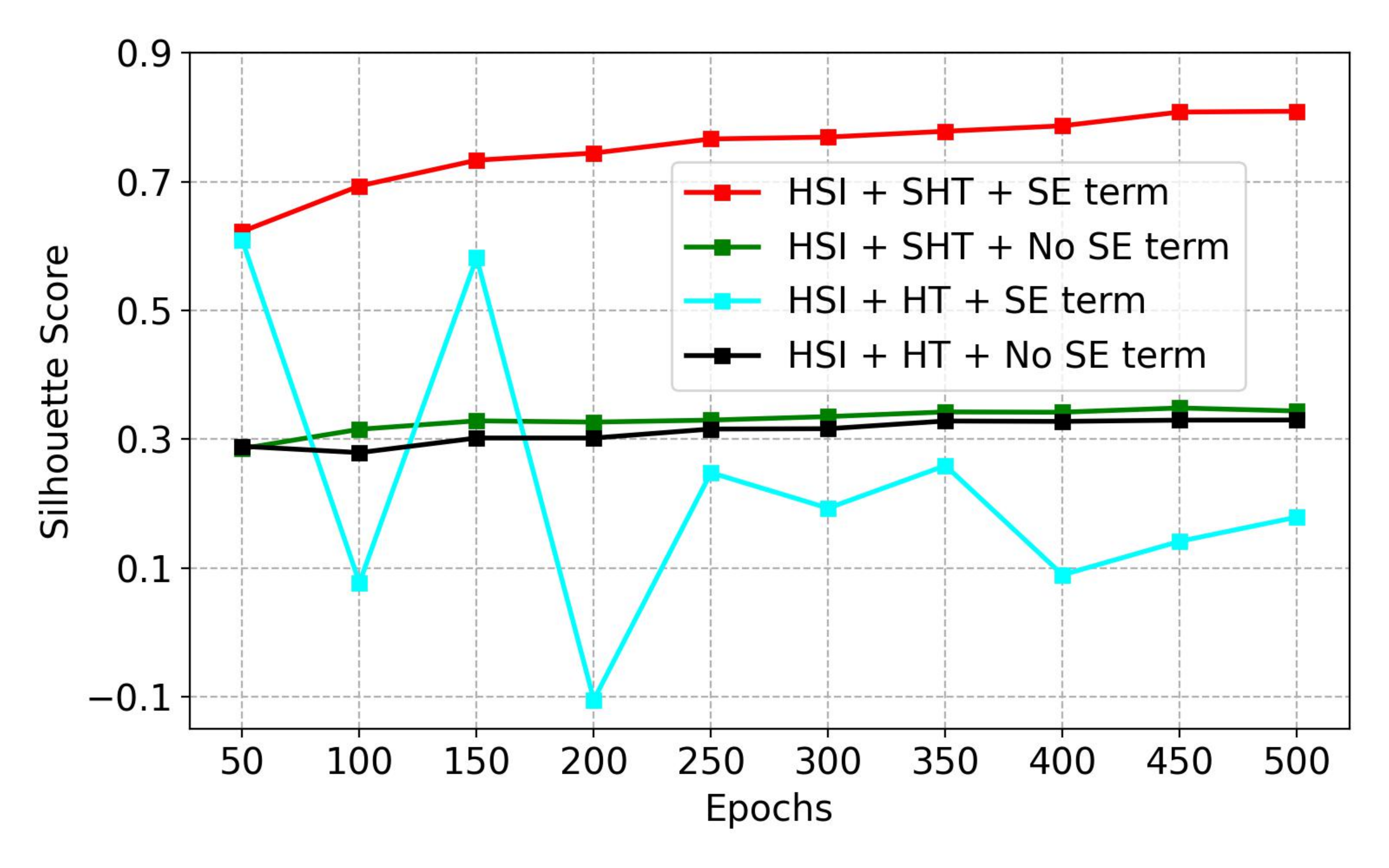} }}
        \hspace{0.02em}
        \subfloat[\footnotesize{Models trained on the MUUFL LiDAR Data}]{{\includegraphics[width=5.8cm]{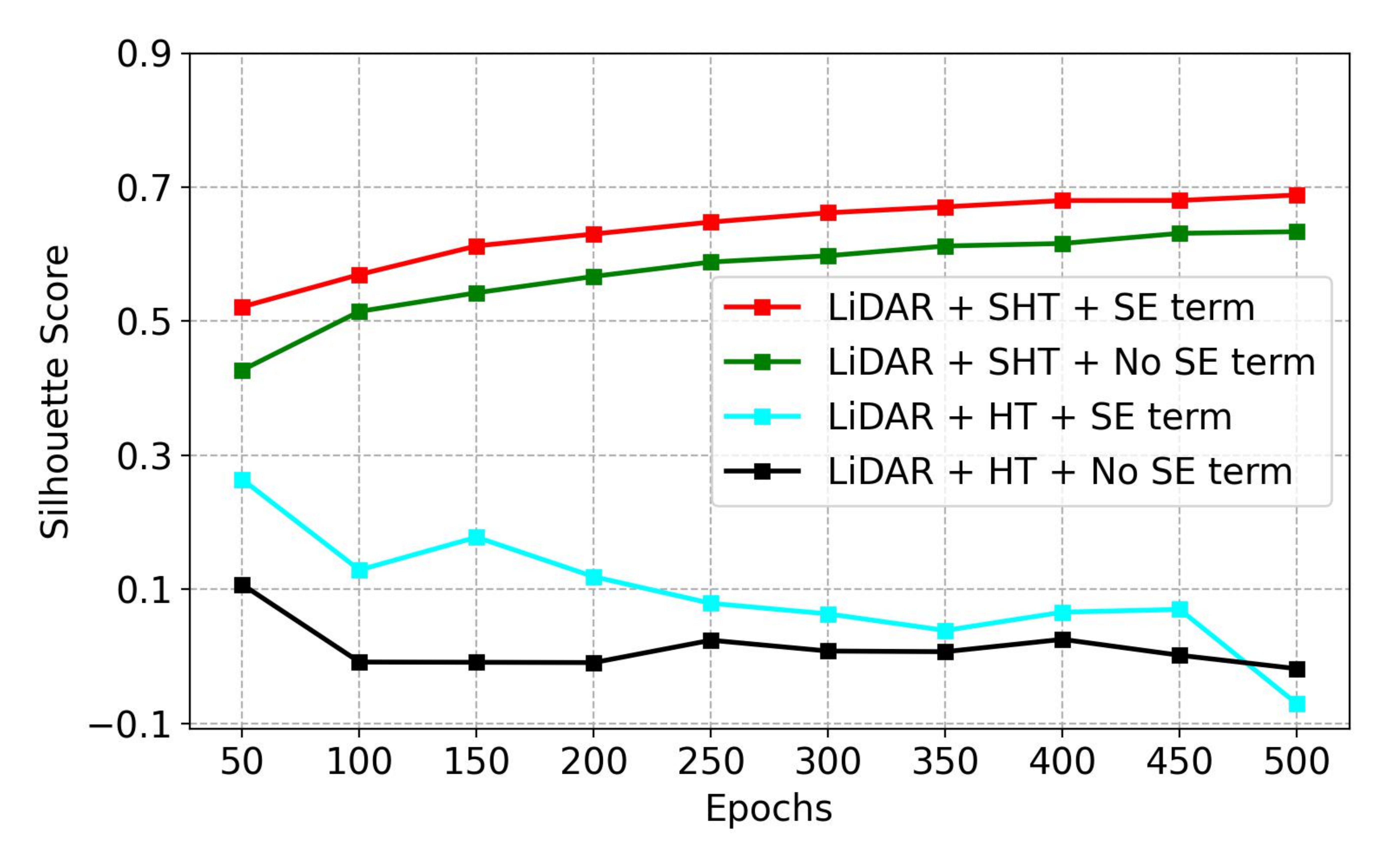} }}
        \hspace{0.02em}
        \subfloat[\footnotesize{MUUFL Data: Effect of the Sensor used for the Triplet Network }]{{\includegraphics[width=5.8cm]{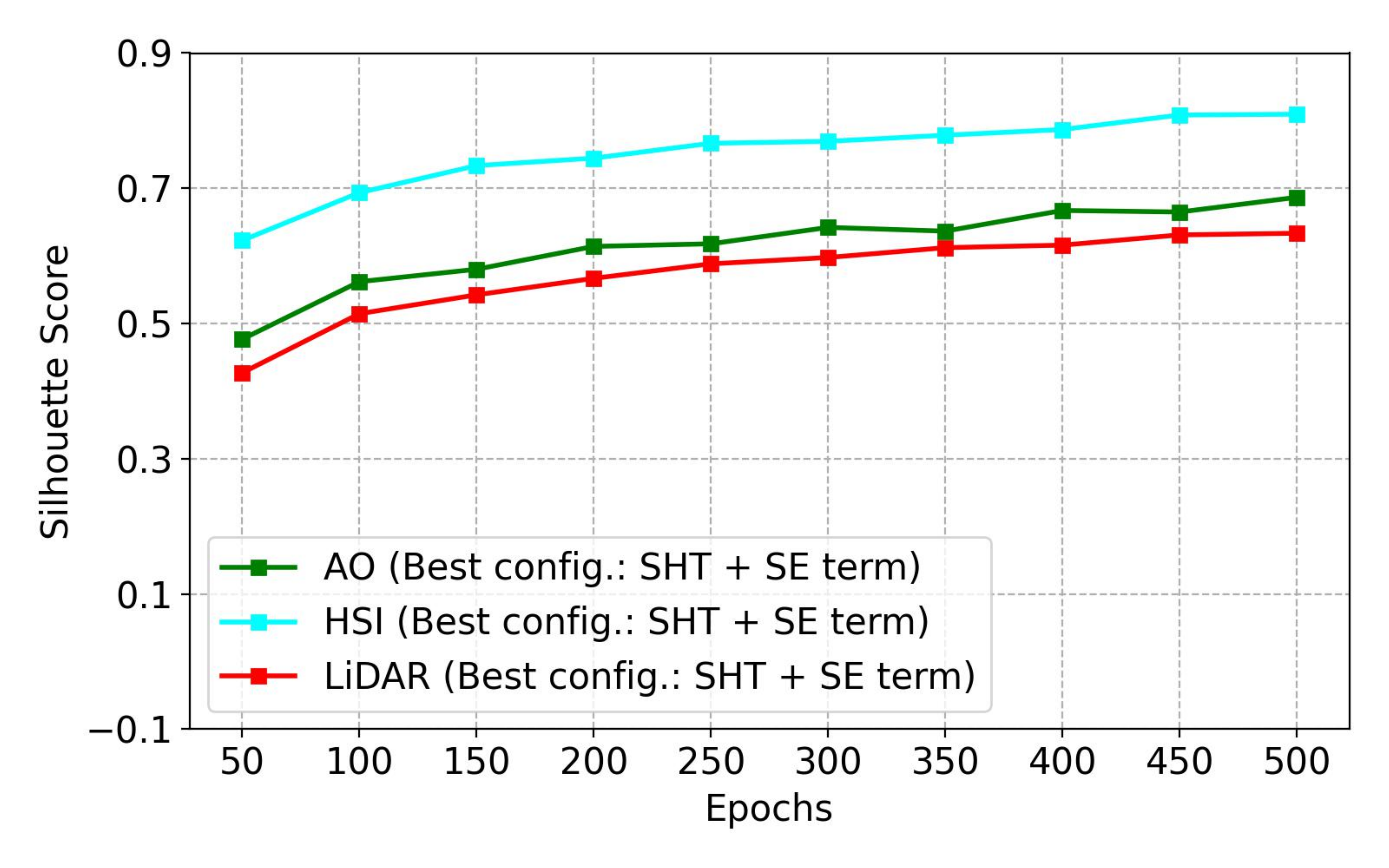} }}
        \\
        \centering
        \subfloat[\footnotesize{Models trained on the Berlin HSI Data }]{{\includegraphics[width=5.8cm]{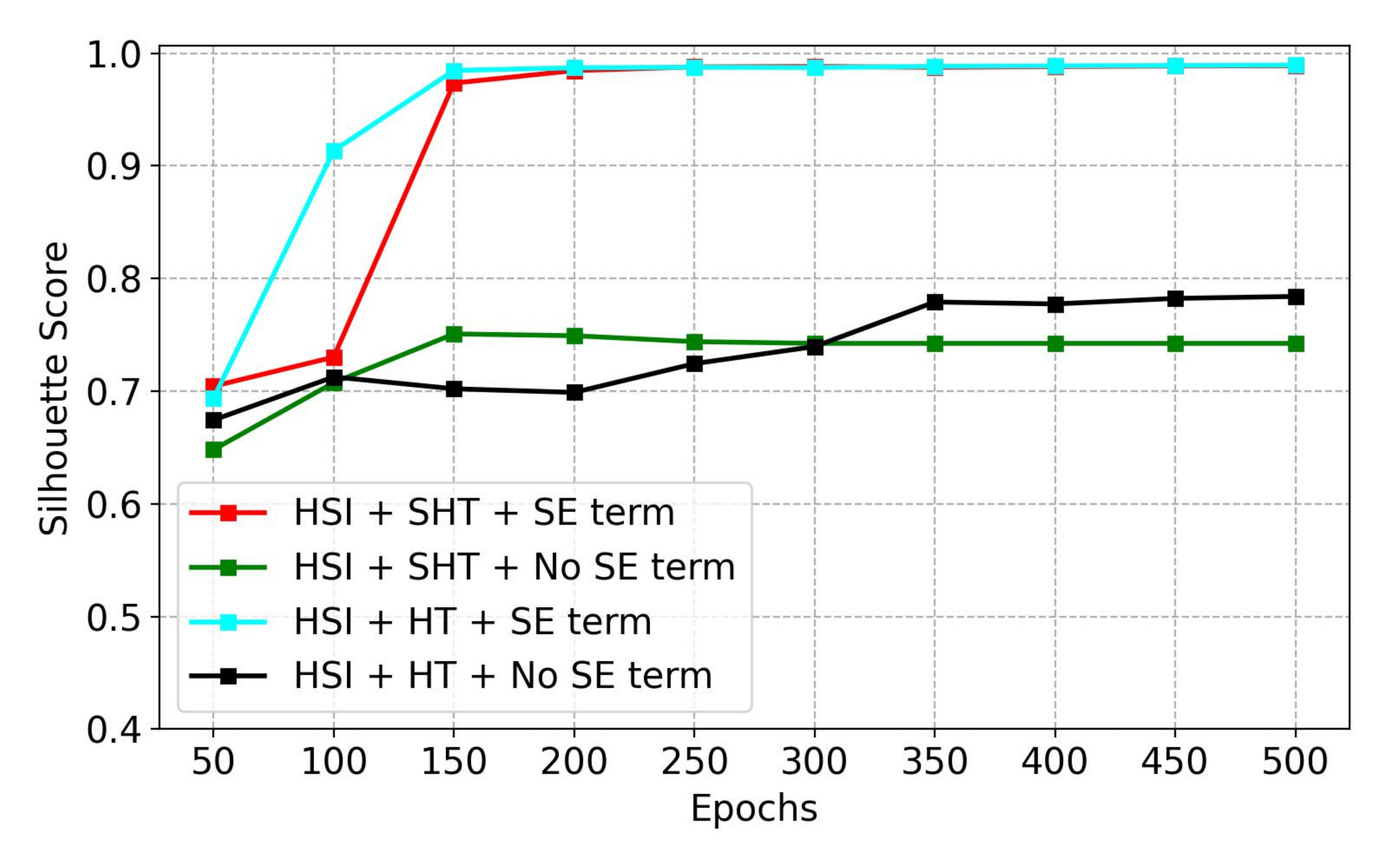} }}
        \hspace{0.02em}
        \subfloat[\footnotesize{Models trained on the Berlin SAR Data}]{{\includegraphics[width=5.8cm]{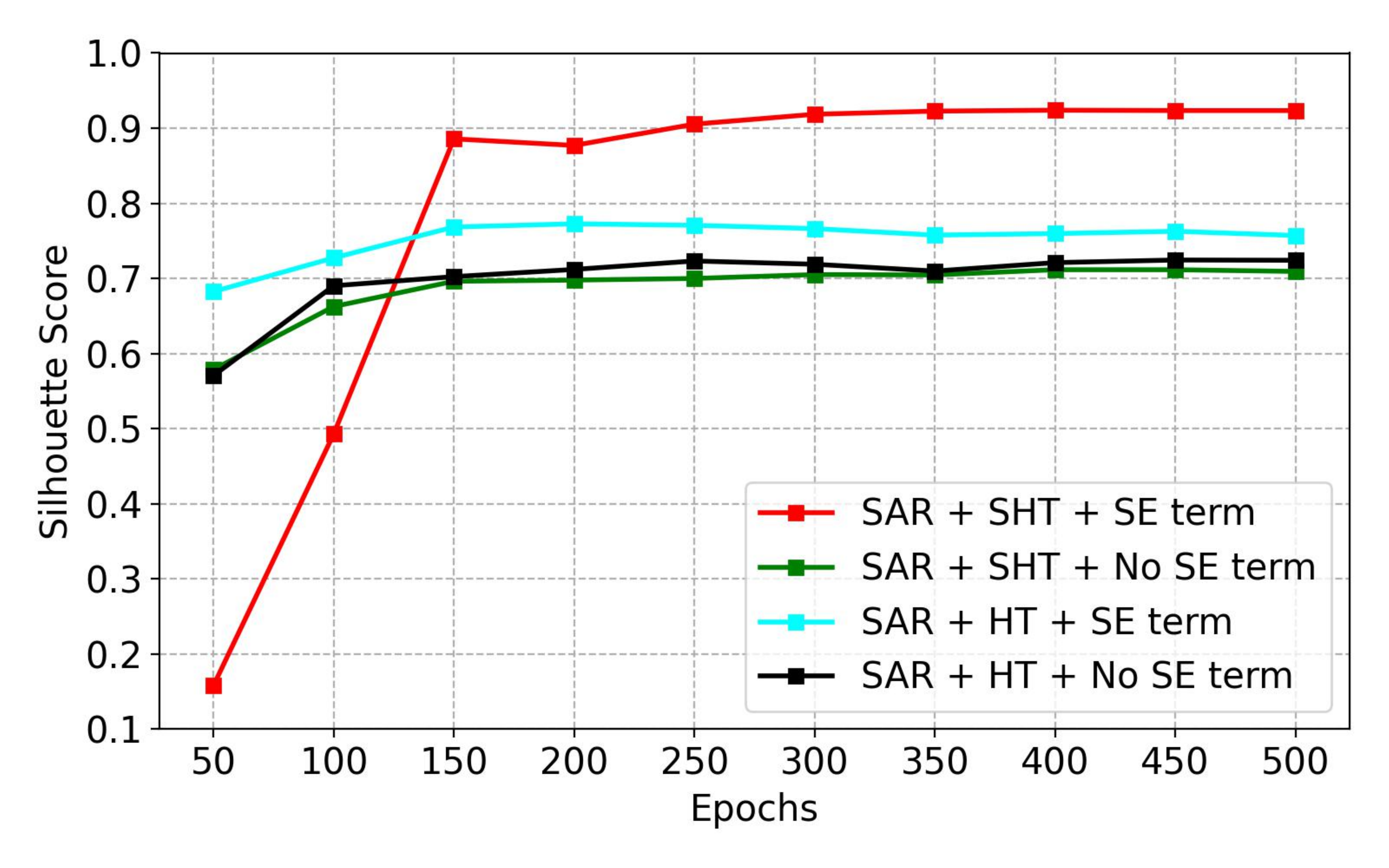} }}
        \hspace{0.02em}
        \subfloat[\footnotesize{Berlin Data: Effect of the Sensor used for the Triplet Network }]{{\includegraphics[width=5.8cm]{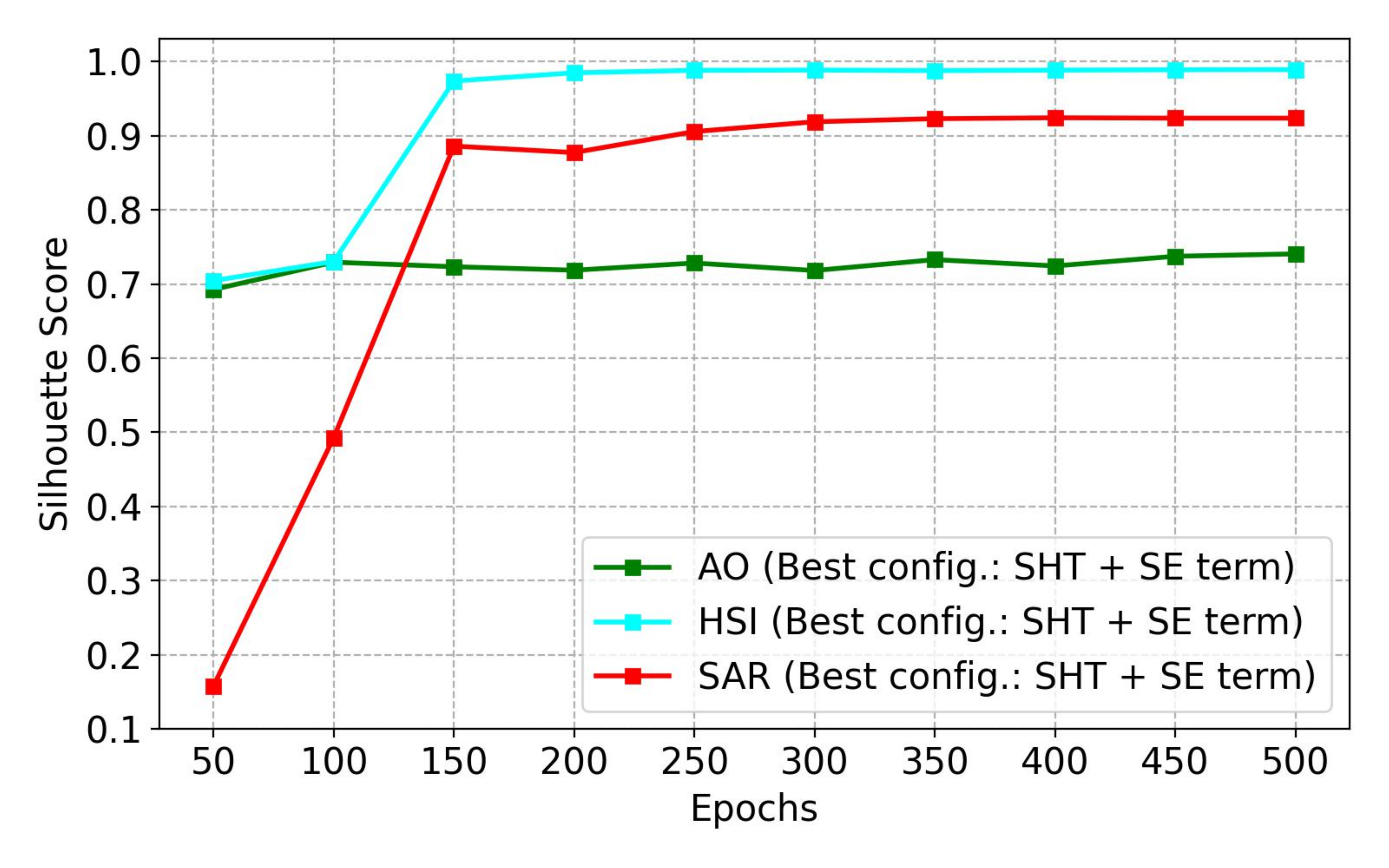} }}

        \caption{Results of the Ablation studies conducted on the similarity enhancement term, triplet selection strategy, and sensor used for the triplet network are shown here. The acronyms are: HT for Hard Triplets, SHT for Semi-Hard Triplets, SE for Similarity Enhancement, AO for Alternating Optimization, HSI for Hyperspectral Imagery, LiDAR for Light detection and ranging, and SAR for Synthetic Aperture Radar.}

        \label{fig:AblationGraph}
    \end{figure*}

However, in a missing sensor scenario, it is quite the opposite. In this case, the value of $\gamma$ should be higher. A higher value of $\gamma$ will bring the latent spaces of both sensors close to each other and form tightly packed clusters. Now, it is easier to predict one sensor's latent space from another sensor's latent space which makes the sensor translation process more accurate. In the MUUFL data, for the classification task, a $\gamma$ between $0$ and $0.1$ shows the best performance. For predicting a missing sensor's latent space, the optimal value of $\gamma$ was 0.4. For the Berlin dataset, a $\gamma$ of 0.4 worked best for both classification and the missing sensor's latent space prediction. The embeddings with and without the SE term are shown in Fig. \ref{fig:ScatterSE} and Fig. \ref{fig:ScatterWSE}, respectively. The embeddings with the SE term appear to be more compact, which shows the improvement in clustering after incorporating the SE term in the loss function. In Fig. \ref{fig:AblationGraph} also, it is shown that the models trained using the SE term perform significantly better than the models trained without the SE term.

\subsection{Effect of the triplet mining strategy} \label{sssec:triplet_effect}
Experiments are conducted using semi-hard and hard triplets. The silhouette scores are computed for comparison of performance. If the dataset contains fewer outliers, both semi-hard and hard triplet strategies show a similar performance. However, if the dataset contains a significant number of outliers, then semi-hard triplets seem to be a better option because the hard triplets strategy is sensitive to the presence of outliers. Additionally, a few easy triplets can be added to the semi-hard triplets to make the training smoother. The effect of the triplet selection strategy is shown in Fig. \ref{fig:AblationGraph}. \par 
In the case of AVIRIS-NG/ NEON data, both the strategies show a similar performance irrespective of the sensor used for the triplet network. It is due to the fewer classes and less ambiguity among them. However, in MUUFL data, both strategies show a similar result when the triplet network is applied on HSI data without using the SE term. But when the SE term is applied, and HSI data is used for the triplet network, the results of the semi-hard triplet mining strategy are significantly better than the hard triplet mining strategy. When the LiDAR data is used for the triplet network, then also the semi-hard triplet strategy performs significantly better than the hard triplet strategy. For training the MUUFL data, 40\% of the samples were used, and there is a significant variation in the distribution of some classes. On such data, using the semi-hard triplet mining strategy makes the training process more stable. Therefore, the semi-hard triplet strategy outperforms the hard triplet mining strategy in this case. \par 

The Berlin data also shows slightly better performance using semi-hard triplets when SAR data is used for the triplet network along with the SE term. When the SE term is not applied, and the SAR data is used for the triplet network, both the strategies yield a similar result. Both the strategies show a similar performance when the HSI data is used for the triplet network because the Berlin Dataset contains very few samples ($<=$ 443) from each class, and there are very few outliers.

\subsection{Effect of the sensor used for the Triplet Network}
The sensor chosen for the triplet network significantly affects the model's performance. If a sensor is more capable of distinguishing between different classes compared to the other sensors, then using its data as input to the triplet network gives better results. If both the sensors are comparable, then the triplet network can be used on either of the sensors, and a similar performance is observed. A comparison is shown in Fig. \ref{fig:AblationGraph}. Using the triplet network on the better sensor in all the datasets yields better performance. However, the performance drops when a triplet model is applied on a sensor with a low resolution or a low discriminative ability.

\subsection{Effect of Alternating Optimization}
One obvious concern is how the performance is affected if a triplet network is used on both sensors. A triplet network can be used on both the sensors, but it is computationally expensive because now we have a large number of triplets from two sensors. However, to reduce the computations significantly, we propose an Alternating Optimization approach. As mentioned in Section \ref{sec:Method}, the CoMMANet is trained in several checkpoints, and triplets are computed at the beginning of every checkpoint. In Alternating Optimization, the triplet network is used on one sensor in one checkpoint and the other sensor in the next checkpoint. The triplet network alternates between both the sensors in this way. This ensures that both the sensors' embeddings are clustered efficiently when the triplet network is applied.  A comparison is shown in Fig. \ref{fig:AblationGraph}. The best training parameters configuration from each sensor is used for comparison. \par

In AVIRIS-NG/ NEON data, using a triplet network on NEON HSI, NEON LiDAR, and Alternating Optimization shows almost the same performance. The reason is that both the sensors are hyperspectral sensors. The AVIRIS-NG HSI has a lower resolution, but it still captures more information than LiDAR. Therefore, both NEON HSI and AVIRIS-NG HSI are capable of distinguishing between different classes accurately.\par

In the case of MUUFL data also, the best results are obtained using the triplet model on HSI. When the LiDAR is chosen for the triplet network, the performance is worst because LiDAR alone cannot identify all land-cover classes. The alternating optimization model lies in the middle. In the case of Berlin data also, using the triplet network on HSI gives the best performance. The second-best model uses SAR on the triplet network. It is because HSI can identify the land-cover classes with more precision compared to SAR. The Alternating Optimization model shows the worst performance. The nature of the Alternating Optimization model is slightly unpredictable. However, the models trained using the better sensor always yield the best results, and Alternating Optimization does not improve the model's clustering performance.

\section{Conclusion}
\label{sec:Conclusion}
In this paper, a novel architecture, called CoMMANet, is proposed, which can map data from heterogeneous modalities onto a shared manifold in a discriminative manner. The proposed model can cluster the target classes from all the sensors effectively. Additionally, the proposed architecture allows missing sensor data reconstruction or sensor translation. The fused embeddings from all the sensors allow a robust and accurate classification. The discriminative ability of CoMMANet embeddings allows robust classification using even a simple method like KNN also. However, the proposed CoMMANet is not limited to remote sensing applications. The CoMMANet is a generalized architecture that can be used on any kind of multi-modal data like audio, video, text, RGB images, and ECG. Additionally, different features like Gabor filters, residual connections, dilated convolutions, and the Attention mechanism can be incorporated into the CoMMANet architecture to improve the performance. Experimental results and comparison with the state-of-the-art multi-modal classification methods indicate the effectiveness of the proposed CoMMANet. However, the current architecture still needs some improvements. Firstly, the model requires many samples from each class to generalize on unseen samples. Secondly, the model needs to be equipped with a better fusion strategy. The classification can be further improved if the embeddings contain complementary information from multiple sources. Thirdly, the sensor translation process needs to be improved to predict the missing sensor data with higher accuracy. Our future work will focus on overcoming these shortcomings and developing a more efficient architecture.

\section*{Acknowledgement}
\label{sec:Acknowledgement}
The authors would like to thank the Geomatics Laboratory, Geography Department, Humboldt-Universität zu Berlin, for providing the simulated EnMAP hyperspectral data for the Berlin Urban area.\par
This material is based upon work supported by the National Science Foundation under Grant No. CNS-1747783.

\bibliographystyle{IEEEtran}
\bibliography{References}

\begin{IEEEbiography}[{\includegraphics[width=1in,height=1.25in,clip,keepaspectratio]{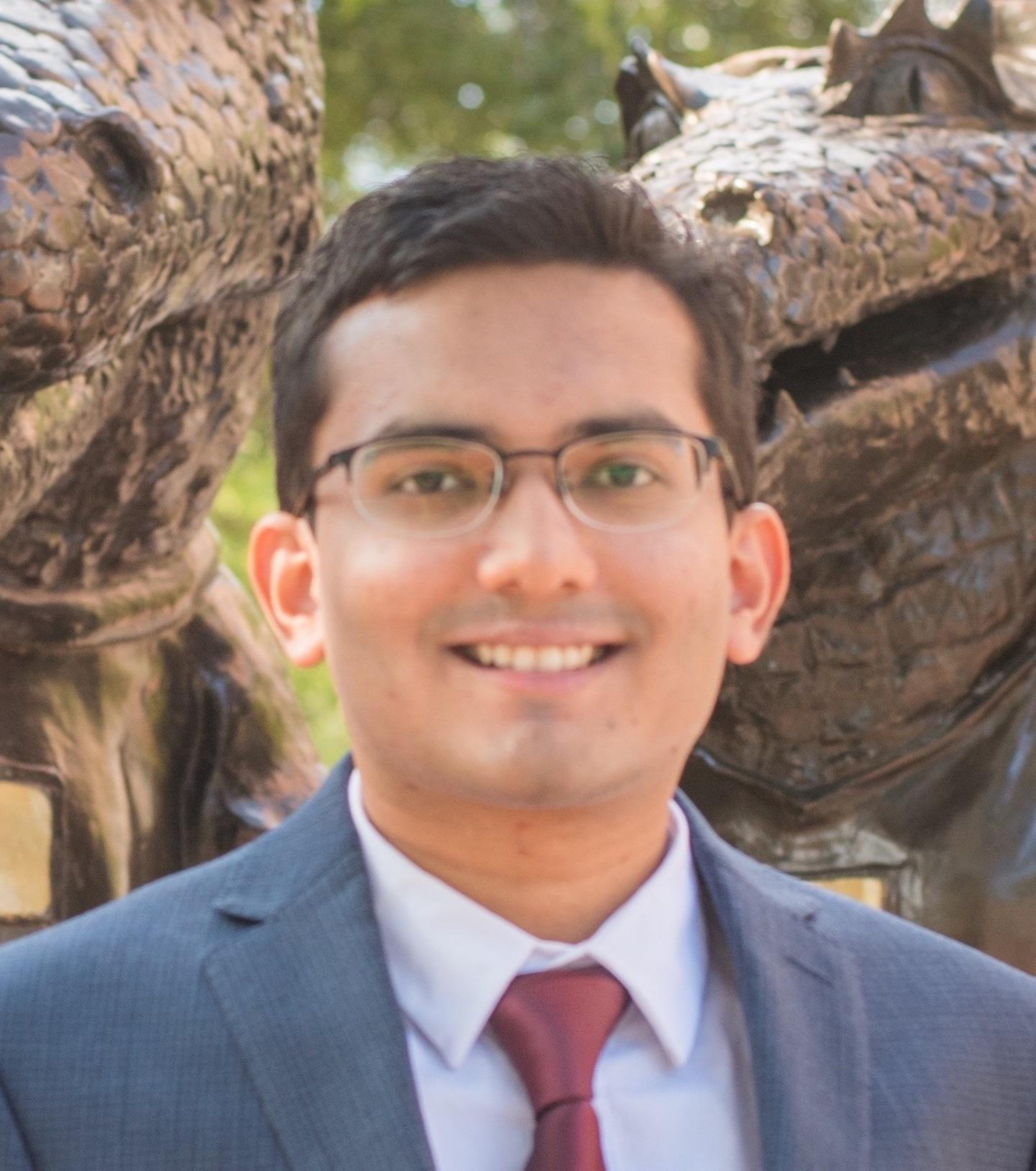}}]{Aditya Dutt}(Graduate Student Member, IEEE) received his M.S. degree in computer science from the University of Florida, Gainesville, FL, USA, in 2019. He is currently pursuing a Ph.D. degree in computer science with the Department of Computer and Information Science and Engineering, University of Florida.\par 
His research interests include machine learning, metric learning, multimodal data fusion, speech analysis, and speech emotion recognition.
\end{IEEEbiography}

\begin{IEEEbiography}[{\includegraphics[width=1in,height=1.25in,clip,keepaspectratio]{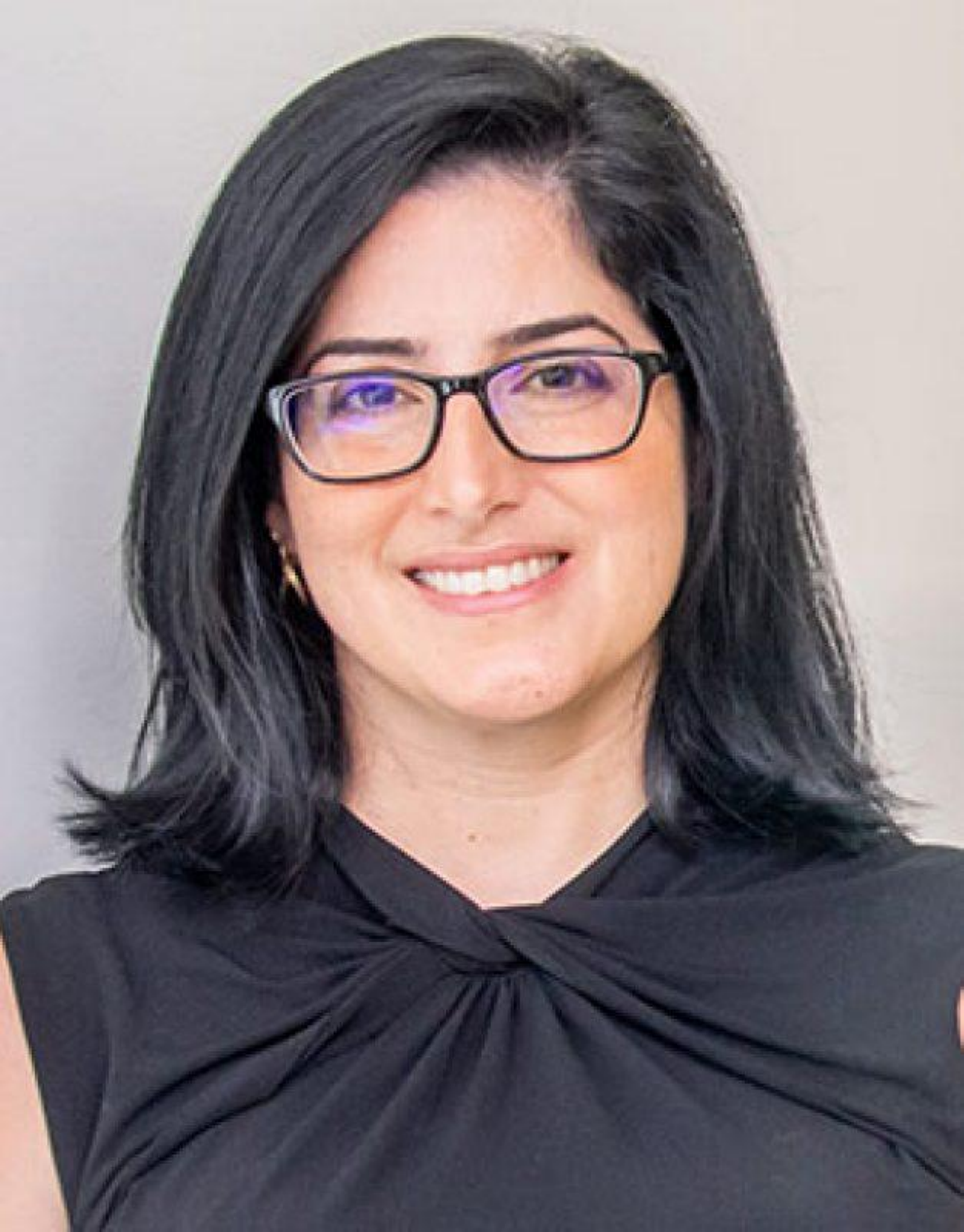}}]{Alina Zare} (Senior Member, IEEE) received a Ph.D. degree in Computer and Information Science and Engineering from the University of Florida, Gainesville, FL, USA, in 2008.\par
Alina Zare teaches and conducts research in the area of machine learning and artificial intelligence as a Professor in the Electrical and Computer Engineering Department at the University of Florida. Dr. Zare’s research has focused primarily on developing new machine learning algorithms to automatically understand and process data and imagery. Her research work includes automated plant root phenotyping, sub-pixel hyperspectral image analysis, target detection, and underwater scene understanding using synthetic aperture sonar, LIDAR data analysis, Ground Penetrating Radar analysis, and buried landmine and explosive hazard detection.\par 
She is currently an Associate Editor for the IEEE Transactions on Artificial Intelligence.
\end{IEEEbiography}

\begin{IEEEbiography}[{\includegraphics[width=1in,height=1.25in,clip,keepaspectratio]{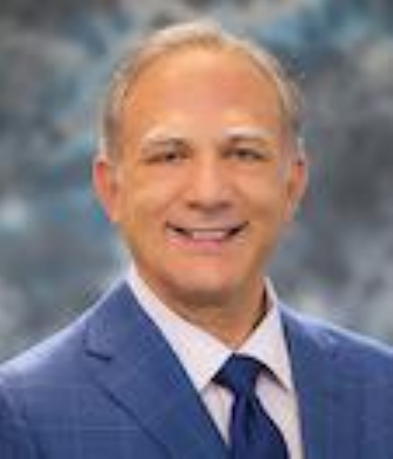}}]{Paul Gader} (Fellow, IEEE) received a Ph.D. degree in mathematics for image-processing-related research from the University of Florida, Gainesville, FL, USA, in 1986. \par
He is a Professor with the Department of Computer and Information Science and Engineering and the Engineering School of Sustainable Infrastructure and Environment, University of Florida. He performed his first research in image processing in 1984, working on algorithms for detecting bridges in forward-looking infrared imagery as a Summer Student Fellow at Eglin Air Force Base. He has been a leading figure in handwriting recognition and landmine detection. He led the development of a 5th-ranked handwritten character recognizer and a top-ranked handwritten word recognizer in two National Institute of Standards and Technology (NIST) competitions in the early 1990s. He has published over 100 journals and over 300 total papers and was an Associate Editor of IEEE Geoscience and Remote Sensing Letters.\par 
He has worked on a wide variety of theoretical and applied research problems, including fast computing with linear algebra, mathematical morphology, fuzzy sets, Bayesian methods, handwriting recognition, automatic target recognition, biomedical image analysis, landmine detection, human geography, and hyperspectral and light detection, and ranging image analysis projects.
\end{IEEEbiography}

\end{document}